\documentclass[review]{elsarticle}

\usepackage{lineno,hyperref}
\modulolinenumbers[5]

\journal{Elsevier}

\usepackage{url}
\usepackage{makeidx}         
\usepackage{graphicx}        
\usepackage{multicol}        
\usepackage[bottom]{footmisc}

\usepackage{amsmath,bm,amsfonts,amssymb,amsthm}

\usepackage{arcs}

\usepackage{commath}      
\usepackage{colortbl}  
\usepackage{color,xcolor,ucs}
\usepackage[font=small,labelfont=bf]{caption}

\usepackage{floatrow}
\usepackage{tabularx}
\usepackage{float}
\usepackage{hyperref}
\usepackage{xr-hyper}
\usepackage{wrapfig}

\usepackage{algorithm}
\usepackage{algpseudocode}
\usepackage{tikz}
\usetikzlibrary{tikzmark}
\usepackage{multirow}
\usepackage{listings}
\usepackage{lipsum}
\usepackage{textcomp} 

\usepackage[nocomma]{optidef}  
\usepackage{subcaption}
\usepackage{graphicx}
\usepackage{caption}
\usepackage{comment}
\usepackage[export]{adjustbox}

\theoremstyle{remark}
\newtheorem{definition}{Definition}

\newcommand{\B}{\mathbf}
\makeatletter
\newcommand{\clearsubcaptcounter}{\setcounter{sub\@captype}{0}}
\makeatother    

\definecolor{revisioncolor}{rgb}{0.1,0.1,1}

\newcommand{\tikzarc}[1]{%
\tikzmarknode{a}{#1}
\begin{tikzpicture}[overlay,remember picture]
\draw ([yshift=1pt]a.north west) to[bend left=20] ([yshift=1pt]a.north east);
\end{tikzpicture}%
}

\begin{document}

\begin{frontmatter}

\title{A Horizon-slicing Approach to Minimum Obstacle Displacement Planning for Robot Navigation}

\author[label1]{Antony Thomas}\ead{antony.thomas@iiit.ac.in}
\address[label1]{Robotics Research Center, IIIT Hyderabad, Hyderabad 500032, India.}

\author[label2]{Giulio Ferro}
\ead{giulio.ferro@unige.it}
\address[label2]{Department of Informatics, Bioengineering, Robotics, and Systems Engineering, University of Genoa, Via All'Opera Pia 13, 16145 Genoa, Italy. }

\author[label2]{Fulvio Mastrogiovanni}
\ead{fulvio.mastrogiovanni@unige.it}

\author[label2]{Michela Robba}
\ead{michela.robba@unige.it}

\author[label2]{Marco Baglietto}
\ead{marco.baglietto@unige.it}





\begin{abstract}
In this paper, we investigate the \textit{Minimum Obstacle Displacement Planning} problem from a robot motion planning perspective. The problem involves determining a feasible path to a goal location by displacing movable obstacles when no collision-free path initially exists. We show that this problem is computationally challenging and, in particular, NP-hard when obstacles are modeled as polygons in the plane.
Besides an exact formulation of the minimum obstacle displacement problem generalizing other problems in the literature, and the associated optimal solution, this paper proposes an approximate solution that is \textit{less intensive} from a computational standpoint, and differs from the optimal solution by a fraction of the optimal cost, being able to trade-off between path length and amount of obstacle displacements. 
\end{abstract}


\end{frontmatter}
\section{Introduction}
Classical approaches to robot motion planning aim at finding a feasible path from a start to a goal location \cite{Lav06}. 
The feasibility of a motion plan is scenario-specific. 
It involves different constraints, such as the robot's mechanical design, its actuators' performance, the robot's perception capabilities, uncertainties in state estimation and control signals, and the employed collision avoidance strategy. 
In principle, a complete path planner systematically explores the search space induced by the environment and the available planning information, examining all candidate paths consistent with its search strategy. If no feasible path exists because of the workspace configuration or due to constraints violation, the planner terminates reporting failure, or otherwise planners complete in probability can be used \cite{Mastrogiovannietal2009}.
However, in several application domains, it could be legitimate to evaluate the possibility of relaxing some of the constraints hindering a feasible path. 
For example, let us consider a robot arm tasked with reaching a target object in a cluttered tabletop setting. 
It may be the case that no collision-free path exists due to the specific arrangement of objects on the table.
However, objects themselves could be moved aside, therefore allowing the robot to reach the target object. 
Likewise, mobile robots may reach goal locations provided that certain obstacles could be rearranged or moved aside. 
This may involve, for example, pushing aside chairs or other furniture, as well as opening doors along the path.
The \textit{Minimum Constraint Displacement} (\textsf{MCD} in short) problem introduced by Hauser~\cite{hauser2013RSS} seeks to minimize the amount of \textit{space} by which one or more objects may have to be displaced in order to yield a feasible path. 
The approach by Hauser treats objects in so far as they can be modeled as \textit{constraints} for the motion planner. 
Therefore, in addition to relocating obstacles or modifying their configurations, the same framework demonstrates how a humanoid robot may modify its own body configuration to climb a ladder, for instance, by slightly increasing its hip angle limits, as well as finger and thumb strengths, that is, the mechanical constraints. 
All the constraints are represented in a graph, in particular as sub-graphs involving, for each obstacle, its candidate displacements (that is, a set of sampled displacements), which are associated with cost functions subject to a minimization procedure. 
One of the main contributions of \cite{hauser2013RSS} is the proof that the discrete version of \textsf{MCD} is NP-hard. 
Two approaches to solve the problem are discussed, namely, a theoretically exact method, which has exponential time complexity in the worst case, and an approximate (greedy) approach providing \textit{fast} solutions, but characterized by an unbounded approximation error.
\begin{figure}[t!]
\centering
\subfloat[A Pepper robot navigating amidst circular, movable objects.]{\includegraphics[scale=0.33]{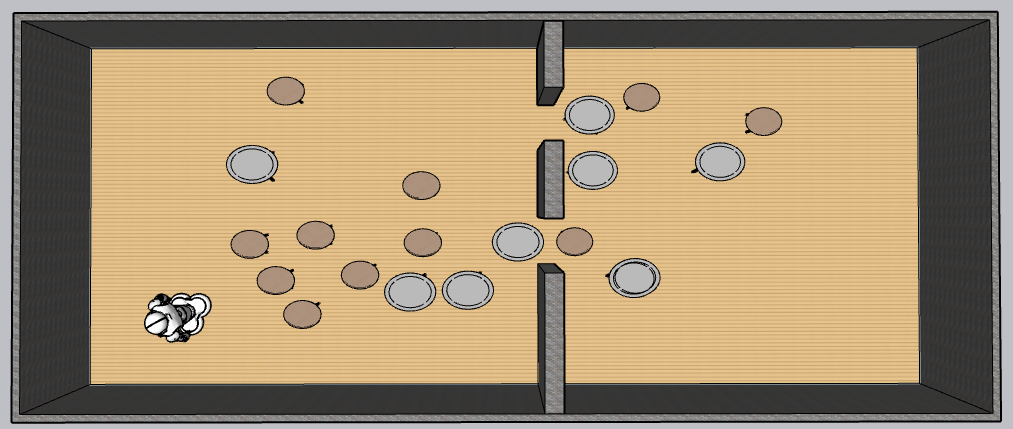}\label{fig:example_scenario}} \\
\subfloat[A mobile manipulator in a warehouse scenario.]{\includegraphics[trim=10 10 10 10,clip,scale=0.16]{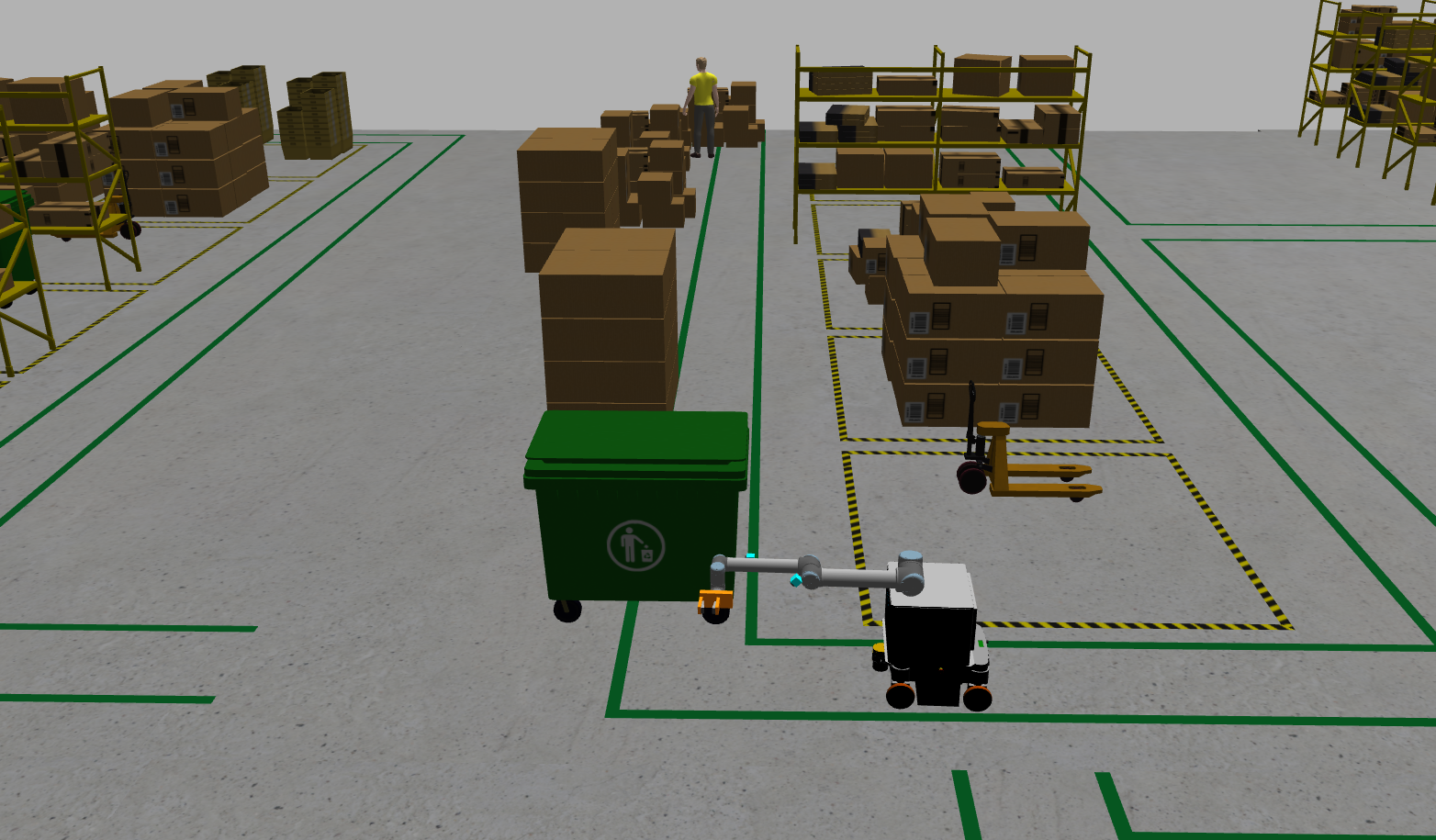}\label{fig:industrial}} 
\caption{
(a) A top-down view of a possible scenario where a simulated Pepper humanoid robot is required to navigate to the right-hand side of the environment, and must displace obstacles to that aim. 
(b) A mobile manipulator, restricted to moving between the green lines, encounters an obstacle blocking its path. 
To proceed, the manipulator must push the obstacle aside before continuing along the designated trajectory. 
The warehouse environment is simulated in Gazebo by modifying an off-the-shelf model available at \url{https://github.com/belal-ibrahim/dynamic_logistics_warehouse}.}
\end{figure}

\textit{This work focuses on scenarios in which a mobile robot must employ simple manipulation actions (such as, for example, \textit{pushing an object}) when the goal location is not reachable due to the peculiar configuration of the environment.}
A first example is shown in Fig. \ref{fig:example_scenario}, where a Pepper robot must displace objects (tables and chairs) to navigate through the passages between two areas.
A second example, illustrated in Fig.~\ref{fig:industrial}, shows a mobile manipulator operating in an warehouse environment, which has to move obstacles out of its path, which is constrained by the green lines.
Specifically, this paper is about how a robot can reach an assigned goal location by displacing obstacles while minimizing the total amount of obstacle displacements, and being able to trade off the number of obstacles to be displaced and the corresponding amount.
In this work, we formulate a \textit{Minimum Obstacle Displacement} (\textsf{MOD}) problem that minimizes a weighted cost of 
(a) total obstacle displacement magnitudes, and
(b) robot path length, as the robot navigates to the goal location. 

\begin{figure*}[t!]
\subfloat[]{\includegraphics[scale=0.25]{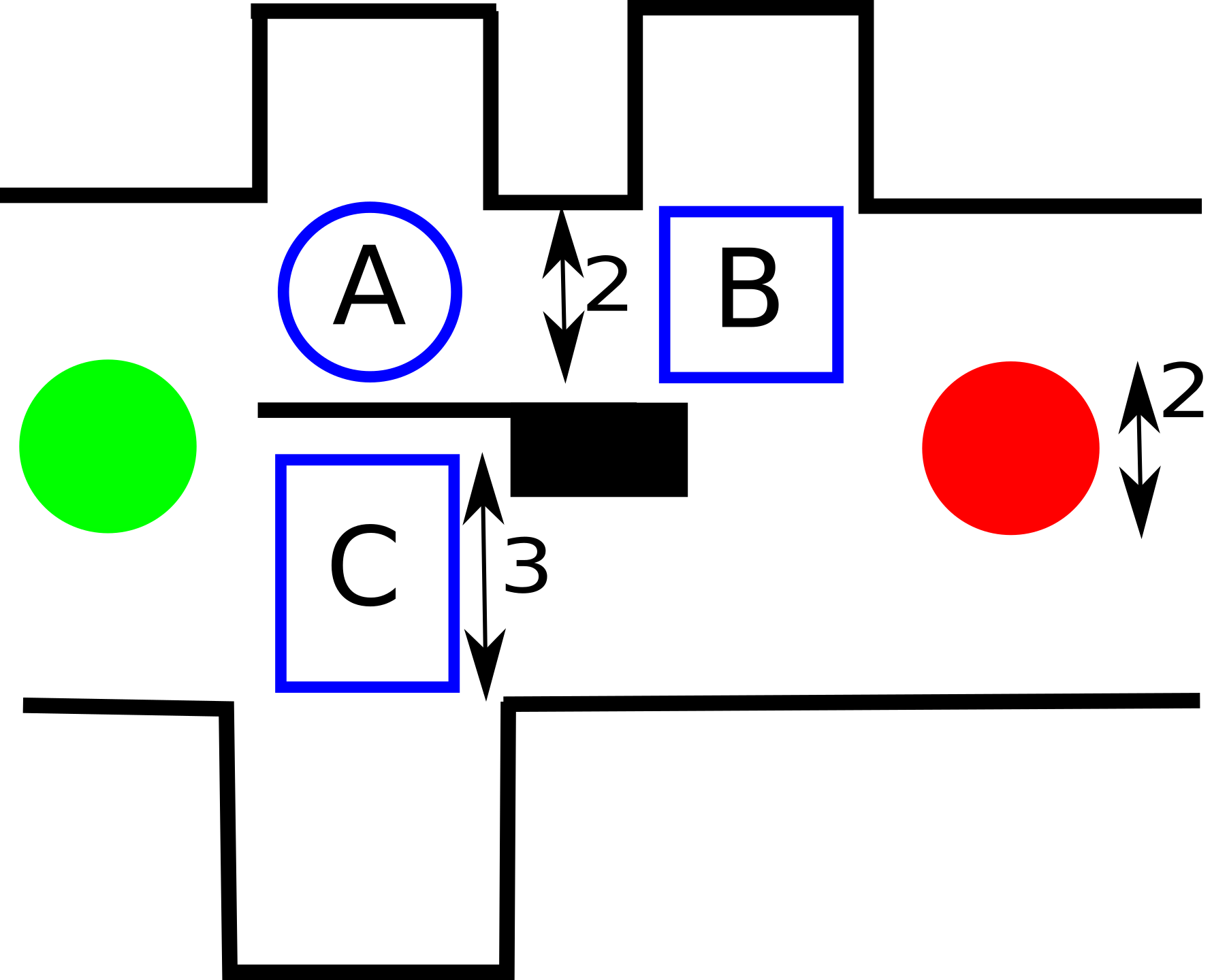}\label{fig:ex1}} \hspace{12pt} 
\subfloat[]{\includegraphics[scale=0.248]{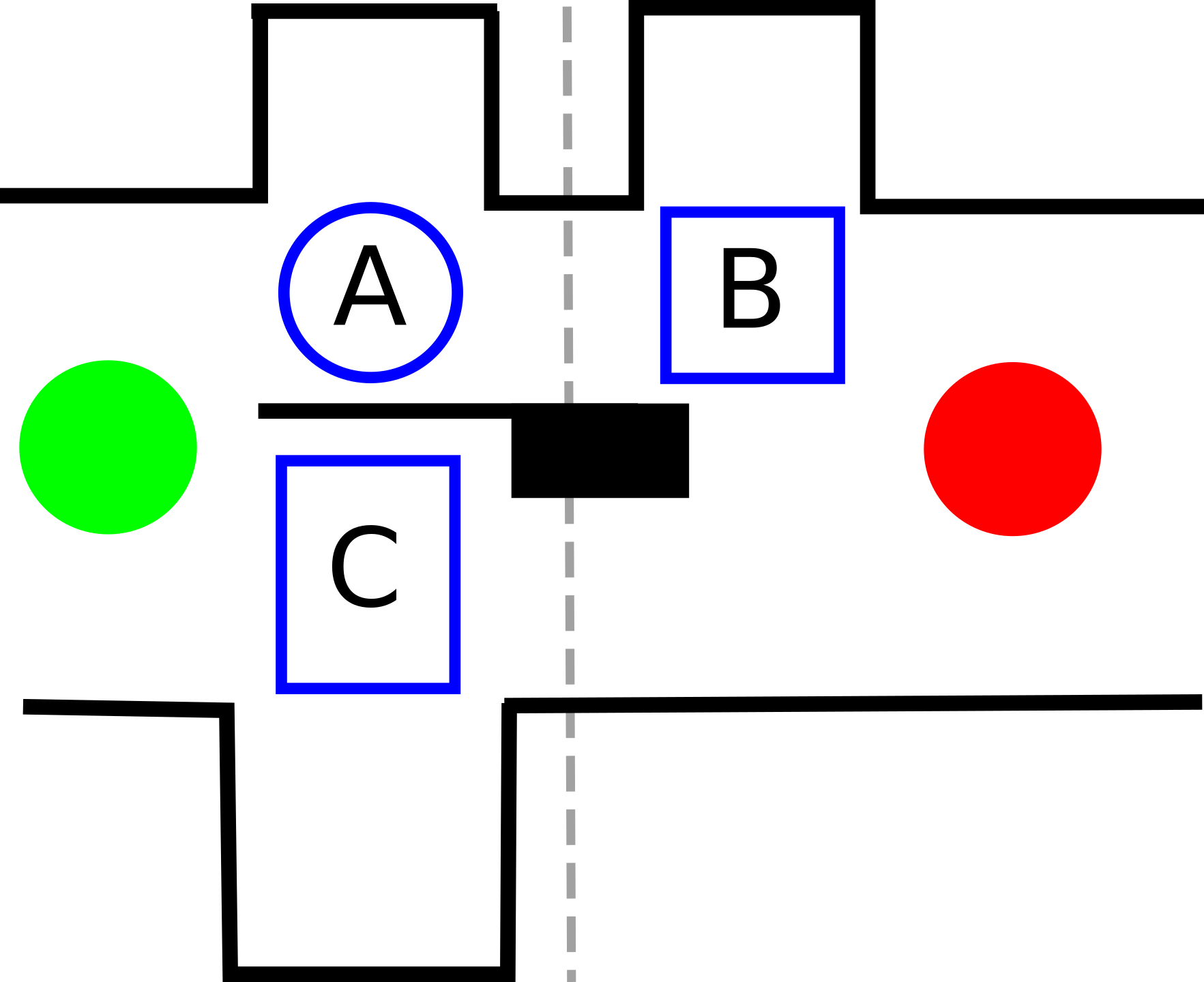}\label{fig:ex2}} \hspace{12pt} 
\subfloat[]{\includegraphics[scale=0.246]{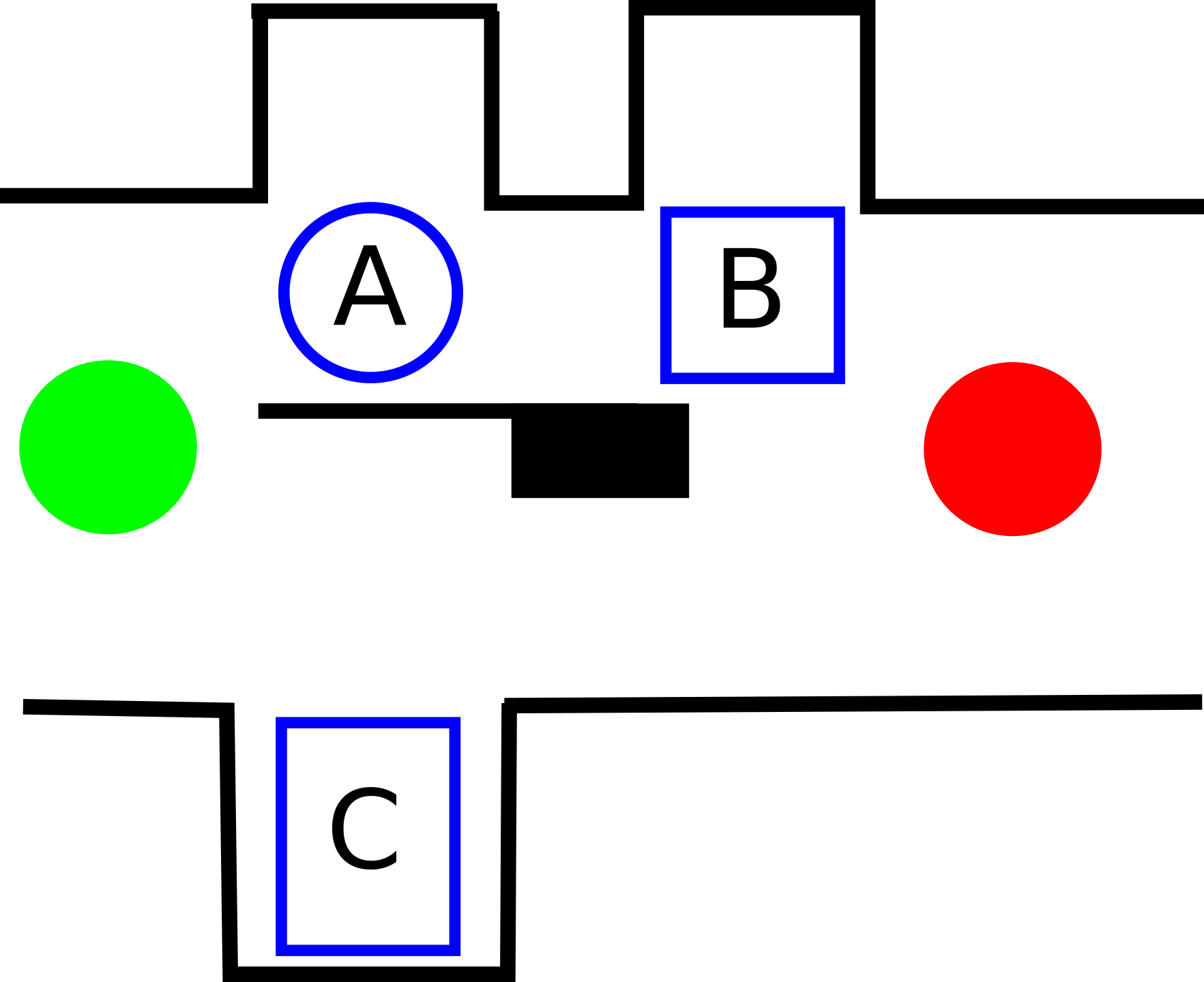}\label{fig:ex3}}
\caption{
An example of the \textsf{MOD} problem not satisfying the optimal substructure property.
(a) The robot must move from left (start location in green) to right (goal location in red) in the presence of three movable obstacles, referred to as $A$, $B$, and $C$.
(b) A solution to the \textsf{MOD} problem splitting it into two sub-problems (left and right of the dashed line).
This leads to a feasible path by moving obstacles $A$ and $B$ with a total displacement magnitude of 4 units.
(c) The optimal solution is obtained by displacing obstacle $C$ by 3 units.
}
\label{fig:ex}
\end{figure*}

This problem is shown to be NP-hard in Section~\ref{sec:nphard}, and therefore it is appropriate to look into approximate methods. 
We introduce one such method, which we refer to as \textit{Horizon Slicing for \textsf{MOD}} (\textsf{HS4MOD} in short), by dividing a \textsf{MOD} problem into (a sequence of) sub-problems. 
Solving \textsf{MOD} by separating it into sub-problems is an approximation as \textsf{MOD} does not exhibit the optimal substructure property \cite{cormen2009book}, that is, optimal solutions to sub-problems -- when combined -- do not necessarily form an optimal solution to the overall problem.
This can be easily illustrated by an example.
Let us consider the simple scenario shown in Fig.~\ref{fig:ex}. 
The robot is required to move from a start (in green) to a goal (in red) location in the presence of movable obstacles ($A$, $B$, and $C$ in the Figure). 
The robot can reach the goal location either by displacing obstacles $A$ and $B$ (that is, following the \textit{upper} path) or by displacing obstacle $C$ (the \textit{lower} path). 
We note here that the start and goal locations are chosen for the sake of this example such that the robot path length for both the upper and the lower paths are equal. 
In Fig.~\ref{fig:ex2}, the problem is split into two sub-problems. 
In the first sub-problem, the robot can either displace obstacle $A$ by 2 space units or obstacle $C$ by 3 units. 
Since minimum displacement is solicited (and path lengths are equal as argued above), the robot chooses to move obstacle $A$ by 2 units. 
On the one hand, if the robot moved through the upper path by displacing $A$, in the next sub-problem it would have to move obstacle $B$ by 2 units to reach the goal location, and therefore the total displacement magnitude would amount to $4$ space units. 
On the other hand, displacing $C$ by 3 units would generate a feasible path for the robot as shown in Fig.~\ref{fig:ex3}, and it would clearly be the optimal solution. 
This shows with a counterexample that the optimal solution to \textsf{MOD} cannot be obtained simply by juxtaposing optimal solutions to its sub-problems. 

\textit{It is important to note that, in this work, we are not concerned about \textit{how} objects can be displaced, both in terms of task or motion planning \cite{Capitanellietal2018}.
Neither are we concerned about the weight of an obstacle, or whether it would be necessary to push or pull it, in case of displacement. 
What we are concerned with is the amount by which obstacles must be displaced to yield a feasible path and assume that the computed displacements can be achieved by appropriate robot-obstacle physical interactions.} 
From an algorithmic perspective, this assumption allows us to 
(1) assume a \textit{planar}, 2D projection of the 3D environment, and 
(2) compute only the locations on such a plane where the obstacles are to be displaced. 
The same assumption allows us to consider events (both robot motions as well as object displacements) as \textit{collapsed} to discrete time instants.
If the robot took $T$ discrete time instants to reach the goal location, with $0 \leq k\leq T$, we could associate both control actions and obstacle displacements (and zero displacements if an obstacle is not affected) to each time instant $k$ in the sequence.
In this work, we also disregard obstacle interactions, that is, the displaced obstacles are allowed to overlap with other obstacles, and therefore obstacle-obstacle collision avoidance is not considered. 
\newpage


The main contributions of the paper are:
\begin{itemize}
\item
We prove that \textsf{MOD} is NP-hard even if all the obsatcles are line segments by a reduction from the \textit{Partition Problem}, which is NP-hard~\cite{michael1979book}. We further show that \textsf{MOD} remains NP-hard when the obstacles are restricted to axis-aligned rectangles. In addition, we establish the NP-hardness of \textsf{MOD} for circular obstacles. Together, these results substantially strengthen the NP-hardness result for the discrete \textsf{MCD} problem reported in~\cite{hauser2013RSS}.    
\item
We present an exact approach to \textsf{MOD} by formulating the problem as a \textit{Mixed Integer Quadratic Programming} problem. 
This approach returns exact solutions able to weight between path length and displacement amount but has worst-case exponential time. 
\item
We introduce \textsf{HS4MOD}, an approximation method for \textsf{MOD} that divides MOD into $m$ sub-problems and obtains a solution from a combination of optimal -- yet local -- solutions to these sub-problems. 
\textsf{HS4MOD} is shown to exhibit significant computational improvement over the exact approach even when the number of sub-problems is limited to two or three. 
For all the results presented in Section~\ref{sec:results}, we also compute the \textit{optimality gap}, that is, the relative distance between the approximate and the optimal solution, with the aim of evaluating the quality of the solutions provided by \textsf{HS4MOD}.
\end{itemize}

The paper is organized as follows. 
We discuss related literature in Section~\ref{sec:related_work}. 
In Section~\ref{sec:problem_definition}, we provide a formal definition of the problem. 
We show in Section~\ref{sec:nphard} that \textsf{MOD} is NP-hard for polygonal obstacles in a plane. 
Section~\ref{sec:approach} presents an exact mixed integer quadratic program optimization for \textsf{MOD}, and in Section~\ref{sec:horizion_slicing} we introduce an approximate strategy for its solution. 
\textsf{HS4MOD} obtains a solution to \textsf{MOD} that combines optimal yet local solutions to its sub-problems. 
In Section~\ref{sec:results}, we evaluate \textsf{HS4MOD} under varying domains, distinct robot models, and different numbers of obstacles. 
The discussion in Section~\ref{sec:discussion} concludes the paper. 
\section{Background and Problem Definition}
\subsection{Related Work}
\label{sec:related_work}
The problem we address in this paper shares some resemblance with a number of apparently similar problems discussed in the literature.
Herein, we describe the related classes of problems highlighting similarities and differences, and we argue that \textsf{MOD} deserves a principled treatment on its own. 

The \textsf{MCD} planner presented and discussed in~\cite{hauser2013RSS} proceeds by 
(1) building a road map of robot configurations, and 
(2) sampling random obstacle displacements. 
The quality of the solution is improved by iteratively expanding the road map and the displacement samples.
The method is shown to asymptotically approach the true optimum as the number of samples (both robot configurations and displacements) grow in size. 
In contrast, we formulate an exact approach, incorporting the robot kinematics and solve it find the optimal solution. 
Furthermore, we present an approximate method and compute its \textit{deviation} from the computed optimal solution. 
\textsf{MCD} focuses on pure motion planning in the sense that it ignores robot dynamics and other differential constraints, and finds the translations and rotations required to move the robot. 
In our case, we plan a trajectory taking into account robot dynamics and other constraints such as control limits and collision constraints, to find the sequence of control actions to move the robot to the goal location. 

The work in~\cite{thomas2022IAS} examines minimum displacement planning for movable obstacles through a two-stage process. 
The first stage proceeds by finding a path through a set of movable obstacles while minimizing robot-obstacle overlaps. 
In the second stage, the overlapping obstacles are displaced iteratively by a distance factor until no overlap is found. 
However, the synthesized solutions are not necessarily optimal, and no guarantees or discussion regarding their quality are provided. Our approach computes the exact optimum and, for the proposed approximation algorithm, quantifies its solution quality with respect to the exact method across all experimental evaluations.

In a general sense, the problem we address is closely related to the \textit{Navigation Among Movable Obstacles} (\textsf{NAMO}) class of problems~\cite{stilman2005IJHR, nieuwenhuisen2008WAFR, stilman2008IJRR, van2009WAFR,weeda2025ICRA}. 
Most approaches for \textsf{NAMO} focus on connecting pairwise separated configuration regions by displacing objects as the robot proceeds to the goal location, and seek to find a motion plan that displaces the least number of obstacles while minimizing the \textit{work} done to do it. Learning-based and language model-based approaches have also been explored for \textsf{NAMO}~\cite{Zhang2025RAL,yang2025IROS}. However, these methods do not explicitly account for minimizing the displacement of manipulated objects during task execution.

\textsf{NAMO} problems are shown to be NP-hard if the final obstacle locations are unspecified, and PSPACE-hard when the locations are specified instead~\cite{wilfong1991AMAI}. 
Stilman and Kuffner~\cite{stilman2005IJHR} address linear problems of degree 1 (\textsf{$LP_1$}), that is, two disjoint regions in the configuration space are linked by displacing a \textit{single} obstacle.
Subsequent studies~\cite{nieuwenhuisen2008WAFR, stilman2008IJRR, van2009WAFR} extend this concept to \textsf{$LP_k$}, where \textit{at most} $k$ obstacles are displaced to connect two free configuration regions. 
As such, \textsf{NAMO} relies on \textit{keyhole} solutions, where a keyhole is the sub-problem of moving one (\textsf{$LP_1$}) or more objects (\textsf{$LP_k$}) to connect two regions of the free configuration space.
Unlike these approaches, our method does not partition the configuration space into disconnected regions, nor does it impose restrictions on the number of obstacles that may be displaced.
Though \textsf{NAMO} incorporates manipulation actions by selecting from obstacle contact points, the obstacles may be displaced unnecessarily far. 
Exact \textsf{MOD} tackles both displacement and path optimality concurrently.

\textit{Disconnection Proving in Motion Planning}~\cite{basch2001ICRA, zhang2008WAFR, li2021RSS, li2023RAL} poses related challenges since it tries to find constraints preventing a robot to move from start to goal locations.
Our approach does not need to consider explicitly hindering constraints, but adopts a horizon-slicing approximation, which requires connecting all the sub-optimal solutions.

The work in~\cite{thomas2026RAS} considers a unified approach for solving constraint displacement problems (MOD, MCR, NAMO) by varying the objective function of a common optimization framework. However, the solutions obtained are suboptimal and lack guarantees with respect to optimality.

\textit{Manipulation Among Clutter} or \textit{Rearrangement Planning}~\cite{stilman2007ICRA, dogar2011RSS, krontiris2015RSS, karami2021AIIA} is another related class of problems wherein objects may need to be displaced or rearranged to perform a given task. 
However, in these problems, the final location of the objects is either specified, or it could occur that displaced objects may be placed anywhere in the (free) workspace. 
Similarly, approaches for integrated \textit{Task and Motion Planning} (\textsf{TaMP})~\cite{kaelbling2013IJRR, srivastava2014ICRA, dantam2016RSS, garrett2018IJRR, thomas2021RAS, karami2025RAS} also bear resemblance to the problem discussed in this paper since task execution may require dealing with, at the motion planning level, constraints not considered at the task planning level.
It is worth noting that the problem we consider here could be one of the key elements in such classes of problems.

The \textit{Minimum Constraint Removal} problem (\textsf{MCR})~\cite{hauser2014IJRR} finds the minimum number of movable obstacles that should be removed to yield a feasible path, but does not determine where the obstacles should be moved for creating such a path. 
Different algorithms for \textsf{MCR} exist in the literature~\cite{castro2013CDC, xu2016ASC, krontiris2017AR, xu2020ICMLC, thomas2023IAS}, and the problem is proven to be NP-hard for convex polygonal obstacles~\cite{hauser2014IJRR}. 
Our formulation supersedes this class of approaches in that it determines also each object's displacement.
\subsection{The Minimum Obstacle Displacement Problem}
\label{sec:problem_definition}
Let us consider a generic robot, and let $\B{x}_k$ denote the robot state at any time instant $k$. 
Let us consider, for the sake of the discussion, a scenario where a mobile robot should reach a goal location, and therefore $\B{x}_k$ may denote its pose in terms of position and orientation. 
We assume a standard motion model for the robot given by
\begin{equation}
\B{x}_{k+1} = f(\B{x}_k,\B{u}_k),
\label{eq:motion_model}
\end{equation}
\noindent where $\B{u}_k$ is the control action applied at time instant $k$, where $k \in \{0, \ldots, T-1\}$.
Furthermore, let $\B{x}^s$ and $\B{x}^g$ denote the start and goal configurations, respectively.
We shall denote the robot position at time $k$ by $\B{p}_k \subseteq \B{x}_k$, and the sequence of all the positions from $0$ to $T$ as $\B{p}$.  
Let 
\[
\mathcal{O} = \{o^i \mid 1 \leq i \leq |\mathcal{O}|\}
\]
represent the set of obstacles in the environment, where $\mathcal{M} \subseteq \mathcal{O}$ is the set of movable obstacles.
We will use $\B{o}^i_k$ to denote the obstacle state associated with the obstacle $o^i$ at time $k$. 
As it has been argued in the previous Section, at each time instant $k$, obstacles in $\mathcal{M}$ are associated with a displacement set
\[\mathcal{D}_k = \{d^i_k \in \mathbb{R}^n| \ 1 \leq i \leq \mathcal{M}\}
\]
representing the corresponding obstacle displacements, where $d^i_k = 0$ in case the object $o^i \in \mathcal{M}$ is not displaced at time $k$. 
Both $\B{x}_k$ and all obstacle locations $\B{o}^i_k$ are, in principle, elements of the configuration space.
It may be useful to define a function $O$ that, when applied to the robot pose at time instant $k$, maps to the set of points corresponding to the robot body, namely $O(\B{x}_k)$, and when applied to an obstacle pose $\B{o}^i_k$, maps to the set of points corresponding to the obstacle itself, namely $O(\B{o}^i_k)$. 
Before formally defining the problem addressed in this paper, we will rephrase the conceptual perimeter of this work:
\begin{itemize}
\item We do not focus on how objects are displaced;
\item We do not take into account the weight of the obstacles, and instead we focus solely on their 2D projections;
\item This work does not take into consideration collision avoidance among obstacles.
\item We assume that a map of the environment is available a priori. Consequently, the proposed method is an offline planning approach that computes the required displacement for each obstacle before execution.
\end{itemize}
Now, we proceed to formally define the problem that is the focus of this paper.
\begin{definition}
The \textit{Minimum Obstacle Displacement} (\textsf{MOD}) planning problem finds a sequence of control actions $\B{u}_0, \ldots, \B{u}_{T-1}$ which make the robot navigate from $\B{x}^s$ to $\B{x}^g$ with obstacle displacement magnitudes $D^1, \ldots, D^{\mathcal{M}}$ and is defined as

\begin{mini!}|s|[2]
{ }{w^x c(\B{p}) + w^d \sum_{i=1}^{\mathcal{M}} c^i(D^i),}{\label{eq:mdmp}}{\label{eq:cost}}
\addConstraint{c(\B{p}) = \sum\limits_{k=0}^{T-1} \norm{\B{p}_{k+1}- \B{p}_{k}}^2 + w^g\norm{\B{p}_T - \B{p}^g}^2},{\label{eq:cost_pos}}
\addConstraint{O(\B{x}_k) \cap \bigcup\limits_{i=1}^{|\mathcal{O}|} O(\B{o}_k^i) = \{\emptyset\}, \ \forall k \in [1,T]},{\label{eq:coll_avoid}}
\addConstraint{c(D^i) = \sum_{k=0}^T d^i_k},{\label{eq:disp_def}}
\addConstraint{\B{x}_1 = \B{x}^s}.{\label{eq:init}} 
\end{mini!}
\label{def:mdmp}
\end{definition}

The objective function \eqref{eq:cost} aims to minimize two components:
the first component $c(\B{p})$ is composed of two terms as given in \eqref{eq:cost_pos}, which represent the length of the robot path while keeping track of the goal $(\B{p}^g)$ as terminal position, weighted by $w^g$;
in the second component, $c^i(D^i)$ are functions of displacement magnitudes;
$w^x$, $w^d$ are the respective weights for the two components.
By tuning the weights in an appropriate way, it is possible to trade off path length and the magnitude of displacements, thereby allowing for shortest paths or more disruptive obstacle displacements.
Condition \eqref{eq:coll_avoid} guarantees that the robot does not intersect obstacles at any time.
We note here that for obstacles with fixed position, $O(\B{o}_k^i)$ remains constant throughout.
Condition \eqref{eq:disp_def} defines the total displacement of obstacles, and the initial condition is established in \eqref{eq:init}. 

It can be easily verified that when all displacements $D^i$ are equal to $0$, \textsf{MOD} solves the classical motion planning problem (see Section~\ref{sec:results} for an example). 
Furthermore, \textsf{MOD} solves the shortest path problem if $c(\B{p})$ measures the path length, and $w^d = w^x L^s$, where $L^s$ is equal to the length of the shortest feasible path. 
\textsf{MOD} can also be employed to determine the minimum number of obstacles to be removed to find a feasible path, which is basically the \textsf{MCR} problem with constraints being obstacles. 
This is achieved by assigning $w^x=0$ and limiting $d^i = \{0,1\}$, that is, $d^i=0$ for the obstacles being present and $d^i=1$ for the obstacles that are removed (an example is given in Section~\ref{sec:results}). 
It is noteworthy that by modifying $c^i(D^i)$ to incorporate the work done by the robot, \textsf{MOD} solves a variant of the \textsf{NAMO} problem~\cite{stilman2005IJHR}. 
Similarly, by relaxing $w^d$, \textsf{MOD} may be employed to solve variants of the re-arrangement planning problem. 

Therefore, it can be deduced that \textsf{MOD} can generalize different classes of motion planning problems, and it deserves a treatment on its own.
\section{Complexity of the \textsf{MOD} Problem}
\label{sec:nphard}
\begin{figure}[t!]
\centering
\includegraphics[scale=.9]{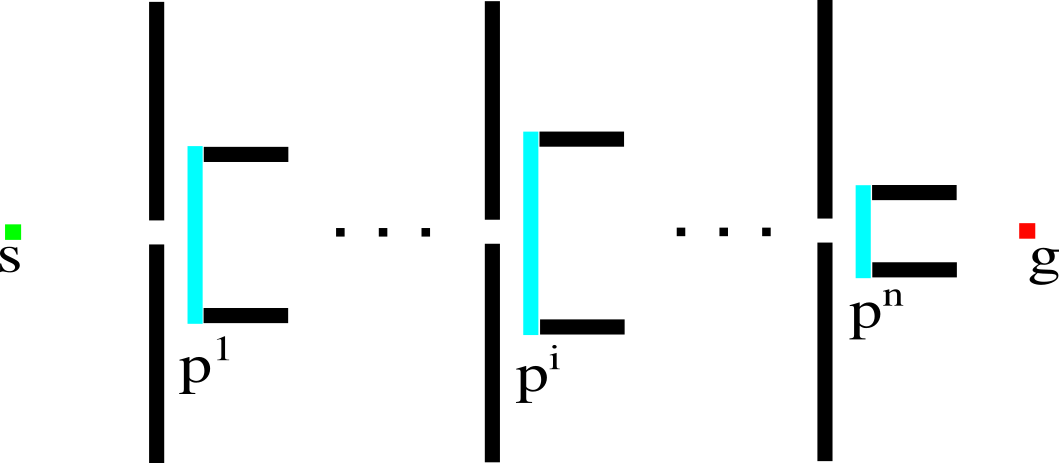}
\caption{
Construction by reduction from the \textit{Partition} problem: movable line segments (in cyan) may be displaced only in the vertical direction.
The segments in black are static obstacles. 
To pass along the $x$-axis (which connects start and goal locations) at location $i$, the obstacle at $i$ has to be displaced at least by $p^i/2$. 
Otherwise, the robot has to move vertically for a distance of $p^i/2$ to reach one of the ends of obstacle at $i$, and then vertically by the same amount to reach again the $x$-axis.
}
\label{fig:vertical_line}
\end{figure}

In this Section, we show that \textsf{MOD} is NP-hard for polygonal obstacles in the plane by considering two special cases, namely obstacles 
(1) as line segments, and
(2) as axis-aligned rectangles with a unique displacement value.
Furthermore, we also prove that \textsf{MOD} is NP-hard for circular-shaped obstacles.

\subsection{\textsf{MOD} for Line Segments is NP-hard}
We consider the class of problems in which obstacles are disjoint (vertical) line segments and do not touch each other. 
Therefore, there exists a feasible path (in the classical motion planning sense).
However, it might be useful to displace some obstacles to obtain a shorter path. 
We consider the \textit{decision} variant of the problem, that is, if there is a feasible path to a goal location given a maximum path length of $L^m$ and a maximum displacement \textit{budget} $D^m$. 
This problem is NP-hard by reduction from the \textit{Partition} problem, which is NP-complete~\cite{michael1979book}. 

We now prove the hardness. 
Given a set
\[
P = \left\{\frac{p^1}{2}, \ldots, \frac{p^n}{2}\right\}
\]
of $n$ positive integers, the Partition problem checks whether two sets $P1$ and $P2 = P - P1$ exist such that $S(P1) = S(P2) = S(P)/2$, where $S(\cdot)$ computes the sum of the elements in the set. 
With reference to Fig.~\ref{fig:vertical_line}, we place the goal location $\B{x}^g$ at a distance of $n+1$ units along the $x$-axis from the start location $\B{x}^s$. 
We have $n$ line segments, whereas each line segment $i$ is of length $p^i$, and is placed at a distance of $i$ units from the start location. 
We set
\[D^m = \frac{S(P)}{2}\]
and
\[L^m = 2\left(\frac{S(P)}{2}\right) + (n + 1).\] 
Furthermore, we construct two vertical and two horizontal line segments as static obstacles close to each of the $n$ movable obstacles. 
The fixed vertical line segments have a length of $2L^m+1$ and are separated from each other by a distance $\delta = 1/4n$, which is assumed to be small enough for the robot to pass through. 
They are constructed to lie to the left of each movable obstacle at a distance of $\delta$. 
The horizontal obstacle segments have a length of $(1-2\delta)$, and are placed touching the right end of each obstacle at $i$ and lie symmetrically at a distance of $p^i/2$ from the $x$-axis. 
As a consequence of this construction, the set $P$ can be partitioned into two sets $P1$ and $P2 = P - P1$ such that $S(P1) = S(P2)$ only if there is a path from $\B{x}^s$ to $\B{x}^g$ of length at most $L^m$, and displacement cost of at most $D^m$. 
Thus, \textsf{MOD} for (vertical) line segments in a plane is NP-hard. 

\begin{figure}[t!]
\centering
\includegraphics[scale=0.18]{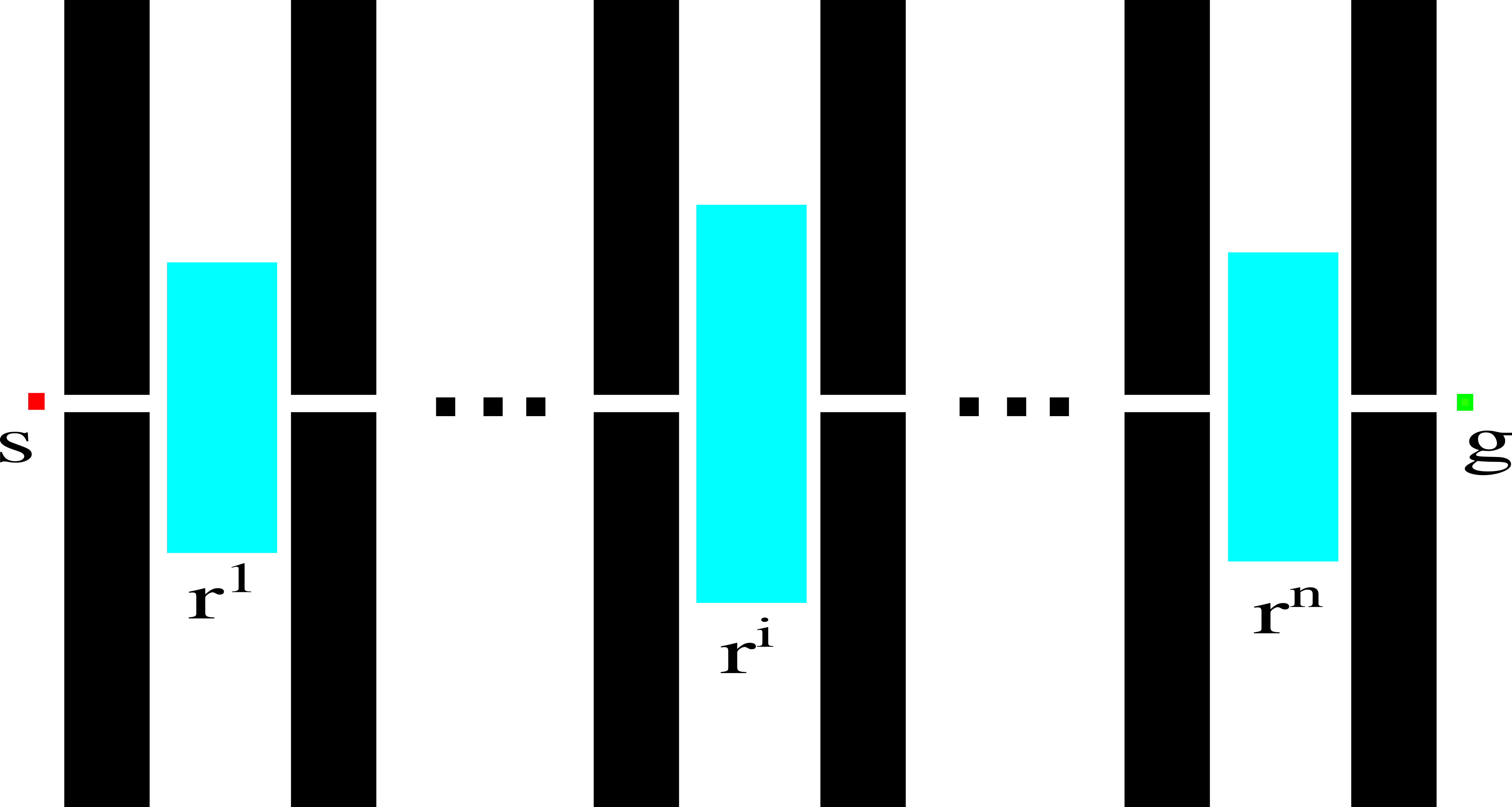}
\caption{
Movable rectangles are shown in cyan and can be displaced only by an amount $r^i$. 
Drawn in black are the obstacles constructed to achieve the reduction of the Partition problem.}
\label{fig:axis_rectangle}
\end{figure}

\subsection{\textsf{MOD} for Axis-aligned Rectangles with Constant Displacement Cost is NP-hard}
In this configuration, each axis-aligned rectangle $i$ is allowed a single displacement\footnote{We note here that while the result presented is of theoretical interest, in the real world the single-displacement value considered here can be mapped to obstacles moving to a specified final location -- for example, the transportation of pallets in logistics \cite{Mohamedetal2019}.} of magnitude $r^i$. 
We consider again the decision version of the problem, that is, given a set of $n$ movable obstacles in the form of axis-aligned rectangles, we want to ascertain if there is a feasible path from $\B{x}^s$ to $\B{x}^g$ such that the maximum path length and displacement do not exceed $L^m$ and $D^m$, respectively. 
The NP-hardness can be proved by reduction from the Partition problem.

In order to prove the hardness, the scenario as shown in Fig.~\ref{fig:axis_rectangle} is constructed. 
We place $\B{x}^s$ and $\B{x}^g$ at a distance of $2n+1$ units along the $x$-axis. 
Each left end of a movable rectangle $i$ is placed at $2i-1$ units from $\B{x}^s$. 
The rectangles have a length of $r^i$ and a width of $1$ unit, and are allowed to be displaced of $r^i$ units in the vertical direction. 
At a distance of $2(i-1)+\delta$ units from $\B{x}^s$ we place two rectangular obstacles, vertically separated by a distance $\delta = 1/4n$, which is small enough for the robot to pass through. 
Each of these rectangles has a length of $2L^m+1$ and a width of $1-2\delta$. 
Now, let us consider a set $R = \{r^1, \ldots, r^n\}$ and let
\[D^m = \frac{S(R)}{2}\]
and
\[L^m = \frac{S(R)}{2} + (2n + 1).\] 
The set $R$ can be partitioned into two sets $R1$ and $R2 = R - R1$ such that $S(R1) = S(R2)$ only if there is a path from $\B{x}^s$ to $\B{x}^g$ of length at most $L^m$, and displacement cost of at most $D^m$. 
Thus, the problem of deciding if there is a path with length at most $L^m$ and displacement cost at most $D^m$ is NP-hard. 

\begin{figure}[t!]
\centering
\includegraphics[scale=1.3]{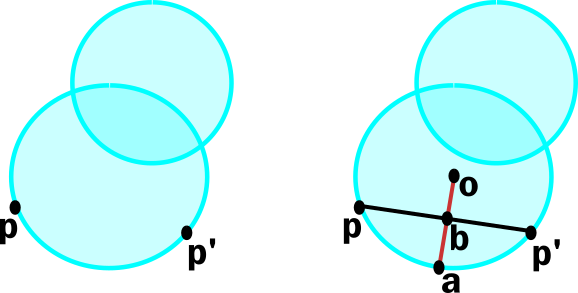}
\caption{Construction for NP-hardness proof of MOD with circular obstacles. See text for further details.}
\label{fig:circle_NP}
\end{figure}

\subsection{\textsf{MOD} for Circular Obstacles is NP-hard}
Let us consider a scenario where circular obstacles are located on the plane.
When the robot encounters an obstacle at point $p$ (refer to Fig.~\ref{fig:circle_NP}), it navigates along the arc $\tikzarc{pp'}$ if the obstacle remains stationary, with $p'$ denoting the point where the robot departs from the obstacle's boundary.
If the obstacle is displaced, the shortest path $\overline{pp'}$ for the robot is achieved by displacing the obstacle at least by an amount $d = \overline{ab}$, with $\overline{oa} \perp \overline{pp'}$, and $o$ being the center of the obstacle.
Thus, for a given displacement $d$, the shortest path $\overline{pp'}$ may easily be computed as
\[\overline{pp'}= 2\sqrt{r^2 - (r-d)^2},\]
with $r$ denoting the radius of the obstacle.
If the obstacle is displaced by an amount $d$, the robot path length decreases by an amount $\tikzarc{pp'}-\overline{pp'}$.
For the discussion that follows, the decrease in the path length for a given $d$ will be referred to as the \textit{value} associated with $d$. 


We now establish the NP-hardness of the \textsf{MOD} problem by a polynomial-time reduction from the \textit{Multiple-Choice Knapsack Problem} (\textsf{MCKP}), which is NP-hard~\cite{kellerer2004multiple}. We consider the budget-constrained discrete special case of \textsf{MOD}, in which a finite set of admissible displacements is specified for each movable obstacle and the total obstacle displacement is bounded by a given budget $D^m$.

In an instance of \textsf{MCKP}, there are $n$ mutually disjoint classes ${M^1,\ldots,M^n}$ of items and a knapsack of capacity $c$. Each item $j\in M^i$ has a weight $W^{ij}$ and a profit $P^{ij}$. The objective is to select exactly one item from each class such that $\sum_{i=1}^{n}W^{ij_i}\leq c $, while maximizing $\sum_{i=1}^{n}P^{ij_i}$. 

We construct an instance of the budget-constrained discrete \textsf{MOD} problem with $n$ movable circular obstacles, where obstacle $i$ corresponds to class $M^i$. For every item $j\in M^i$, we construct one admissible displacement $d^{ij}$ and its corresponding path-length reduction $v^{ij}$. The construction is chosen such that $
d^{ij}=W^{ij}, v^{ij}=P^{ij}
$. The latter correspondence is realizable from the circular-obstacle geometry since the path-length reduction produced by a displacement $d$ of a circular obstacle of radius $r_i$ is $
V_i(d)=2\sqrt{r_i^2-(r_i-d)^2}
      =2\sqrt{2r_i d-d^2}.
$

For a prescribed value $P^{ij}$, an appropriate radius and displacement can be selected so that $V_i(d^{ij})=P^{ij}$. In particular, the geometric parameters can be constructed using the inverse relation $
r_i=\frac{(P^{ij})^2+4(d^{ij})^2}{8d^{ij}}
$, with the construction chosen consistently for the admissible displacement set of obstacle $i$. We then set the maximum total displacement to $D^m=c$. Consequently, selecting one displacement for each obstacle gives $
\sum_{i=1}^{n}d^{ij_i}\leq D^m
\quad\Longleftrightarrow\quad
\sum_{i=1}^{n}W^{ij_i}\leq c
$, while maximizing the total path-length reduction gives $
\max\sum_{i=1}^{n}v^{ij_i}
\quad\Longleftrightarrow\quad
\max\sum_{i=1}^{n}P^{ij_i}
$.

Thus, every feasible selection of one item from each MCKP class corresponds to a feasible selection of one displacement for each obstacle, with identical objective value, and vice versa. Hence, an optimal solution of the constructed \textsf{MOD} instance yields an optimal solution of the original \textsf{MCKP} instance. The construction requires only a polynomial number of obstacle parameters and displacement choices, and therefore can be performed in polynomial time. Hence, the budget-constrained discrete special case of \textsf{MOD}, and consequently \textsf{MOD} itself, is NP-hard.

\\
\section{An Exact Approach to Solve \textsf{MOD}}
\label{sec:approach}
In this Section, we present an exact approach to solve \textsf{MOD} that achieves an optimal, global solution, albeit computationally expensive as the set of obstacles $\mathcal{O}$ and the robot workspace grow. 

\subsection{The Optimization Problem}

Using the models related to robots and obstacles, we formulate \textsf{MOD} as an optimization problem that finds a feasible path for the robot to the goal location while minimizing a trade off between path length and obstacle displacements. 
While doing so, control inputs and state variables must lie within their respective feasible sets, and the robot should not collide with obstacles. 
Starting from the general formulation of the \textsf{MOD} problem, the overall optimization problem can be thus formalized as follows.

\begin{mini!}|s|[2]
{ }{J = w^xc(\B{p}) + w^d \sum_{i=1}^{\mathcal{M}} c^i(D^i),}{\label{eq:optimization_problem}}{\label{eq:cost_fn0}}
\addConstraint{\B{x}_{k+1}=f( \B{x}_k,\B{u}_k ), }{\label{eq:cost_fn1}} 
\addConstraint{\B{o}_{k+1}=g( \B{o}_k,\B{d}_k ), }{\label{eq:cost_fn2}}
\addConstraint{\underline{\B{x}}  \le \B{x}_{k} \le \overline{\B{x}}, }{\label{eq:cost_fn3}}
\addConstraint{\underline{\B{u}}  \le \B{u}_{k} \le \overline{\B{u}}, }{\label{eq:cost_fn4}}
\addConstraint{\underline{\B{o}}  \le \B{o}_{k} \le \overline{\B{o}}, }{\label{eq:cost_fn5}}
\addConstraint{\underline{\B{d}}  \le \B{d}_{k} \le \overline{\B{d}}, }{}{\label{eq:cost_fn6}}
\addConstraint{O(\B{x}_k) \cap \bigcup\limits_{i=1}^{|\mathcal{O}|} O(\B{o}_k^i) = \{\emptyset\}, \ \forall k \in [0,T],  }{}{\label{eq:cost_fn7}}
\addConstraint{\forall k \in \{1,\ldots,T\},} \nonumber
\end{mini!}
\noindent where an obstacle at time $k$, denoted by ${o}_{k}$, is displaced by $\B{d}^i_k$ according to some function $g(\cdot)$, which denotes the employed robot strategy to move away objects. 
The robot and obstacle dynamics are represented in constraints~\eqref{eq:cost_fn1}-\eqref{eq:cost_fn2} subject to the upper and lower bounds on the state and control variables as given in~\eqref{eq:cost_fn3}-\eqref{eq:cost_fn6}. 
Finally, constraint \eqref{eq:cost_fn7} represents the collision avoidance constraint between the robot and the obstacles. 


\subsection{Obstacles' Model}

As we noted before, we are not concerned about \textit{how} obstacles are displaced by a robot, and we assume that they can be re-positioned irrespective of the nature of the required robot-obstacle interaction. 
In the formulation of the optimization problem, this is encoded in a generic function $g(\cdot)$.
Therefore, an obstacle may be moved without any constraints. 
Each obstacle is capable of translation and rotation and hence the state $\B{o}^i = [o^{i,x}, o^{i,y}, o^{i,\theta}]$ includes translations in the $x$ and $y$ directions, as well as a rotation. 
This yields the following \textit{linear} model for an obstacle $o^i$:
\begin{equation}
\begin{split}
o^{i,x}_{k+1} &= o^{i,x}_k + d^{i,x}_k,\\
o^{i,y}_{k+1} &= o^{i,y}_k + d^{i,y}_k,\\
o^{i,\theta}_{k+1} &= o^{i,\theta}_k + d^{i,\theta}_k,\\
\end{split}
\label{eq:obs_dyn}
\end{equation}
\noindent where $\B{d}^i_k = [d^{i,x}_k, d^{i,y}_k, d^{i,\theta}_k]$ specify the obstacle's displacement between time instants $k$ and $k+1$. 
Thus, the displacement between two time instants for the obstacle $o^i$ is $ d^{i}_k =\norm{\B{d}^i_k}$.
In analogy to the standard motion model introduced in~\eqref{eq:motion_model}, we may write~\eqref{eq:obs_dyn} compactly as $\B{o}_{k+1} = g(\B{o}_{k}, \B{d}_k)$.

We note that, in many warehouse, logistics, and other structured environments, the placement regions for movable objects are predefined. In such scenarios, these placement regions can be directly treated as displacement sets from which the object configurations are sampled. This eliminates the need to explicitly formulate the obstacle model as described above, thereby reducing the solution search space and, consequently, the computational complexity of the problem.

\subsection{Robot's Model}
We do not assume any specific robot motion model for the validity of our work, and we refer to any standard motion model as in the form given in~\eqref{eq:motion_model}. 
However, as it will become clear in later Sections, we distinguish between linear and nonlinear robot models. 
In this paper, we consider two different linear models.
The first is a trivial model similar to the obstacle model in~\eqref{eq:obs_dyn}, such as
\begin{equation}
\begin{split}
x_{k+1} &= x_k + \tau u^x_k,\\
y_{k+1} &= y_k + \tau u^y_k,\\
\theta_{k+1} &= \theta_k + \tau u^{\theta}_k,\\
\end{split}
\label{eq:robot_model1}
\end{equation}
\noindent with $\B{x}_k = [x_k,y_k,\theta_k]$ as the robot pose and $\B{u}_k = [u^x_k,u^y_k,u^{\theta}_k]$ as the applied control at time instant $k$. 
We also consider the linear dynamics model borrowed from~\cite{van2011lqg} with a robot state $\B{x}_k = [x_k,y_k,v^x_k,v^y_k]$ consisting of its location and velocity, and with acceleration input $\B{u}_k= [a^x_k,a^y_k]$. 
The corresponding model is 
\begin{equation}
\begin{split}
x_{k+1} &= x_k + \tau v^x_k + \frac{\tau^2}{2}a^x_k,\\
y_{k+1} &= y_k + \tau v^y_k + \frac{\tau^2}{2}a^y_k,\\
v^x_{k+1} &= v^x_k + \tau a^x_k,\\
v^y_{k+1} &= v^y_k + \tau a^y_k,\\
\end{split}
\label{eq:robot_model2}
\end{equation}
\noindent where $\tau$ is, again, the duration of a time step. 
The use of linear models allows for obtaining optimal, global solutions, albeit dependent on the nature of the objective function as well as other constraints.
However, the motion model of mobile robots is often nonlinear. 
Nonlinear and non-convex models provide weak theoretical guarantees and the optimization often results in local minima. Nevertheless, we evaluate our approach on two different nonlinear models. The first model is a simple differential drive model~\cite{Lav06} given by
\begin{equation}
\begin{split}
x_{k+1} &= x_k + \tau u_k\cos{\theta_k},\\
y_{k+1} &= y_k + \tau u_k \sin{\theta_k},\\
\theta_{k+1} &= \theta_k + \tau \omega_k,
\end{split}
\label{eq:robot_model3}
\end{equation}
\noindent where $\B{x}_k = [x_k,y_k,\theta_k]$ is the robot state and $\B{u}_k = [u_k,\omega_k]$ is the applied control at time instant $k$. We then consider the case of a non-holonomic car-like robot~\cite{van2012IJRR}
\begin{equation}
\begin{split}
x_{k+1} &= x_k + \tau v_k\cos{\theta_k},\\
y_{k+1} &= y_k + \tau v_k \sin{\theta_k},\\
\theta_{k+1} &= \theta_k + \tau v_k \frac{\tan{\phi_k}}{l},\\
v_{k+1} &= v_k + \tau a_k,
\end{split}
\label{eq:robot_model4}
\end{equation}
\noindent where $\B{x}_k = [x_k,y_k,\theta_k, v_k]$ is the robot state consisting of its pose and speed, $\B{u}_k = (a_k, \phi_k)$ is the applied control consisting of the acceleration and steering wheel angle, $\tau$ is the duration of a time-step, and $l$ is the length of the robot.

\subsection{Collision Avoidance}
So far we have been agnostic to the structure of the constraint in~\eqref{eq:cost_fn7}. In the following paragraphs, we formulate the collision avoidance constraint more precisely. 
We will consider two such constrains, namely
(1) polygon-shaped obstacles, and
(2) circle-shaped obstacles.  

\subsubsection{Polygonal Obstacles}
A polygonal obstacle can be represented as a polyhedron $\mathcal{P}$, that is, the intersection of $N \in \mathbb{N}$ halfspaces, such that
\begin{equation}
\mathcal{P} = \{x \in \mathbb{R}^2| Ax\le b\},
\end{equation} 
where $A =\left (a^1, \ldots, a^N\right )^T$ and $b = \left (b^1, \ldots, b^N\right )^T$. Each row $a^i x\le b^i$ defines one of the halfspaces of $\mathcal{P}$.

For this specific shape, the collision avoidance constraint can be written by means of a mixed integer formulation as

\begin{equation}
a^i\left(\B{p}_k-\B{o^i_k}\right)\ge b^i-c_k^iM, \quad i = 1,\ldots, N \label{eq: milp_ploy_1}
\end{equation} 
\begin{equation}
\sum_{i=1}^N c^i_k \le N-1, \label{eq: milp_ploy_2}
\end{equation} 
where $c^{i}_k \in \left \{0,1\right \}$ are binary auxiliary variables and $M$ is a constant suitably large. 
It is important to note that \eqref{eq: milp_ploy_1} and \eqref{eq: milp_ploy_2} holds $\forall k \in \{1,\ldots,T-1\}$ and $\forall i \in \{1,\ldots,|\mathcal{O}|\}$.  

\subsubsection{Circular Obstacles}
It is common practice to model robot and obstacles using approximate bounding volume spheres~\cite{park2018IEEE, thomas2022RAS}. 
In such cases, the collision avoidance constraint is given by
\begin{equation}
\norm{\B{p}_k - \B{o^i_k}} \geq r^r + r^o,
\label{eq:coll2}
\end{equation}
\noindent where $\B{p}_k$, $\B{o^i_k}$ are the midpoints of the robot and obstacle circles (that is, the 2D projections of the sphere on the supporting plane), respectively, whereas $r^r$ and $r^o$ are the corresponding radii. 


When the robot and all the obstacles are modeled as circles, the collision avoidance constraint in~\eqref{eq:coll2} can be linearized by introducing $j$ binary auxiliary variables $\xi^{j,i}_k \in \left \{0,1\right \}$ for each obstacle $\B{o}^i \in \mathcal{O}$ at every time instant $k$. 
The constraint~\eqref{eq:coll2} is thus rewritten as

\begin{align}
x_k - o^{i,x}_k \geq r^r + r^o - \xi^{1,i}_k M, \label{eq:linear_constraints1}\\
- \left( x_k - o^{i,x}_k \right) \geq r^r + r^o - \xi^{2,i}_k M, \label{eq:linear_constraints2} \\
y_k - o^{i,y}_k \geq r^r + r^o - \xi^{3,i}_k M, \label{eq:linear_constraints3}\\
-\left( y_k - o^{i,y}_k \right) \geq r^r + r^o - \xi^{4,i}_k M, \label{eq:linear_constraints4} \\
\sum_{j=1}^4 \xi^{j,i}_k \leq 3, \label{eq:linear_constraints5}
\end{align}

\noindent where it holds $\forall k \in \{1,\ldots,T-1\}$ and $\forall i \in \{1,\ldots,|\mathcal{O}|\}$, and $M$ is a suitably large constant. 

As a result of \eqref{eq: milp_ploy_1}, \eqref{eq: milp_ploy_2}, \eqref{eq:linear_constraints1}-\eqref{eq:linear_constraints5} the optimization problem in~\eqref{eq:optimization_problem} can be effectively converted into a \textit{Mixed Integer Quadratic Programming} (\textsf{MIQP}) problem. 
\textsf{MIQP}s are a class of optimization problems with a quadratic objective function subject to linear constraints, and involving both continuous and discrete decision variables. 
It is well known from Operations Research theory that there are algorithms guaranteeing global optimality for this type of optimization problem.

In case of nonlinear robot dynamics, the optimization problem~\eqref{eq:optimization_problem} is an instantiation of \textit{Nonlinear Programming} (\textsf{NLP}). 
However, due to the non-convex nature of the \textsf{NLP} formulation, it is not possible to guarantee a global optimum~\cite{salcedo1992IECR}. 
If we choose to reformulate the collision avoidance constraints by introducing auxiliary variables (as shown in~\eqref{eq: milp_ploy_1}-\eqref{eq: milp_ploy_2} and~\eqref{eq:linear_constraints1}-\eqref{eq:linear_constraints5}, the optimization problem transforms into a \textit{Mixed Integer Nonlinear Programming} (\textsf{MINLP}) problem.
This change in problem class increases complexity further due to the combinatorial aspects introduced by the use of binary variables. 
Moreover, global optimization algorithms for \textsf{MINLP}s often necessitate convexity in both the objective and constraint functions~\cite{belotti2013AN}. 
Therefore, for nonlinear robot models, we resort to solving the \textsf{NLP} formulation to obtain sub-optimal solutions.

%
\section{A Horizon-Slicing Heuristic to Solve \textsf{MOD}}
\label{sec:horizion_slicing}
In the previous Section, we have modeled \textsf{MOD} as an optimization problem, as formulated in~\eqref{eq:optimization_problem}. 
The exact solution method does not solve all problem instances efficiently, because of its inherent intractability, and it seems beneficial to search for approximate solutions. 
This is the case in many practical situations in which we are concerned with obtaining a feasible yet sub-optimal solution that has the major advantage of being computed easily. 
As discussed above, in \textsf{MIQP} problems, the worst-case computational complexity grows exponentially with the number of binary and/or integer decision variables~\cite{naik2018PHD}. 
As a case in point, consider the linearisation of~\eqref{eq:coll2}, where for each obstacle we have introduced four binary variables~\eqref{eq:linear_constraints1}-\eqref{eq:linear_constraints4} to formulate the optimization as an \textsf{MIQP} problem when the robot dynamics is linear. 
The number of feasible combinations of the binary variables is thus $2^{4 \times T \times |\mathcal{O}|}$, where $T$ is the number of time steps required to reach the goal location. 
With the increase in the number of movable obstacles and the environment \textit{size} (in the sense that the larger the environment, the higher the value of $T$ on average), the optimization process tends to be computationally unfeasible. 
The computational burden could be reduced by decreasing the number of variables. 
One \textit{naïve} way to achieve this result is by progressively solving subsets of the overall problem, each subset being computationally less expensive compared to solving the complete optimization problem at once. 
However, we have already seen that \textsf{MOD} does not exhibit the optimal substructure property. 
Therefore, solving subsets of the whole problem does not necessarily guarantee an optimal solution at the global level.

\begin{figure}[t!]
\centering
\includegraphics[scale=0.49]{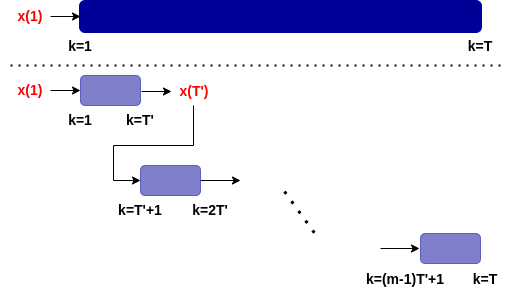}
\caption{
An illustration of the horizon-slicing approach employed in this work. 
$x(\cdot)$ denotes the variables of interest. 
The input (initial state) to the original problem and the first slice of the sub-problem coincide, and are referred to as $x(1)$. 
The output of the first slice $x(T')$ is the initial state for the next slice. 
The values of the time-dependent variables are passed on to the slices that follow.}
\label{fig:slicing}
\end{figure}

\subsection{\textsf{HS4MOD}}

In order to reduce the computational effort (at the expense of giving up optimality) we temporally partition the whole optimization problem into $m$ sub-problems, each one of duration $T'$ such that $mT' = T$. 
In this paper, we simply \textit{slice} the number of time steps required to reach the goal location into a smaller duration, and solve the optimization problem for each slice separately. 
Once each sub-problem undergoes a specific optimization process, the values of time-dependent variables are passed on to the sub-problem that follows. 
This concept is illustrated in Fig.~\ref{fig:slicing}.
It should be pointed out that different, \textit{probably smarter} strategies to horizon-slicing may be employed, for instance related to the adopted motion planning approach.
A comprehensive analysis of such strategies is outside the scope of this paper, whereas the adopted horizon-slicing approach serve the purpose of allowing us to draw a number of general conclusions about sub-optimality and time complexity, which can be used as prerequisite work to ground further strategies.


Let us consider the objective function $J$ in~\eqref{eq:cost_fn0}, and let us denote the optimal objective function \textit{without} horizon-slicing as $J^\star$. 
We thus have:
\begin{equation}
\begin{split}
J^\star &= w^xc(\B{p}) + w^d \sum_{i=1}^{\mathcal{M}} c^i(D^i) \\
&= \sum\limits_{k=0}^{T-1} w^x \norm{\B{p}_{k+1}- \B{p}_{k}}^2 + w^x w^g\norm{\B{p}_T - \B{p}^g}^2 + \\ & \ \ \ \ \ w^d \sum_{i=1}^{\mathcal{M}} \sum_{k=0}^T d^i_k \\
&=\sum\limits_{k=0}^{T-1} \left(w^x \norm{\B{p}_{k+1}- \B{p}_{k}}^2 + w^d \sum_{i=1}^{\mathcal{M}} d^i_k \right) + \\
&\ \ \ \ \ w^x w^g\norm{\B{p}_T - \B{p}^g}^2 + w^d \sum_{i=1}^{\mathcal{M}} d^i_T.\\
\end{split}
\label{eq:horizon_optimal}
\end{equation}
Let us now focus on the terms inside the bracket. 
This sum can be re-written as
\begin{equation}
\begin{split}
&\sum\limits_{k=0}^{T-1} w^x \norm{\B{p}_{k+1}- \B{p}_{k}}^2 + w^d \sum_{i=1}^{\mathcal{M}} d^i_k\\ &=
\left[\sum\limits_{k=0}^{T'-1} w^x \norm{\B{p}_{k+1}- \B{p}_{k}}^2 + w^d \sum_{i=1}^{\mathcal{M}} d^i_k\right] + \\
& \ \ \ \ \ \ldots + \\ & \ \ \ \  \left[\sum\limits_{k=(m-1)T'}^{T-1} w^x \norm{\B{p}_{k+1}- \B{p}_{k}}^2 + w^d \sum_{i=1}^{\mathcal{M}} d^i_k\right]\\
&= J^{\star,s}_{0\rightarrow T'-1} + \ldots + J^{\star,s}_{(m-1)T' \rightarrow T-1}\\
& = \sum_{j=1}^m J^{\star,s}_{(j-1)T'\rightarrow jT'-1},\\
\end{split}
\label{eq:horizon_optimal1}
\end{equation}
\noindent where 
\begin{equation}
\label{subobjective}
J^{\star,s}_{(j-1)T'\rightarrow jT'-1}=\sum_{k=(j-1)T'}^{jT'-1} \left( w^x \norm{\B{p}_{k+1}- \B{p}_{k}}^2 + w^d \sum_{i=1}^{\mathcal{M}} d^i_k\right),
\end{equation}
\noindent and $m$ is the number of slices. 
Thus,~\eqref{eq:horizon_optimal} can be written as
\begin{equation}
 J^\star =  \sum_{j=1}^m J^{\star,s}_{(j-1)T'\rightarrow jT'-1}  +  w^x w^g\norm{\B{p}_T - \B{p}^g}^2 + w^d \sum_{i=1}^{\mathcal{M}} d^i_T.
\label{eq:horizon_optimal2}
\end{equation}
Let us now consider the scenario wherein we partition the horizon and address $m$ distinct optimization problems. 
For each horizon slice $j$, the corresponding objective function is formulated as
\begin{multline}
J^j = \sum_{k=(j-1)T'}^{jT'-1} \left(w^x \norm{\B{p}_{k+1}- \B{p}_{k}}^2 + w^d \sum_{i=1}^{\mathcal{M}} d^i_k \right)+ \\ w^x w^g\norm{\B{p}_{jT'-1} - \B{p}^g}^2 + w^d \sum_{i=1}^{\mathcal{M}} d^i_{jT'-1}.
\label{eq:horizon_sliced}
\end{multline}
Let us define $\eta^j = w^x w^g\norm{\B{p}_{jT'-1} - \B{p}^g}^2 + w^d \sum_{i=1}^{\mathcal{M}} d^i_{jT'-1}$. 
Using the simplification employed in~\eqref{eq:horizon_optimal1},~\eqref{eq:horizon_sliced} can be written as
\begin{equation}
J^j =  J^{j,s}_{(j-1)T'\rightarrow jT'-1}+ \eta^j.
\label{eq:horizon_sliced1}
\end{equation}
Adding the objective function for the $m$ horizon slices and using the fact that 
\begin{equation}
\eta^m =   w^x w^g\norm{\B{p}_T - \B{p}^g}^2 + w^d \sum_{i=1}^{\mathcal{M}} d^i_T,
\end{equation}
we have
\begin{equation}
\begin{split}
J &= \sum_{j=1}^m \left(J^{j,s}_{(j-1)T'\rightarrow jT'-1} + \eta^j\right) \\
& = \sum_{j=1}^m J^{j,s}_{(j-1)T'\rightarrow jT'-1} + \sum_{j=1}^{m-1} \eta^j \ \ + \ \ \\& \ \ \ \ w^x w^g\norm{\B{p}_T - \B{p}^g}^2 + w^d \sum_{i=1}^{\mathcal{M}} d^i_T.   \\
\end{split}
\label{eq:horizon_sliced2}
\end{equation}

\begin{algorithm} 
\caption{\textsf{HS4MOD}}
\label{alg:HSMOD}
\begin{algorithmic}[1]
\Require $T, T', m, \B{x}^s, \B{x}^g, \mathcal{O}, \mathcal{M}, w^x, w^d, w^g, f(\cdot), g(\cdot)$
\For{$j = 1, \dots, m$}
\State $J^j \gets 0 $
\For{$k = (j-1)T',\ldots, jT'-1$}
\State $J^j \gets J^j + w^x \norm{\B{p}_{k+1}- \B{p}_{k}}^2 + w^d \sum_{i=1}^{\mathcal{M}} d^i_k $
\State $\B{x}_{k+1}=f( \B{x}_k,\B{u}_k )$
\State $\B{o}_{k+1}=g( \B{o}_k,\B{d}_k ) $
\Statex \hspace{.94cm} {\text{such that} \ \eqref{eq:cost_fn1}-\eqref{eq:cost_fn7}}
\EndFor
\State $J^j \gets J^j + w^x w^g\norm{\B{p}_{jT'-1} - \B{p}^g}^2 + w^d \sum_{i=1}^{\mathcal{M}} d^i_{jT'-1}$
\State $\text{Minimize} \,\, J^j  \,\,\text{such that} \,\,\eqref{eq:cost_fn1}-\eqref{eq:cost_fn7}$
\EndFor
\Ensure $J, \{\B{x}_0, \ldots, \B{x}_T$\}, $\{d^i_k\} \ \forall i \in \mathcal{M} \ \text{and} \ \forall k \in \{1, \dots, T\}$
\end{algorithmic}
\end{algorithm}

It can be readily verified that equation~\eqref{eq:horizon_sliced2} exhibits a structure similar to~\eqref{eq:horizon_optimal1}, but with an added terminal cost for each slice except the final one. 
This additional term is present in each sub-problem as it enables the robot to keep track of the goal. 
The overall \textsf{HS4MOD} algorithm is summarized in Algorithm~\ref{alg:HSMOD}. 
For each slice $j$ the exact optimization is performed (lines 2-7). 
The summation in~\eqref{eq:horizon_sliced} is carried out (as shown in line 4), and the goal tracking term is added (line 8) to obtain the objective function $J^j$~\eqref{eq:horizon_sliced} for each slice. 
The optimization is then carried out subject to the constraints given in~\eqref{eq:cost_fn1}-\eqref{eq:cost_fn7}.

\subsection{Analysis of the Approximate Solution}

In order to establish a meaningful comparison between $J$ and $J^\star$, we introduce a new objective function, denoted as $J'$, which aggregates the cost obtained by summing up the costs of all the $m$ individual slices, and is thus given by

\begin{equation}
J' = J - \sum_{j=1}^{m-1} \eta^j.
\end{equation}

\begin{figure}[t!]
\subfloat[]{\includegraphics[trim=60 118 45 104,clip,width=7.5cm,height=1.7cm]{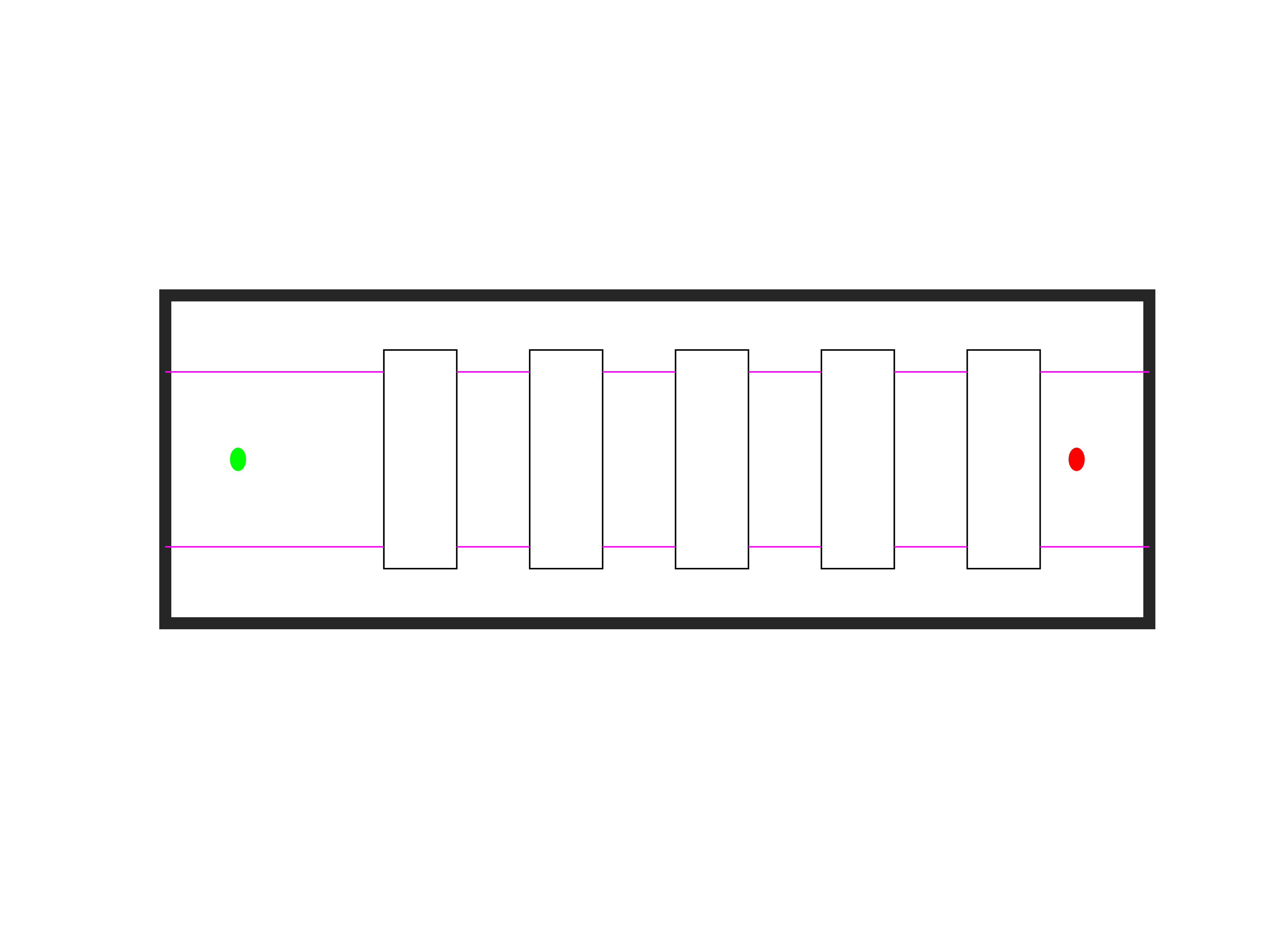}\label{fig:rec1}}\vspace{5pt}\\
\subfloat[]{\includegraphics[trim=60 118 45 104,clip,width=7.5cm,height=1.7cm]{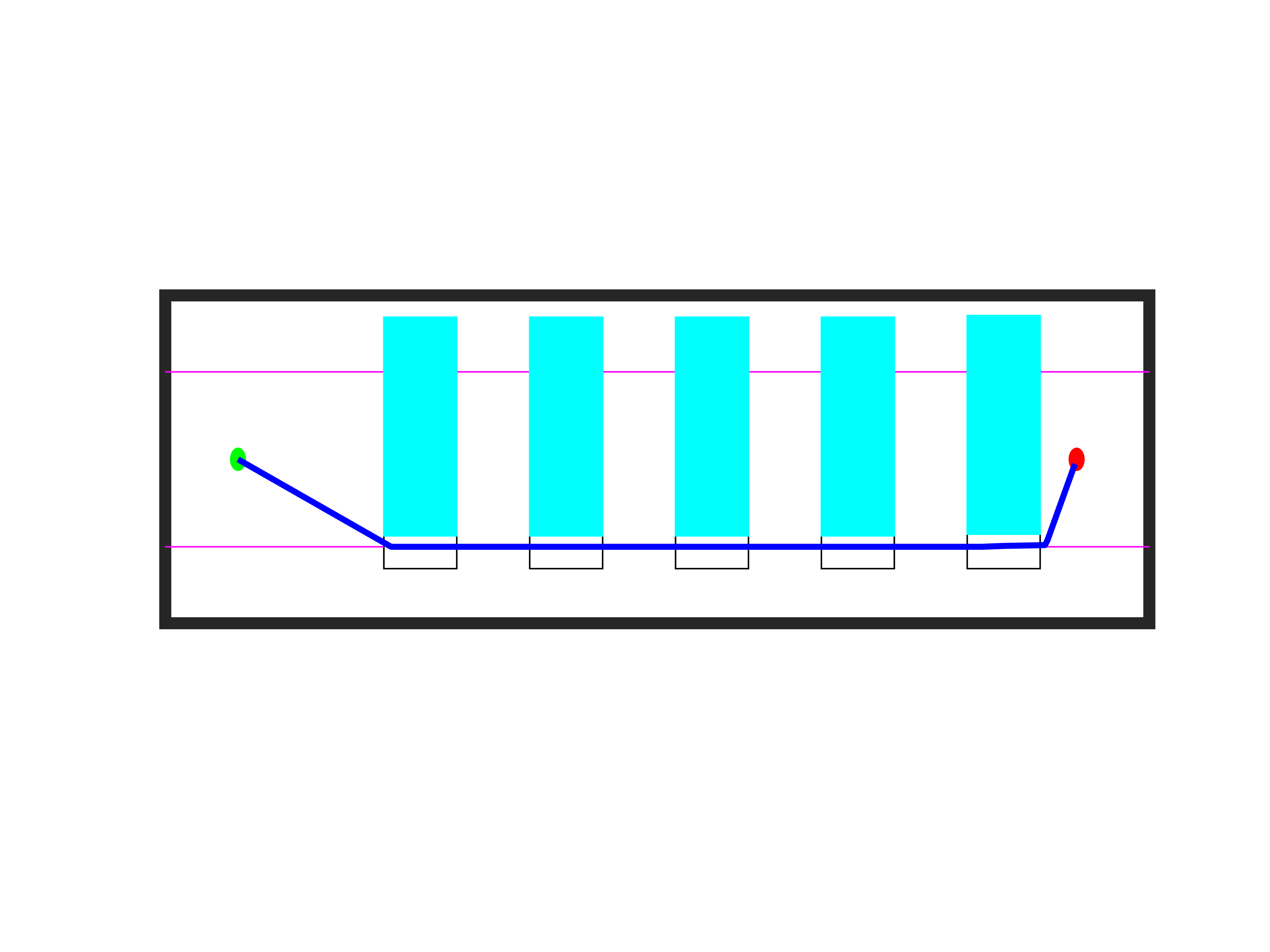}\label{fig:rec2}}\vspace{5pt}\\
\subfloat[]{\includegraphics[trim=60 118 45 104,clip,width=7.5cm,height=1.7cm]{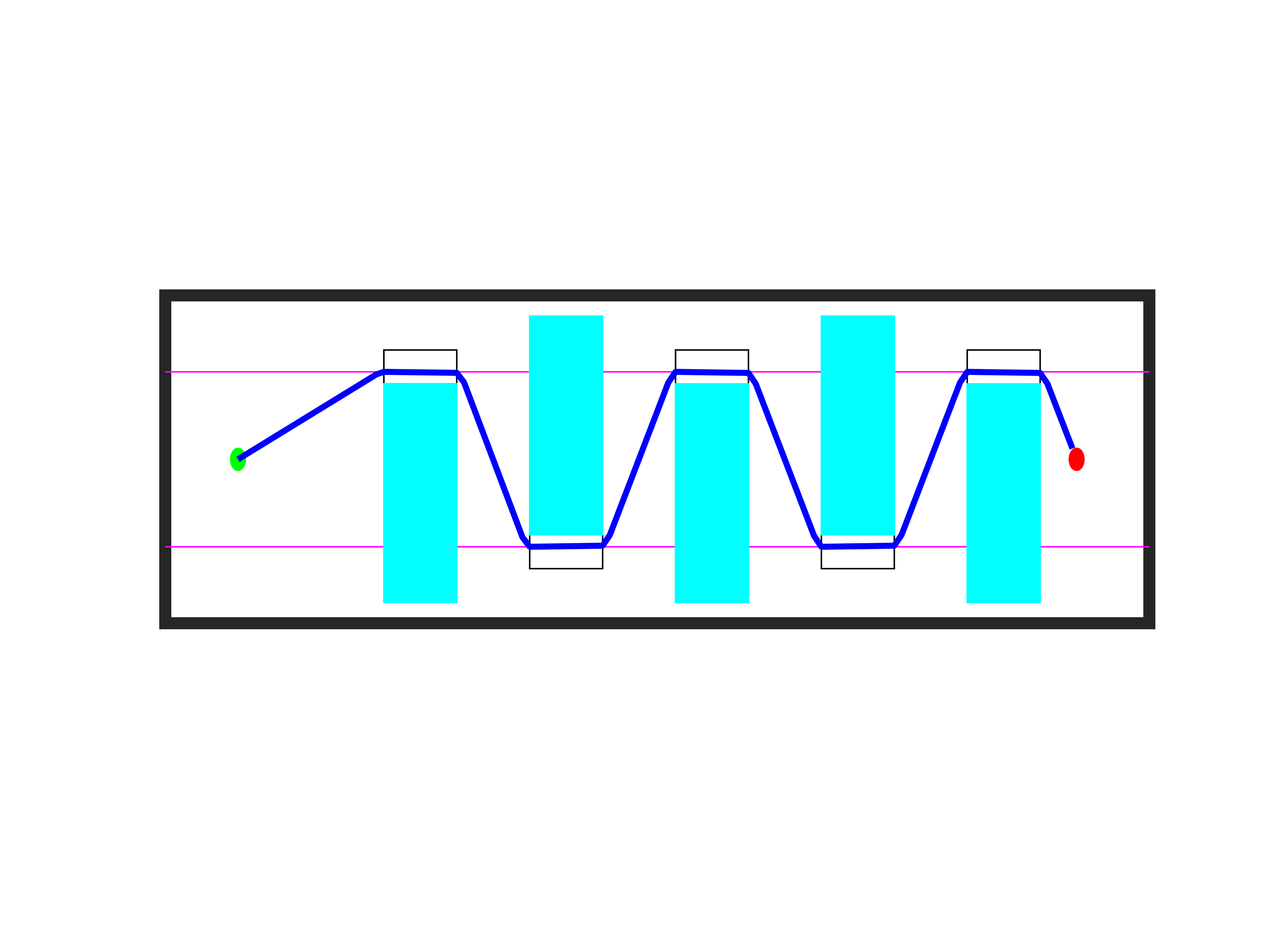}\label{fig:rec3}}\vspace{5pt}\\
\subfloat[]{\includegraphics[trim=60 118 45 104,clip,width=7.5cm,height=1.7cm]{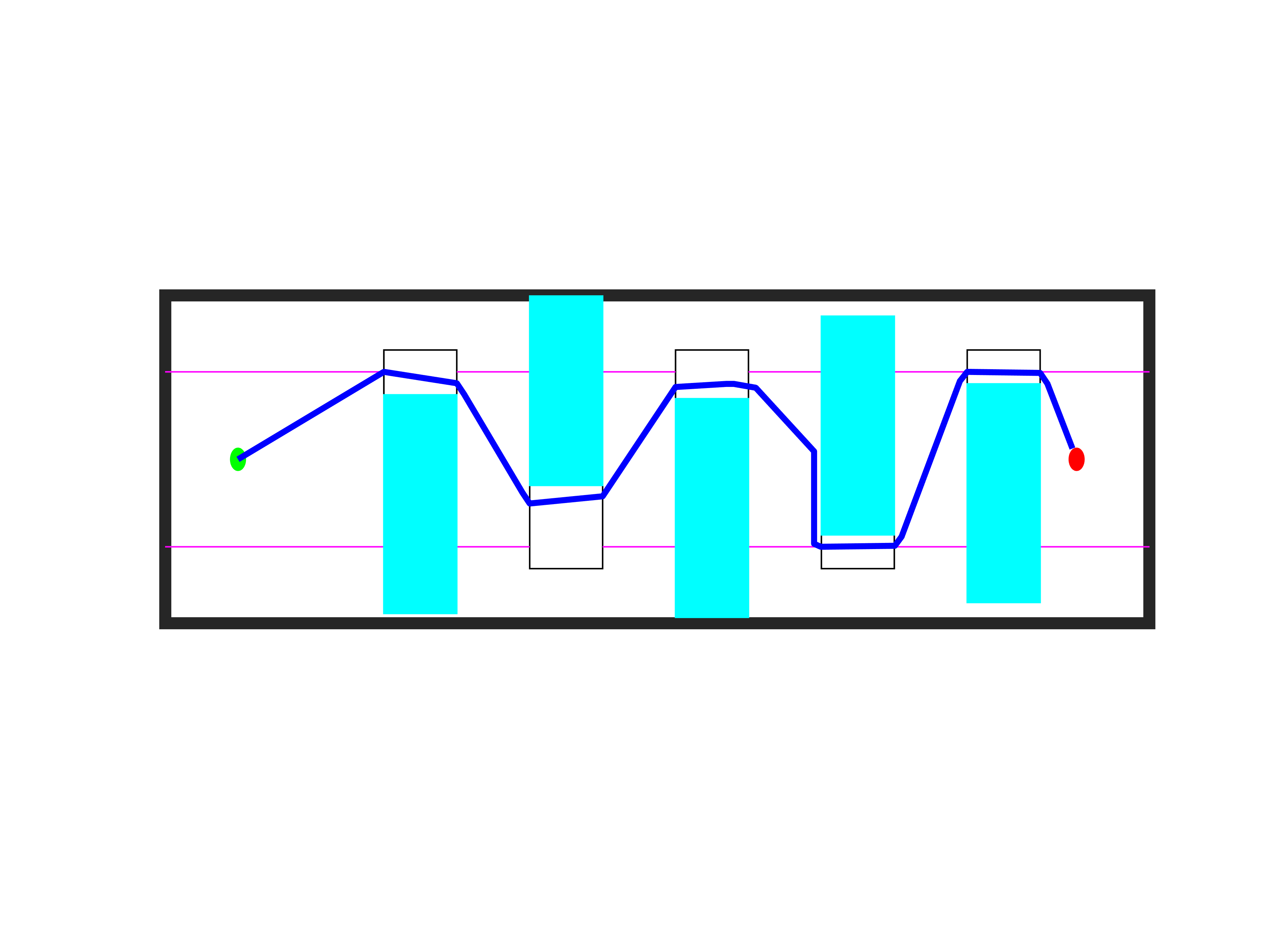}\label{fig:rec4}}\\
\caption{
(a) A robot is tasked to move from the left (start location in green) to the right-hand side (goal location in red) in the presence of 5 movable obstacles. 
The $y$-coordinate of the robot is constrained to lie within the magenta lines. 
(b) Obstacles can only translate in the vertical direction.
The cases $m=1$ and $m=2$ produce similar trajectories (obstacle displacements shown in cyan). 
(c)-(d) Odd obstacles (counting from left) can only move downward and even obstacles can only move upward.
The optimal solution is given in (c).
The case $m=2$ gives a sub-optimal solution (with $\epsilon = 0.81$) as shown in (d) with the first sub-problem involving the first $3$ obstacles, as shown by the vertical trajectory after the third obstacle.}
\label{fig:rectangle}
\end{figure}

$J'$ has a structure similar to that of $J^\star$. 
As observed before, from an algorithmic perspective, a subset of an optimal solution is not necessarily globally optimal for the particular problem we consider.
However, it can be easily verified that the solutions to all the subsets are naturally feasible solutions. 
Therefore, the horizon-slicing approach results in a feasible solution to the original problem, and therefore the value of $J'$ must be an upper bound to $J^\star$, that is, $J' \geq J^\star$. 
For this reason, we have that
\begin{equation}
\begin{split}
J' & = \sum_{j=1}^m J^{j,s}_{(j-1)T'\rightarrow jT'-1} + \eta^m \\
& = \sum_{j=1}^m \left(J^{\star,s}_{(j-1)T'\rightarrow jT'-1} + \delta^i \right) + \eta^m \\
& \leq \sum_{j=1}^m \left(J^{\star,s}_{(j-1)T'\rightarrow jT'-1} + \delta^{max} \right) + \eta^m\\
&  \leq \left(\sum_{j=1}^m J^{\star,s}_{(j-1)T'\rightarrow jT'-1}  + \eta^m \right) + m\delta^{max} \\
& \leq J^\star + m\delta^{max},
\end{split}
\label{eq:upperbound_J}
\end{equation}
\noindent where
\begin{equation}
\delta^i = J^{j,s}_{(j-1)T'\rightarrow jT'-1} - J^{\star,s}_{(j-1)T'\rightarrow jT'-1},
\end{equation}
with each $\delta^i$ bounded by $\pm \delta^{max}$, that is
\begin{equation}
\delta^i \in  [-\delta^{max},+\delta^{max}].
\end{equation}
From~\eqref{eq:upperbound_J}, we thus have
\begin{equation}
J^\star \leq J' \leq J^\star + m\delta^{max}.
\label{eq:epsilon_boud}
\end{equation}
Let the deviation from the optimal solution $J' - J^\star$ be a factor of $J^\star$, that is, $J' - J^\star = \epsilon J^\star$. 
Ideally, we would want $\epsilon \to 0$ with considerable computational efficiency. 
We note here that, in optimization literature, the standard metric used to compute the relative distance between an approximate and the optimal solution is the \textit{optimality gap}. 
Essentially, $\epsilon$ computes this relative distance and is therefore equivalent to the optimality gap since
\begin{equation}
\textrm{optimality gap} = \frac{J'- J^\star}{J^\star} = \frac{\epsilon J^\star}{J^\star} = \epsilon.
\end{equation}
From~\eqref{eq:epsilon_boud}, we have
\begin{equation}
J'- J^\star \leq m \delta^{max} \implies \epsilon J^\star \leq m \delta^{max}
\end{equation}
and therefore
\begin{equation}
\epsilon \leq m \frac{\delta^{max}}{J^\star}.
\label{eq:orderof1}
\end{equation}
This relation characterizes the relative deviation for a given problem instance; it should not be interpreted as an a priori approximation guarantee. Clearly, for a given problem, as the number of horizon slices or sub-problems increase, the value of $\epsilon$ grows as well. 
The exact deviation $\epsilon$ for a given $m$ is problem-specific (see Fig.~\ref{fig:rectangle}) since $\delta^{max}$ depends on a number of parameters such as the robot and obstacle dynamics, size of the environment, distribution of the obstacles in the environment, number of slices. 
In Section~\ref{sec:results}, we provide an empirical analysis to determine the best $m$ by performing different experiments while varying some of the parameters stated above.

\section{Evaluation and Discussion}
\label{sec:results}
In this Section, we evaluate our approach using 
(1) the optimal but computationally demanding exact approach discussed in Section \ref{sec:approach}, and 
(2) the \textsf{HS4MOD} sub-optimal approach, which was introduced in Section \ref{sec:horizion_slicing}. 
The optimization problem in~\eqref{eq:optimization_problem} is performed using the IBM ILOG CPLEX Optimization Studio V12.10.0 in MATLAB under the YALMIP interface \cite{Lofberg2004}. 
When the robot model is nonlinear, optimization is achieved employing the IPOPT solver~\cite{wachter2006MP}. In both cases, the time limit has been set up to 1000 seconds.
The performance is evaluated on an Intel{\small\textregistered} i7-10850H CPU @ 2.70GHz with 32 GB RAM under Windows 11. The horizon \(T\) was set between 23 and 29 for the SQUARE and RANDOM environments and to 50 for the ROBOT environments. All experiments were conducted in MATLAB/YALMIP using CPLEX with its default relative and absolute MIP gaps (\(10^{-4}\) and \(10^{-6}\)) and IPOPT with an overall tolerance of \(10^{-7}\), feasibility and complementarity tolerances of \(10^{-4}\), and acceptable termination after 15 consecutive iterations satisfying a tolerance of \(10^{-6}\). The objective weights were selected to prioritize goal attainment, followed by displacement minimization.

\begin{figure*}[t!]
\subfloat[\textit{SQUARE 1}]    {\includegraphics[trim=75 89 61 72,clip,scale=0.32]{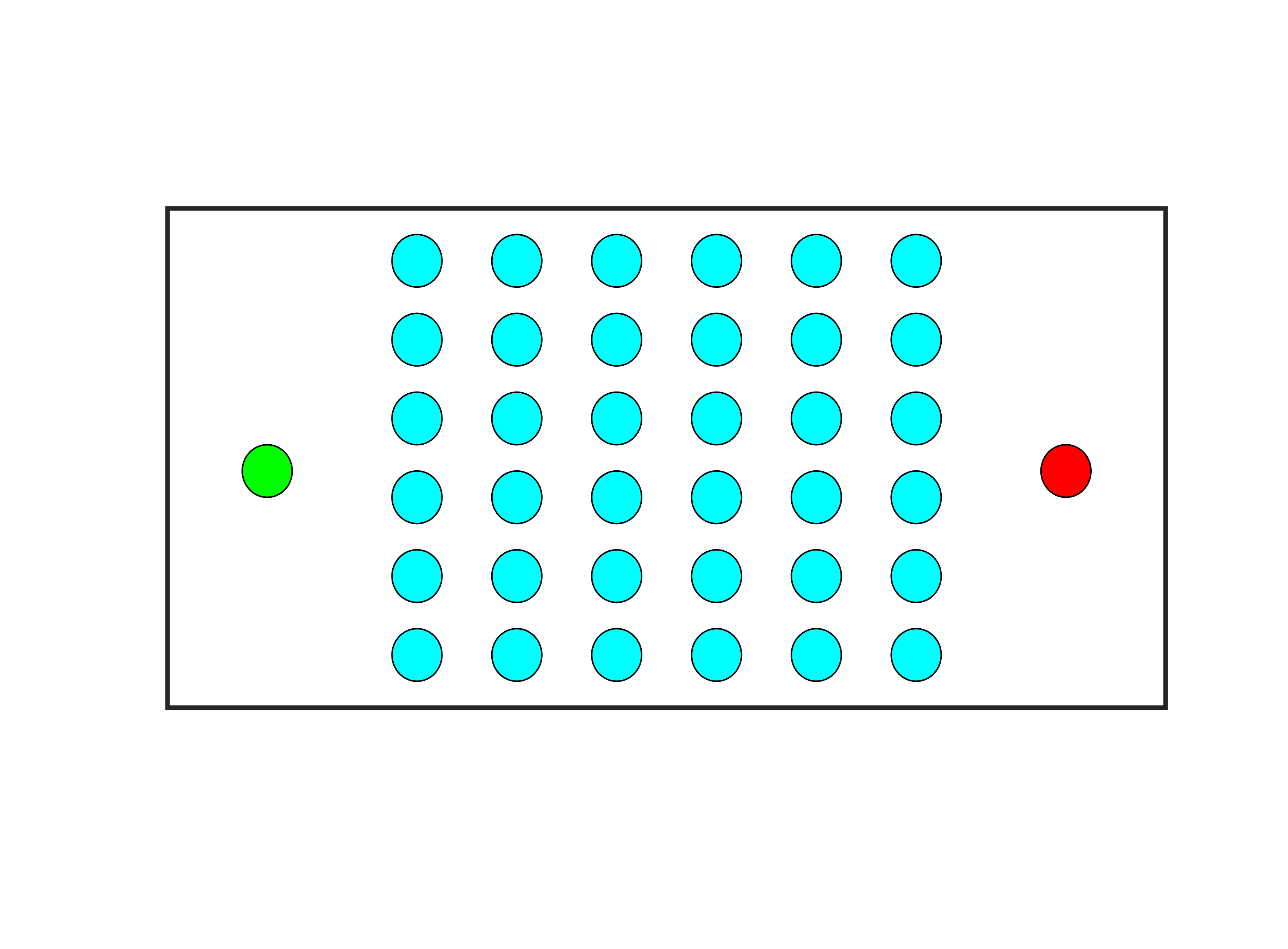}    \label{fig:E1}} \vspace{0.5cm} \\
\subfloat[\textit{SQUARE 2}]    {\includegraphics[trim=75 89 61 72,clip,scale=0.32]{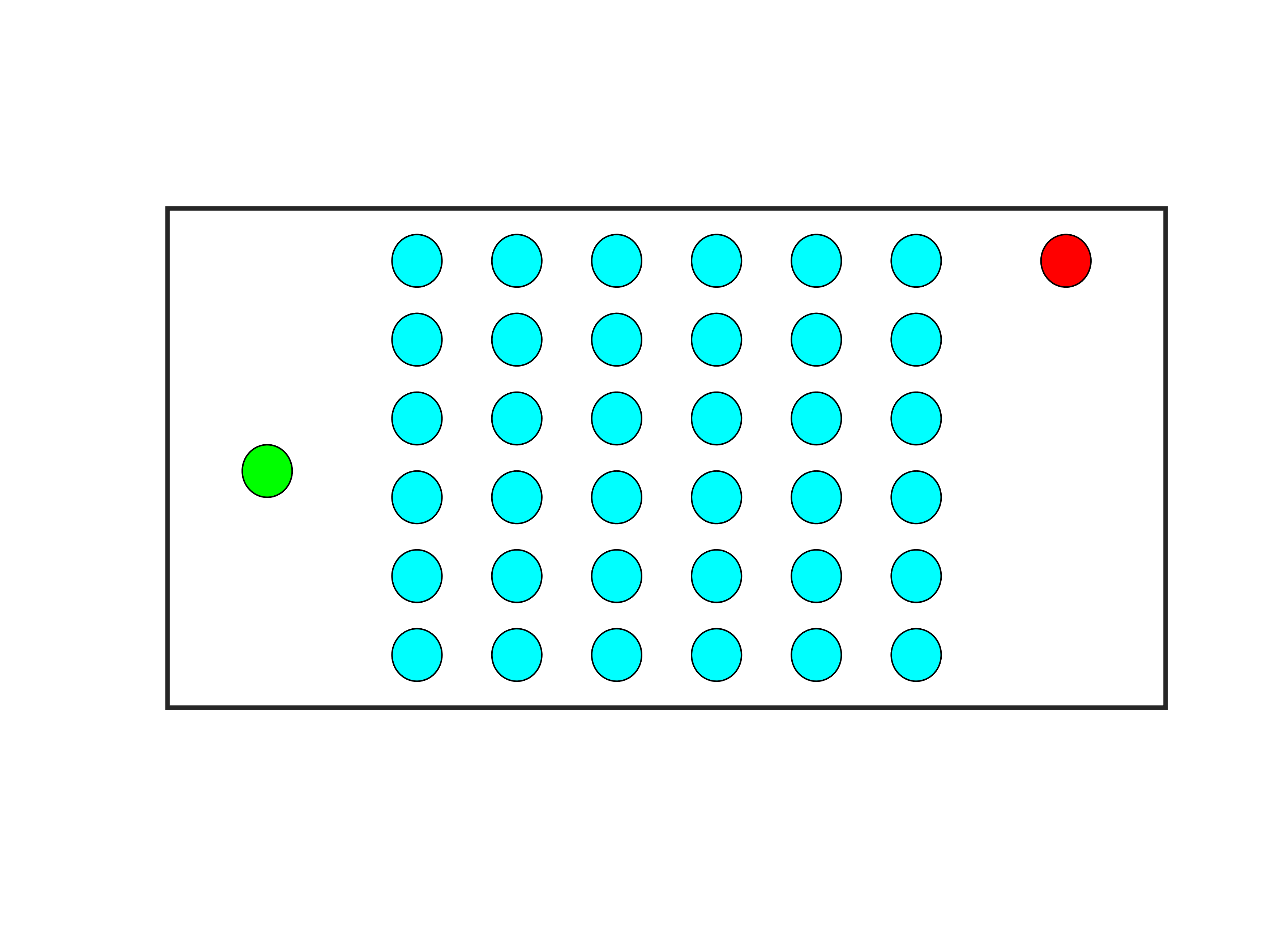}    \label{fig:E2}} \vspace{0.5cm} \\
\subfloat[\textit{ROBOT}]       {\includegraphics[trim=60 124 50 115,clip,scale=0.62]{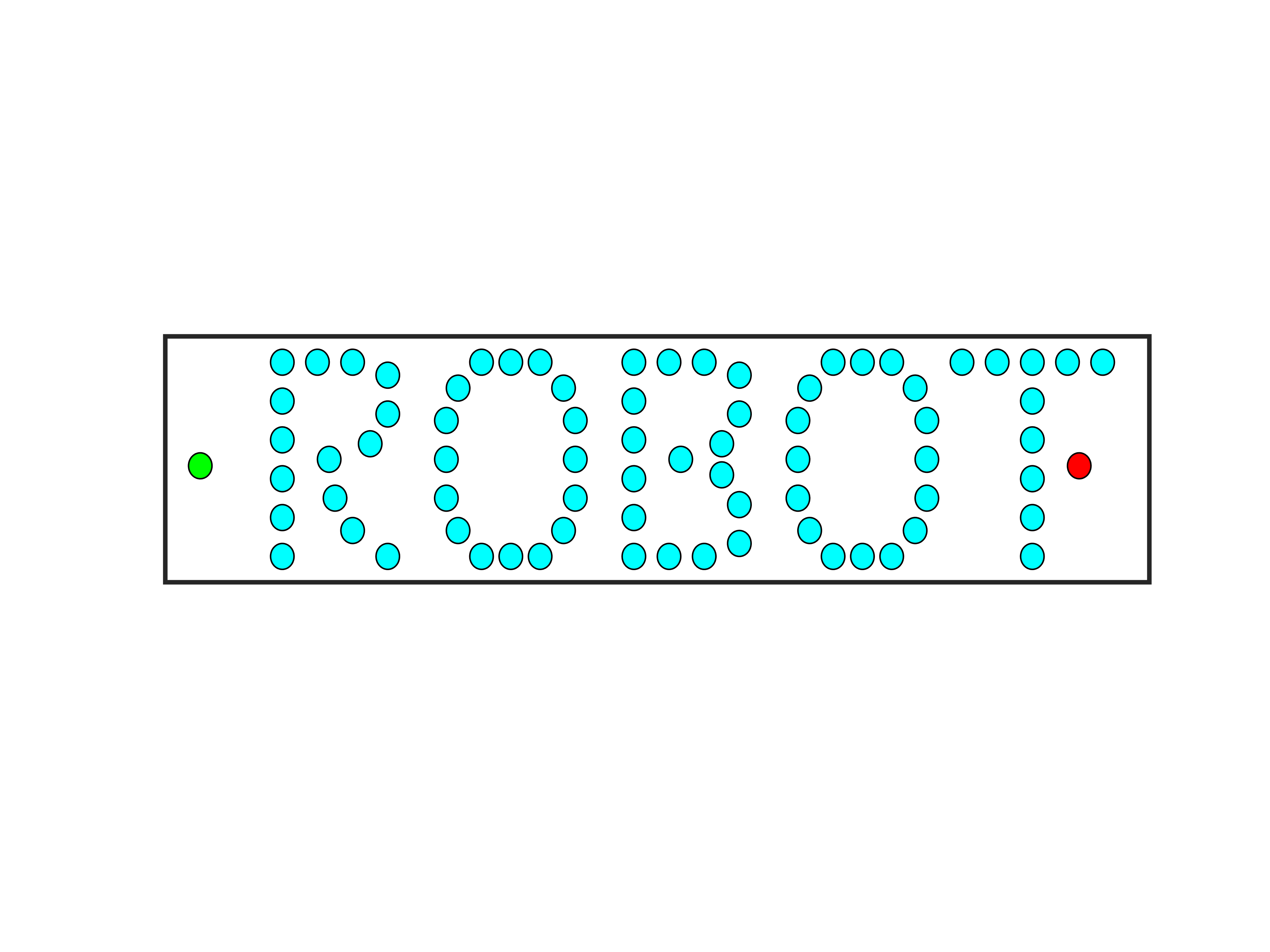}    \label{fig:E3}} \vspace{0.5cm} \\
\subfloat[\textit{RANDOM}]      {\includegraphics[trim=62 42 50 30,clip,scale=0.20]{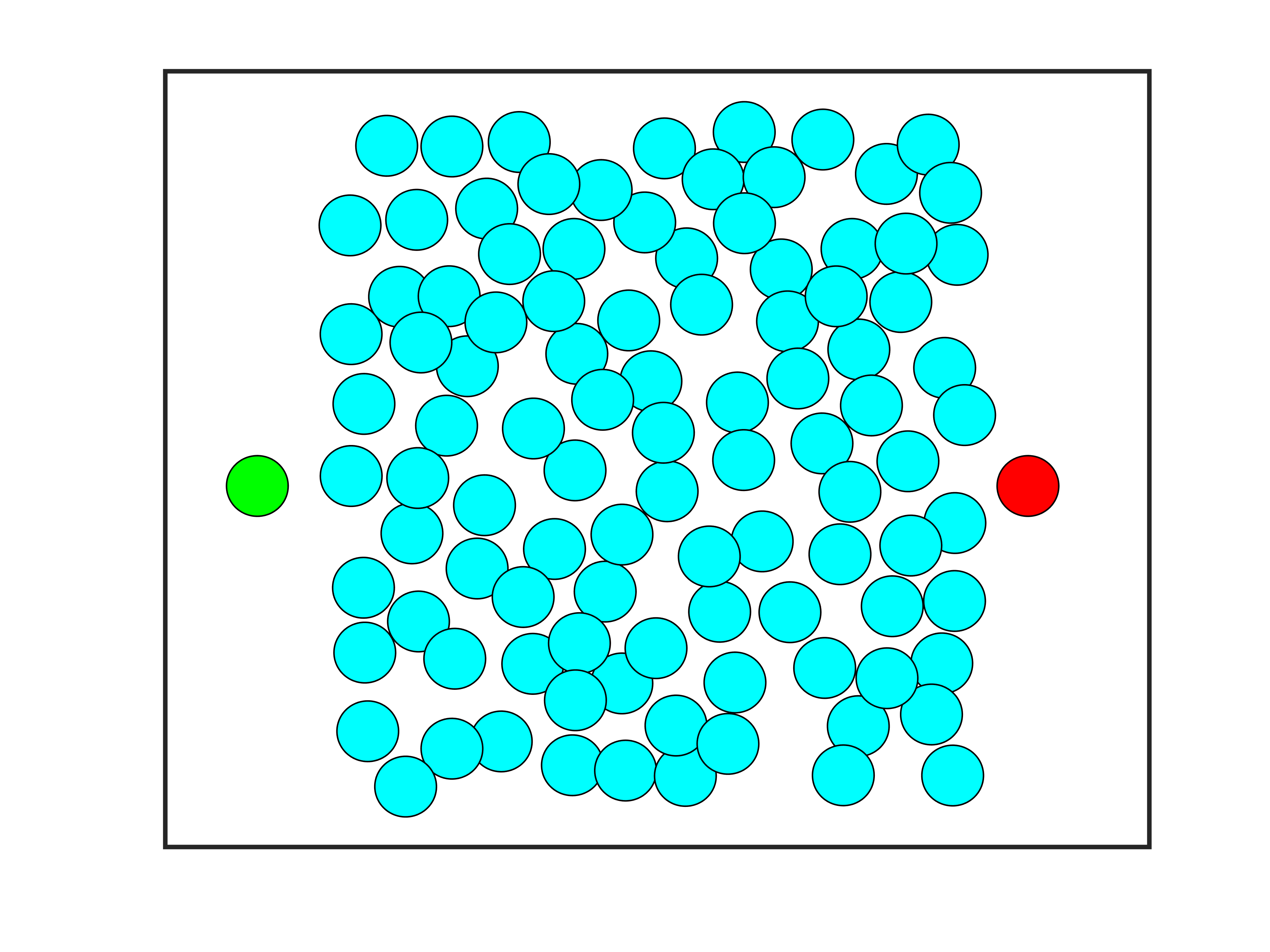}     \label{fig:E4}} 
\caption{
Different domains are used to test our methods. 
In all the experiments, a circular robot is asked to move from the left (start location in green) to the right (goal location in red) hand side in the presence of movable circular obstacles (in cyan).
}
\label{fig:environments}
\end{figure*}

\textsf{HS4MOD} has been tested on three different domains, namely, 
(1) a \textsf{SQUARE} domain, of size $24 \times 24$ meters, with $36$ obstacles --- we consider two domains with the same obstacle configurations but different goal locations, naming them as \textsf{SQUARE\_1} (see Fig.~\ref{fig:E1}) and \textsf{SQUARE\_2} (Fig.~\ref{fig:E2}), 
(2) a \textsf{ROBOT} domain, of size $79 \times 18$ meters, with $74$ obstacles (Fig.~\ref{fig:E3}), and 
(3) a \textsf{RANDOM} domain, of size $24 \times 24$ meters, with $100$ obstacles (Fig.~\ref{fig:E4}). 
The robot must move from a start location (marked in green in Fig.~\ref{fig:environments}) to a goal location (marked in red) in the presence of movable circular obstacles (in cyan). 
As discussed in Section~\ref{sec:approach}, $m$ denotes the number of horizon slices and $m=1$ corresponds to the \textit{exact} approach.
At present, the values of $w^{g}$, $w^x$, and $w^d$ are determined through empirical tuning. 
Unless otherwise specified, we use $w^{g} = 10$, $w^x = 0.5$, and $w^d = 100$ for all the experiments.


\begin{figure*}[]
\subfloat{\vspace{-10pt}\includegraphics[trim=500 315 380 310,clip,width=3.5cm,height=2.4cm]{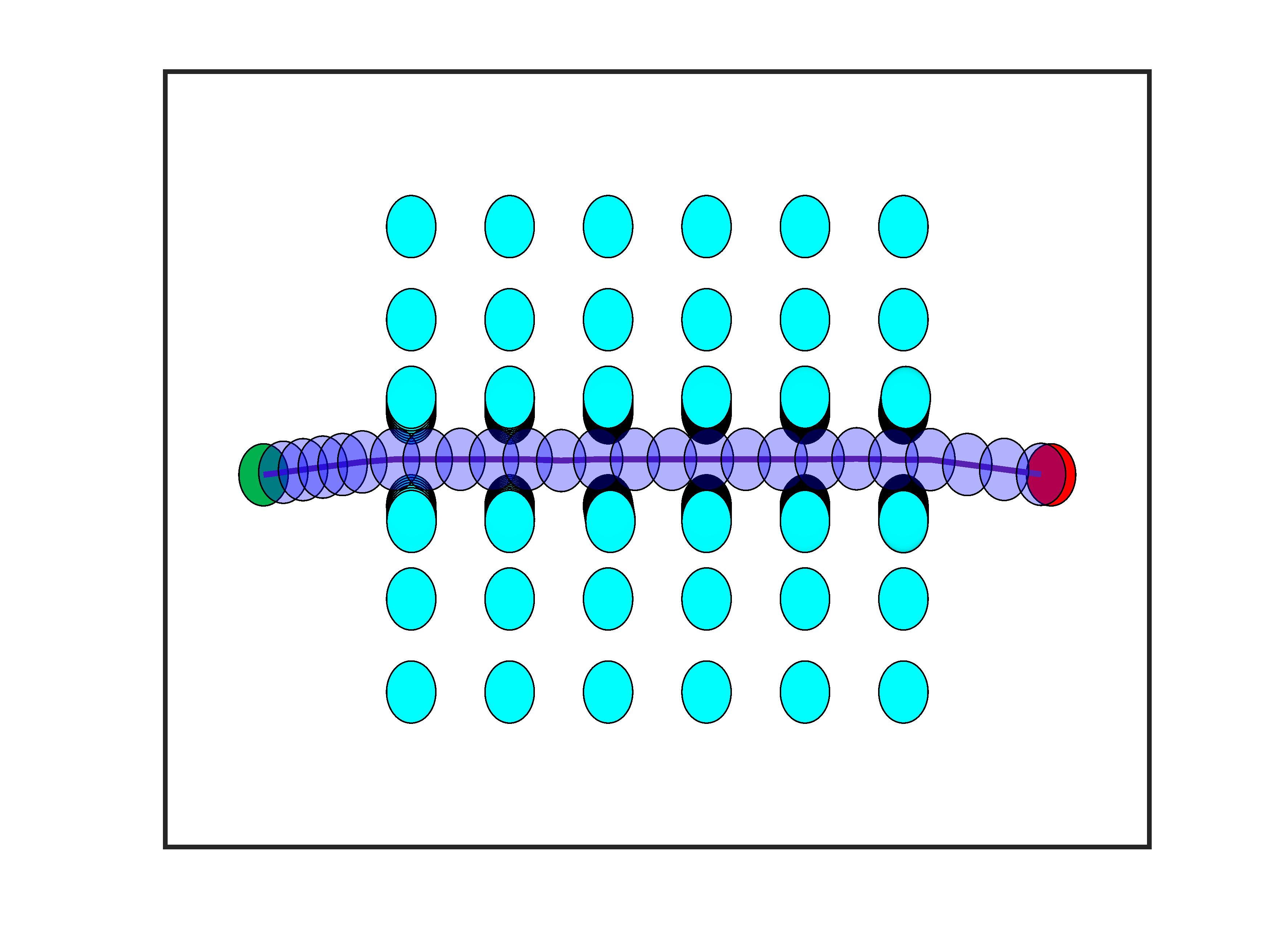}\label{fig:C11}}
\subfloat{\vspace{-10pt}\includegraphics[trim=85 50 57 68,clip,width=3.5cm,height=2.25cm]{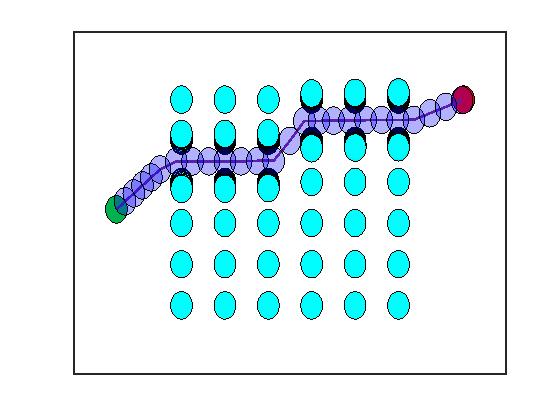}\label{fig:C21}}
    
\subfloat{\vspace{-10pt}\includegraphics[trim=500 315 380 310,clip,width=3.5cm,height=2.4cm]{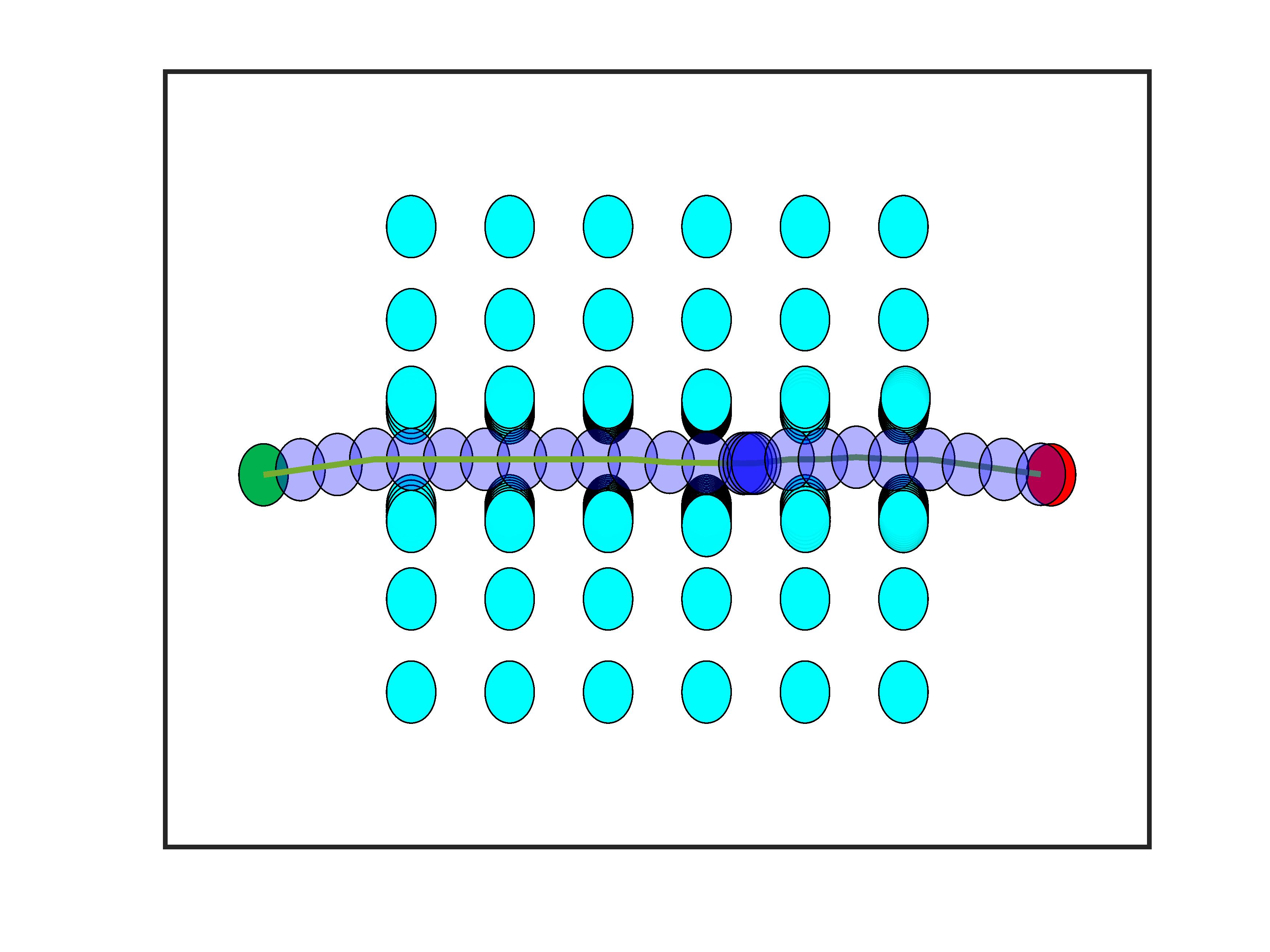}\label{fig:C12}}
\subfloat{\vspace{-10pt}\includegraphics[trim=480 350 360 270,clip,width=3.5cm,height=2.45cm]{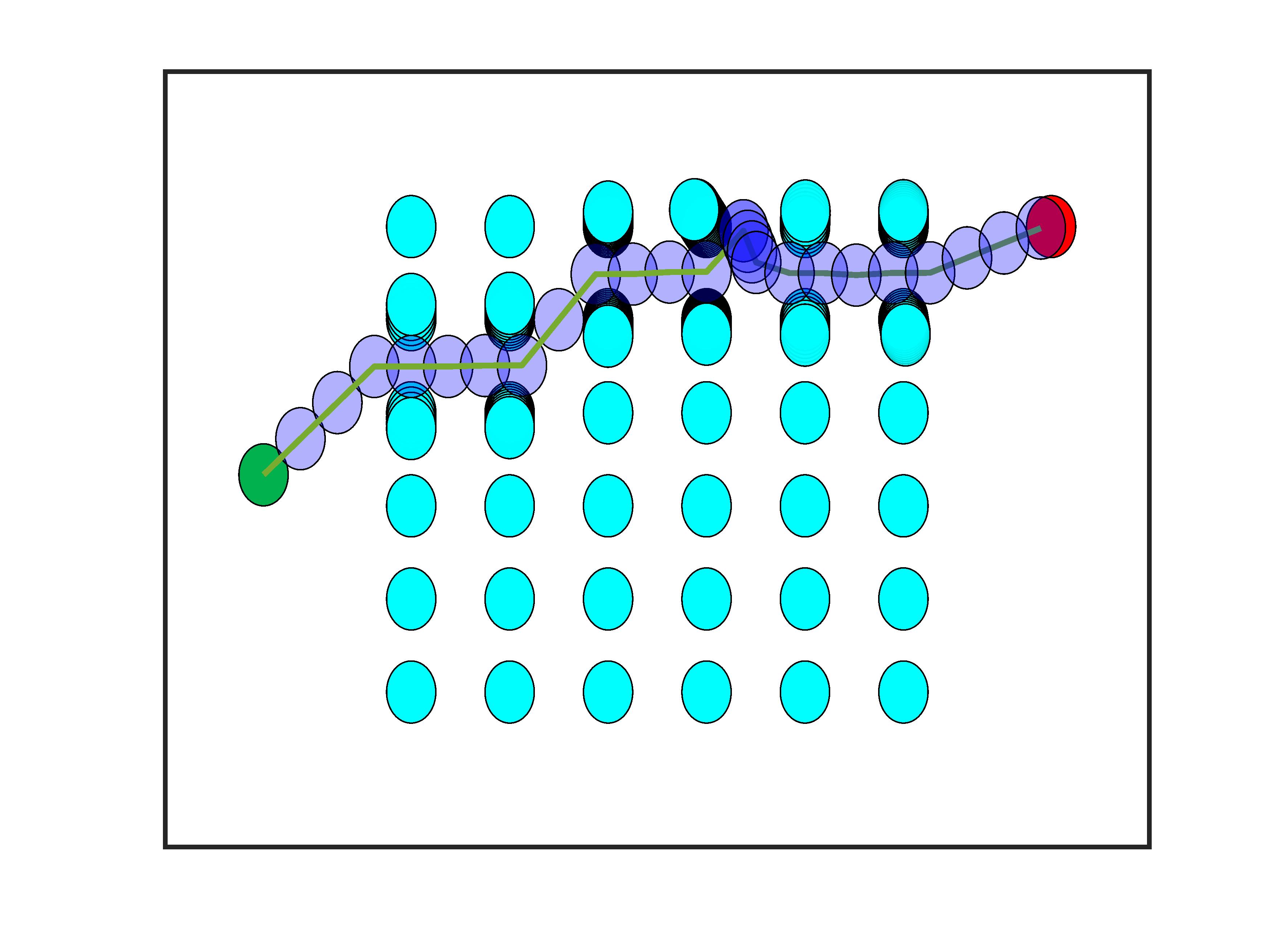}\label{fig:C22}}
     
\subfloat{\vspace{-10pt}\includegraphics[trim=500 315 380 310,clip,width=3.5cm,height=2.4cm]{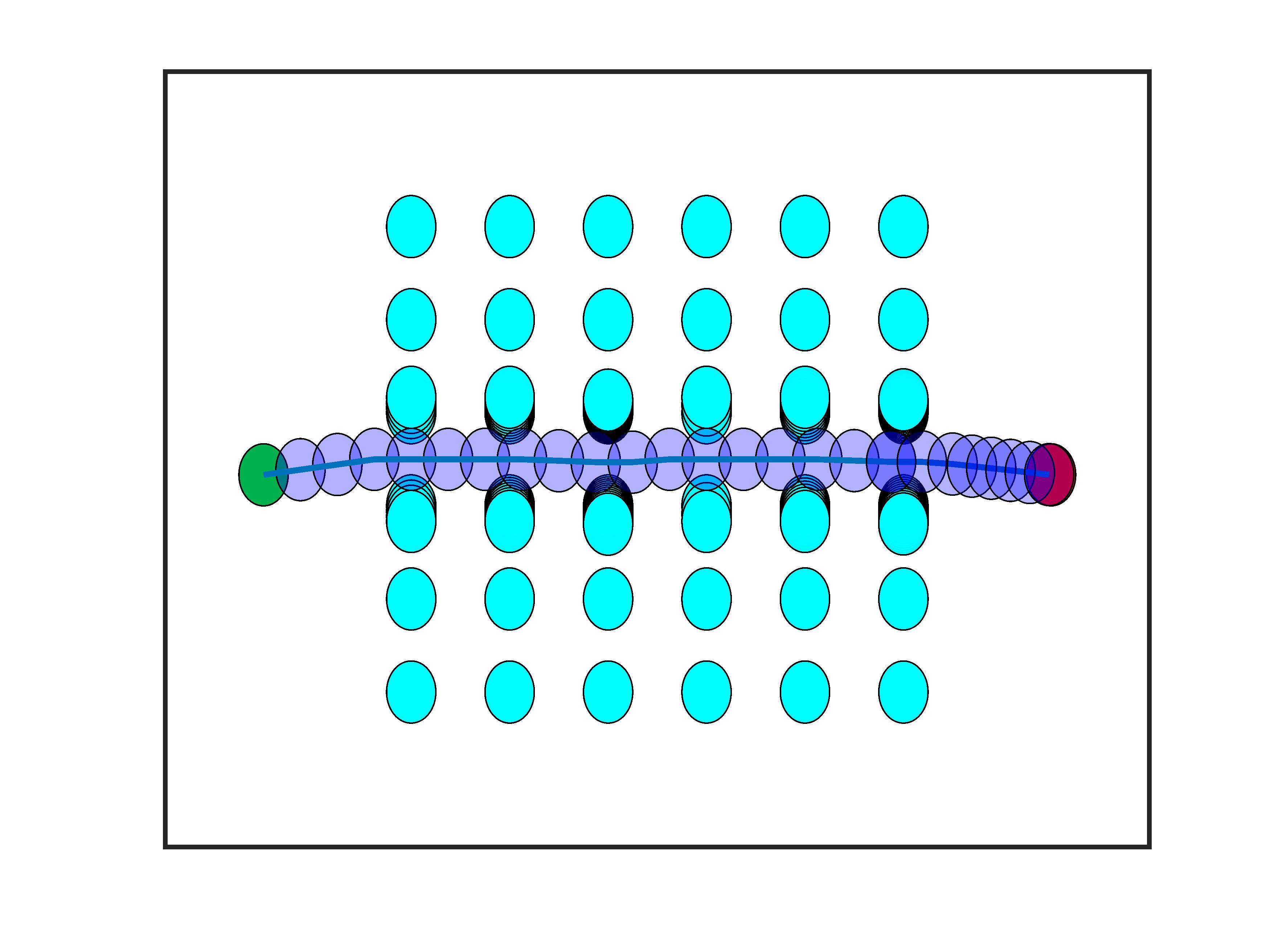}\label{fig:C13}}
\subfloat{\vspace{-10pt}\includegraphics[trim=480 350 360 270,clip,width=3.5cm,height=2.45cm]{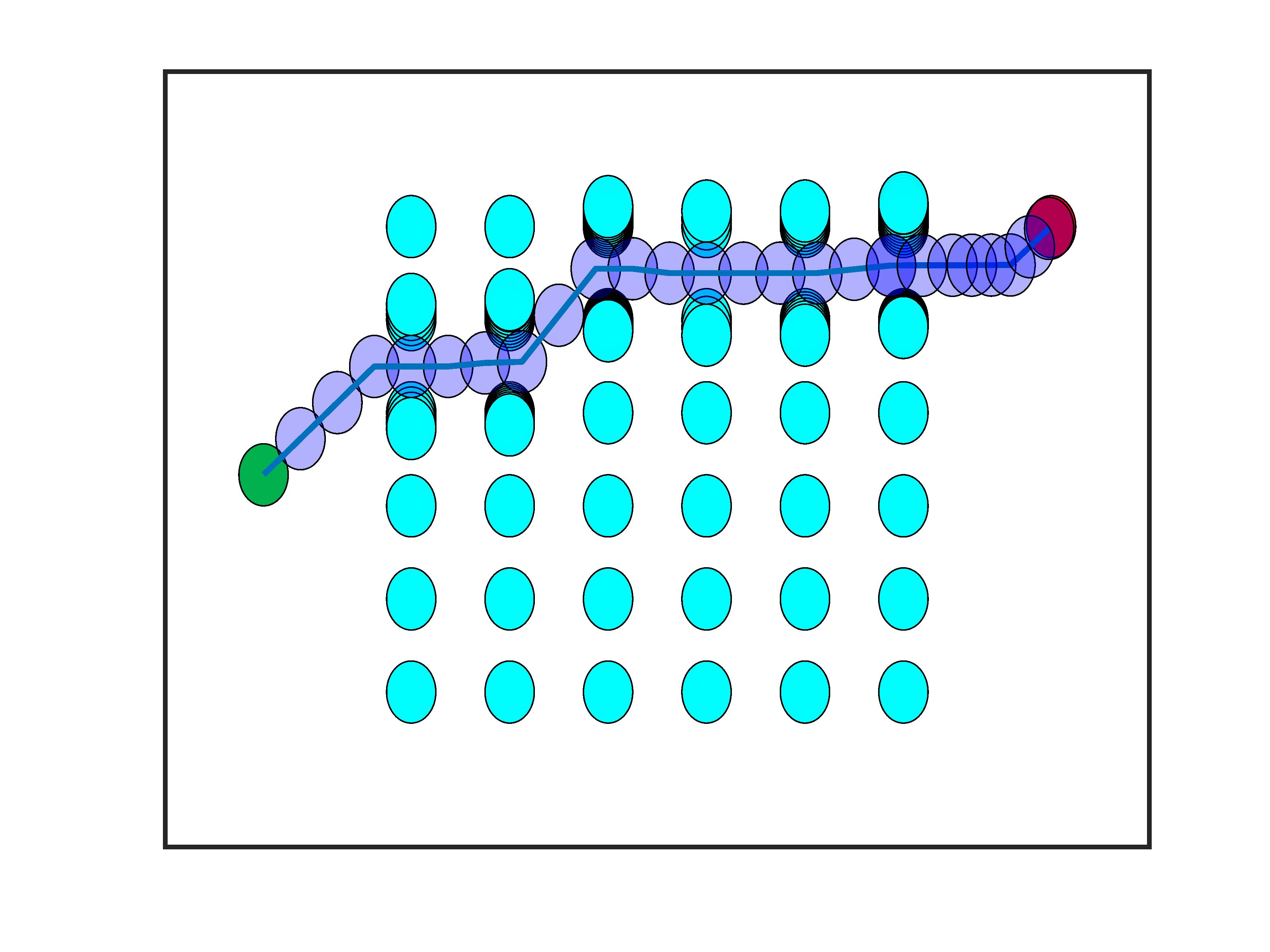}\label{fig:C23}}
     
\subfloat{\vspace{-10pt}\includegraphics[trim=84 49 57 67,clip,width=3.5cm,height=2.4cm]{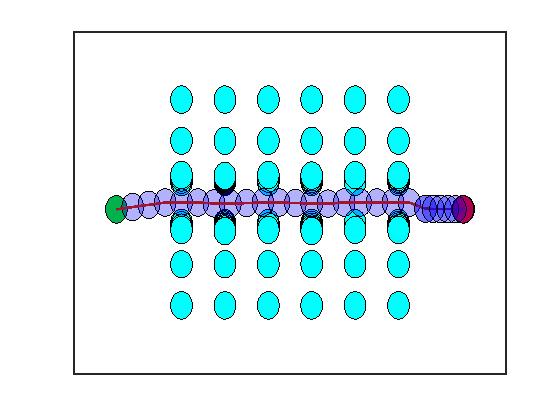}\label{fig:C14}}
\subfloat{\vspace{-10pt}\includegraphics[trim=480 350 360 270,clip,width=3.5cm,height=2.45cm]{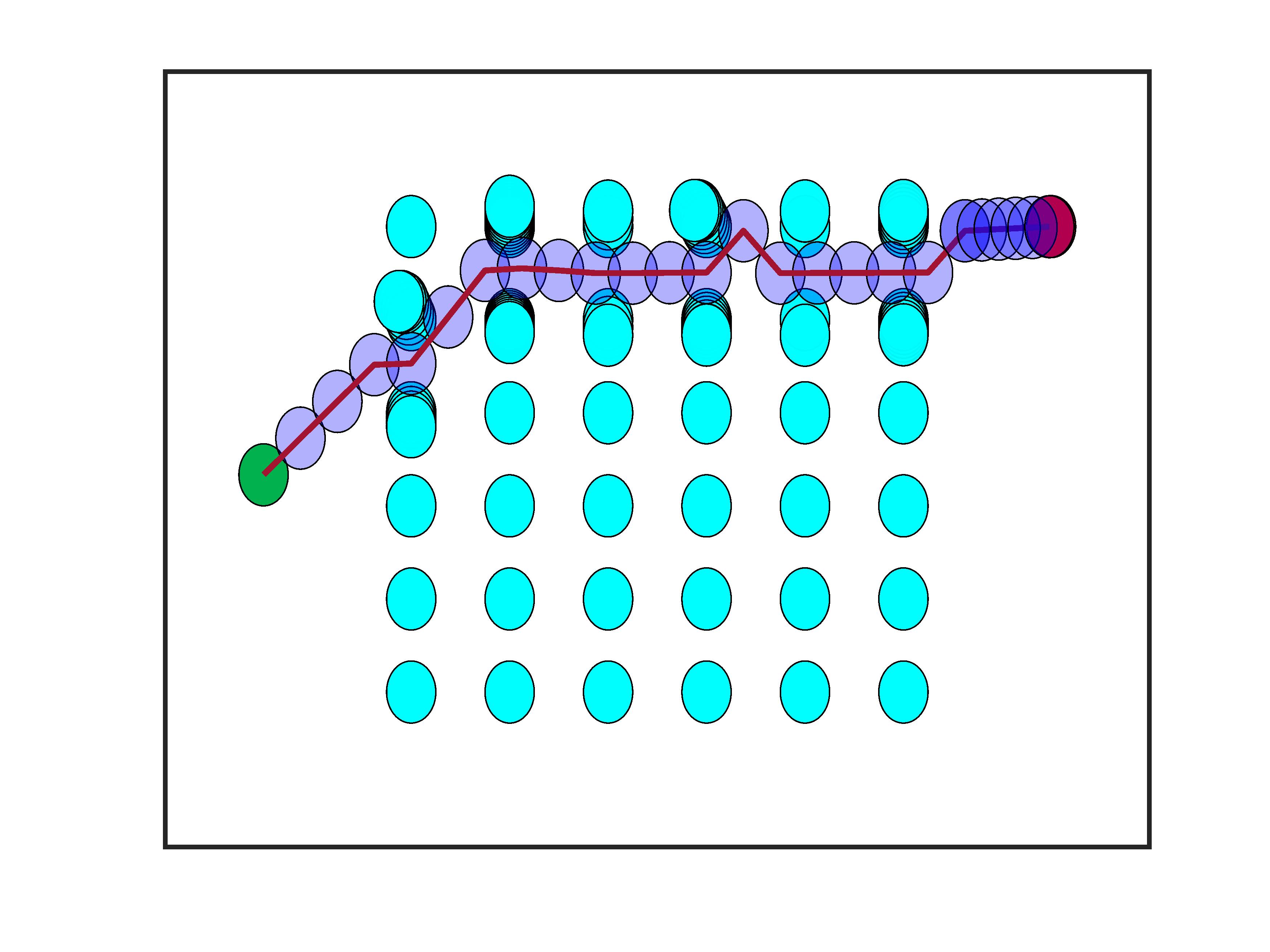}\label{fig:C24}}
    
\clearsubcaptcounter
\subfloat[\textsf{SQUARE\_1}]{\vspace{-10pt}\includegraphics[trim=500 315 380 310,clip,width=3.5cm,height=2.4cm]{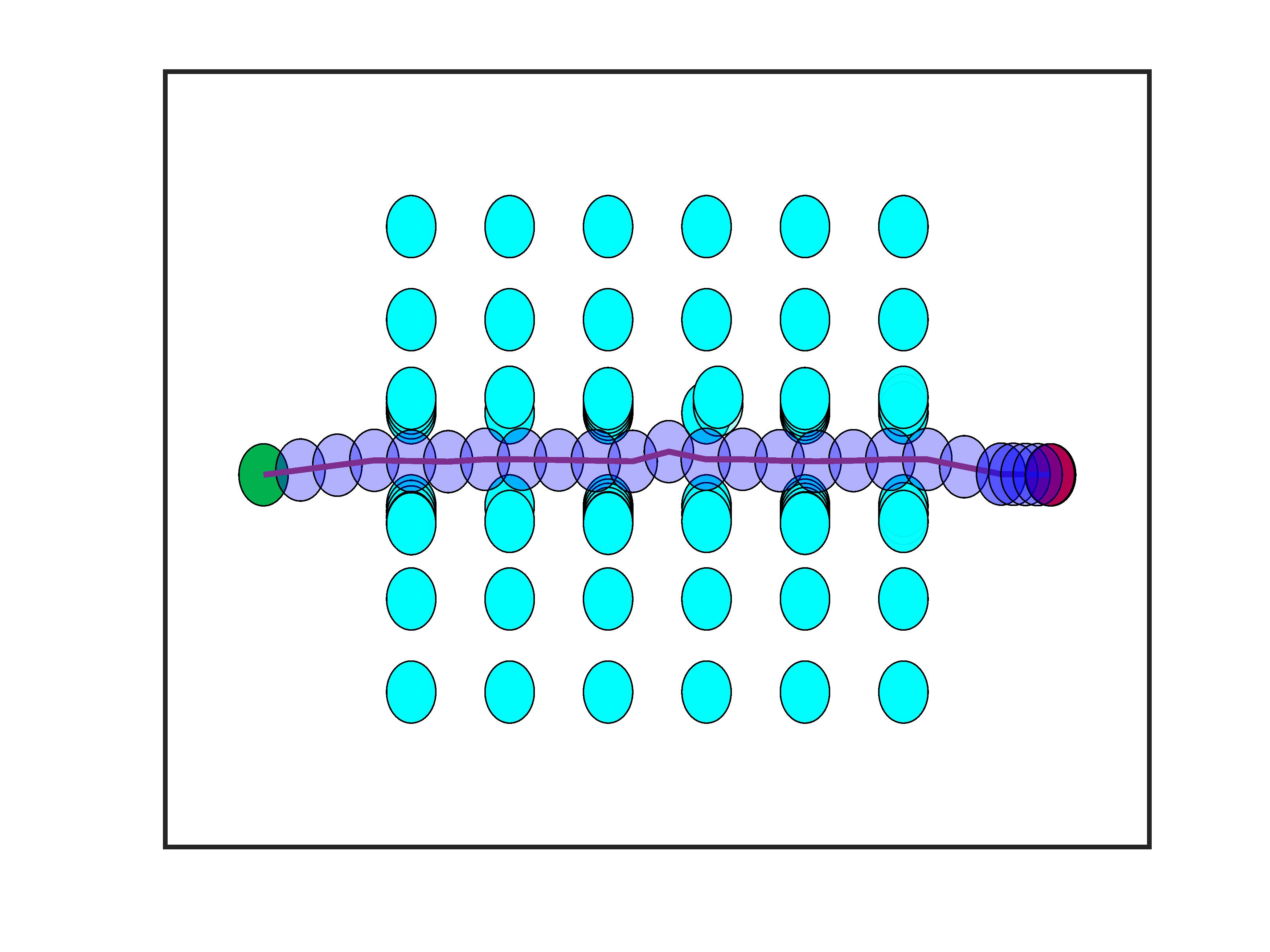}\label{fig:C15}}
\subfloat[\textsf{SQUARE\_2}]{\vspace{-10pt}\includegraphics[trim=480 350 360 270,clip,width=3.5cm,height=2.45cm]{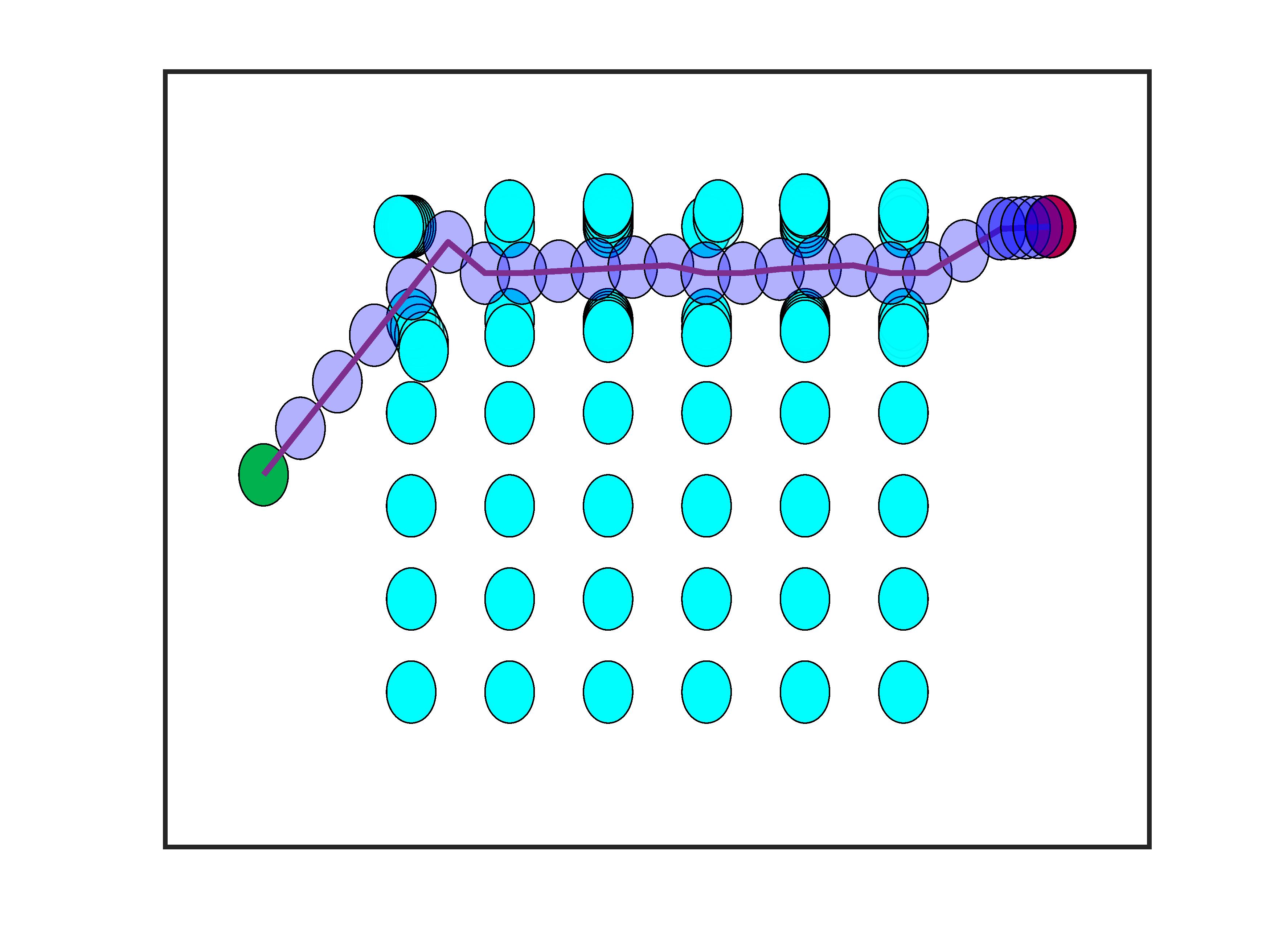}\label{fig:C25}}
\caption{
Robot trajectories and obstacle displacements for \textsf{SQUARE\_1} and \textsf{SQUARE\_2} while the robot moves according to the model in~\eqref{eq:robot_model1}. 
The rows are related to different values for $m$, the top row being related to $m=1$ (exact solution), the second row to $m=2$, and so on.}
\label{fig:linearmodel1a}
\end{figure*}

\begin{figure*}[]
\subfloat{\includegraphics[trim=61 125 50 115,clip,width=6.8cm,height=1.8cm]{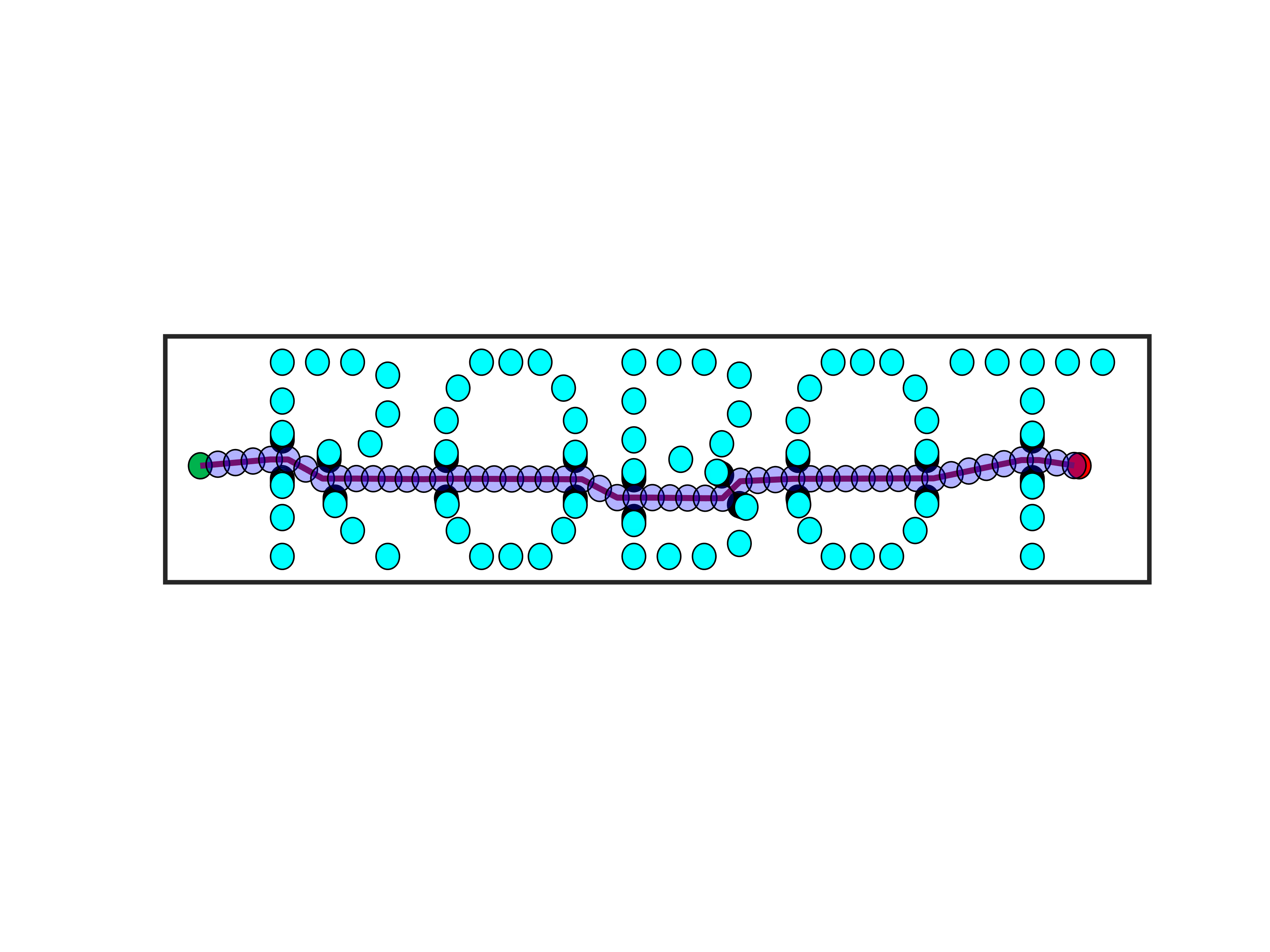}\label{fig:C31}}
\subfloat{\includegraphics[trim=475 330 360 220,clip,width=2.4cm,height=1.85cm]{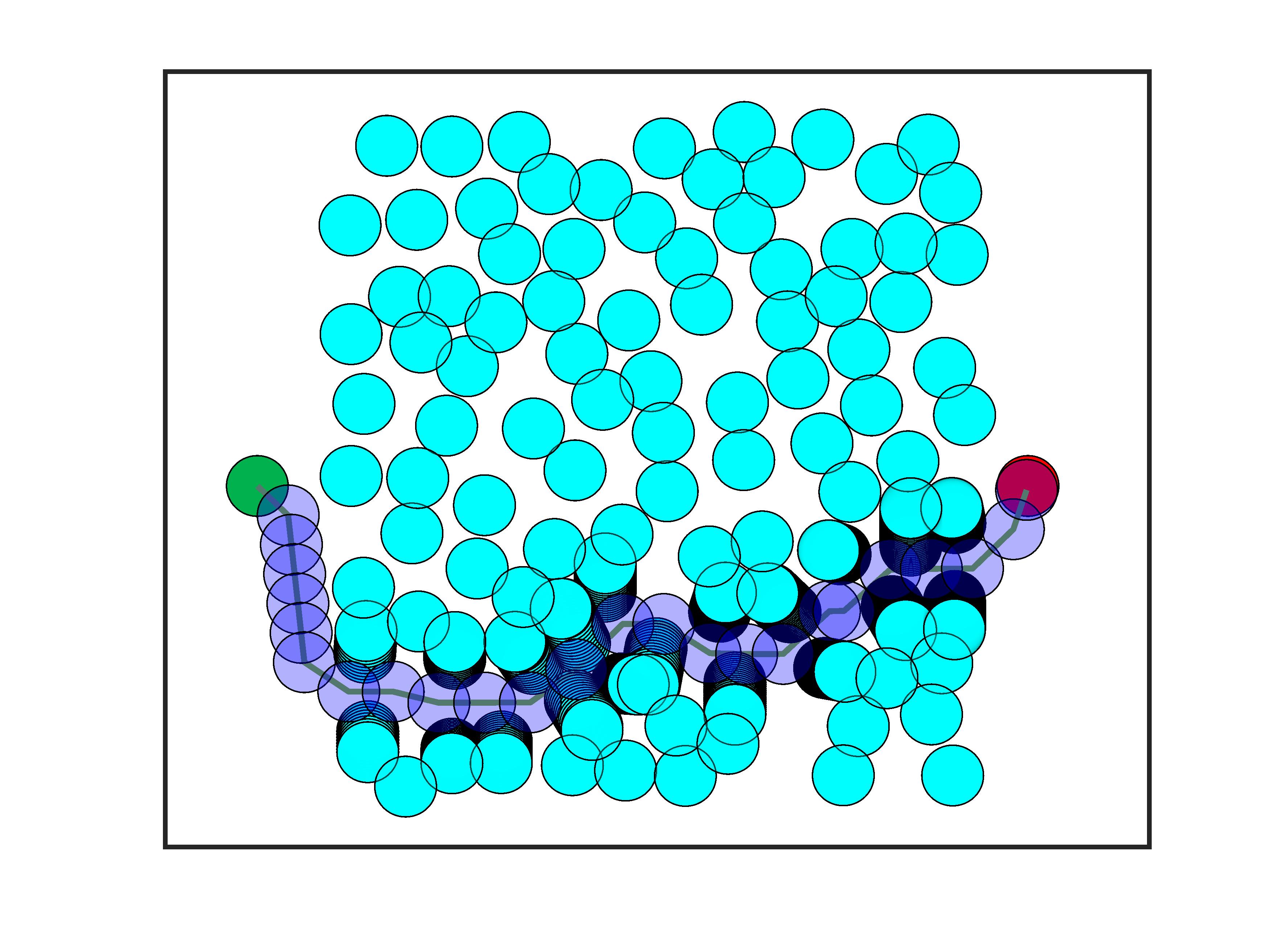}\label{fig:C41}}\\
    
\subfloat{\includegraphics[trim=61 125 50 115,clip,width=6.8cm,height=1.8cm]{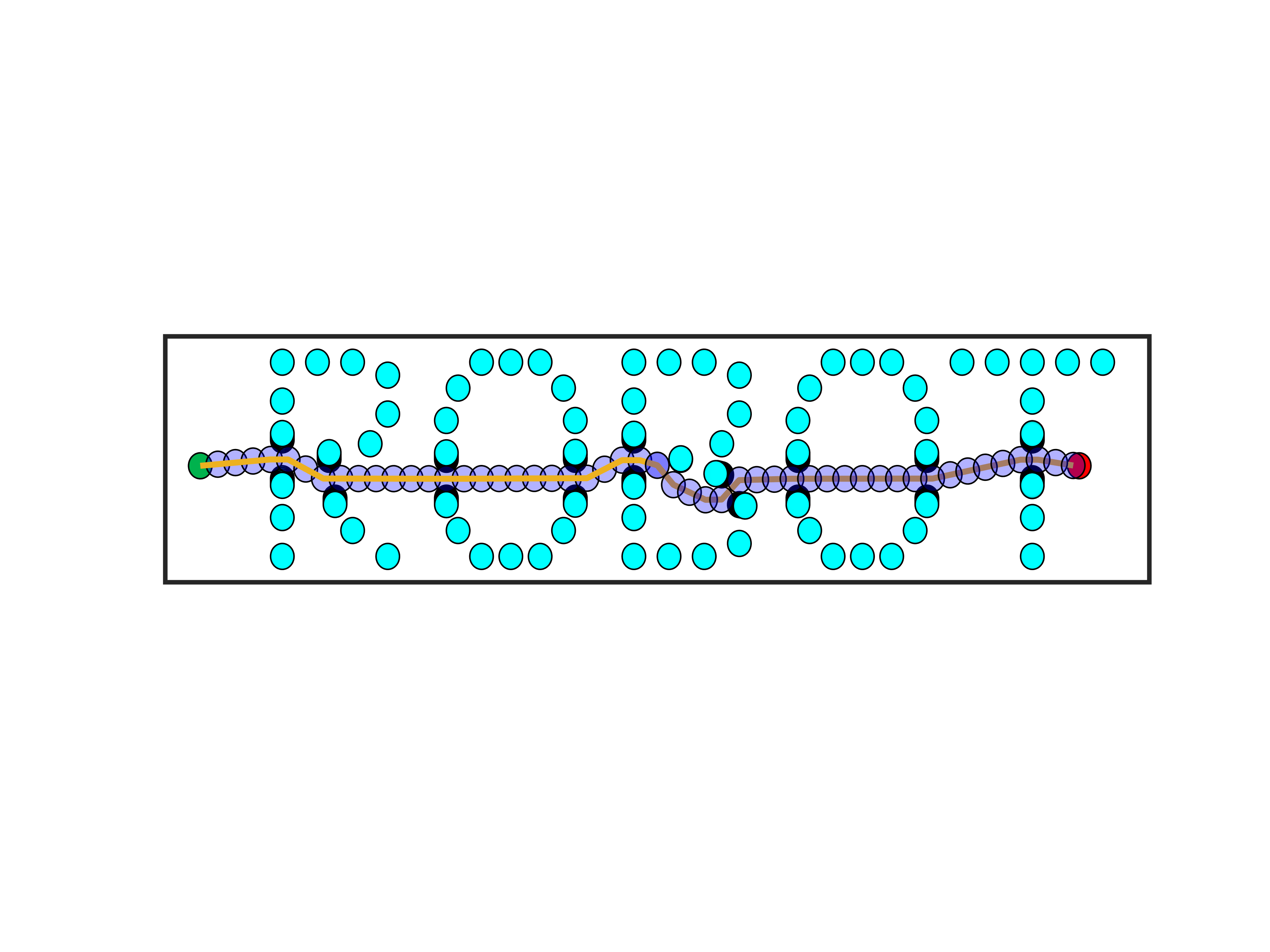}\label{fig:C32}}
\subfloat{\includegraphics[trim=475 330 360 220,clip,width=2.4cm,height=1.85cm]{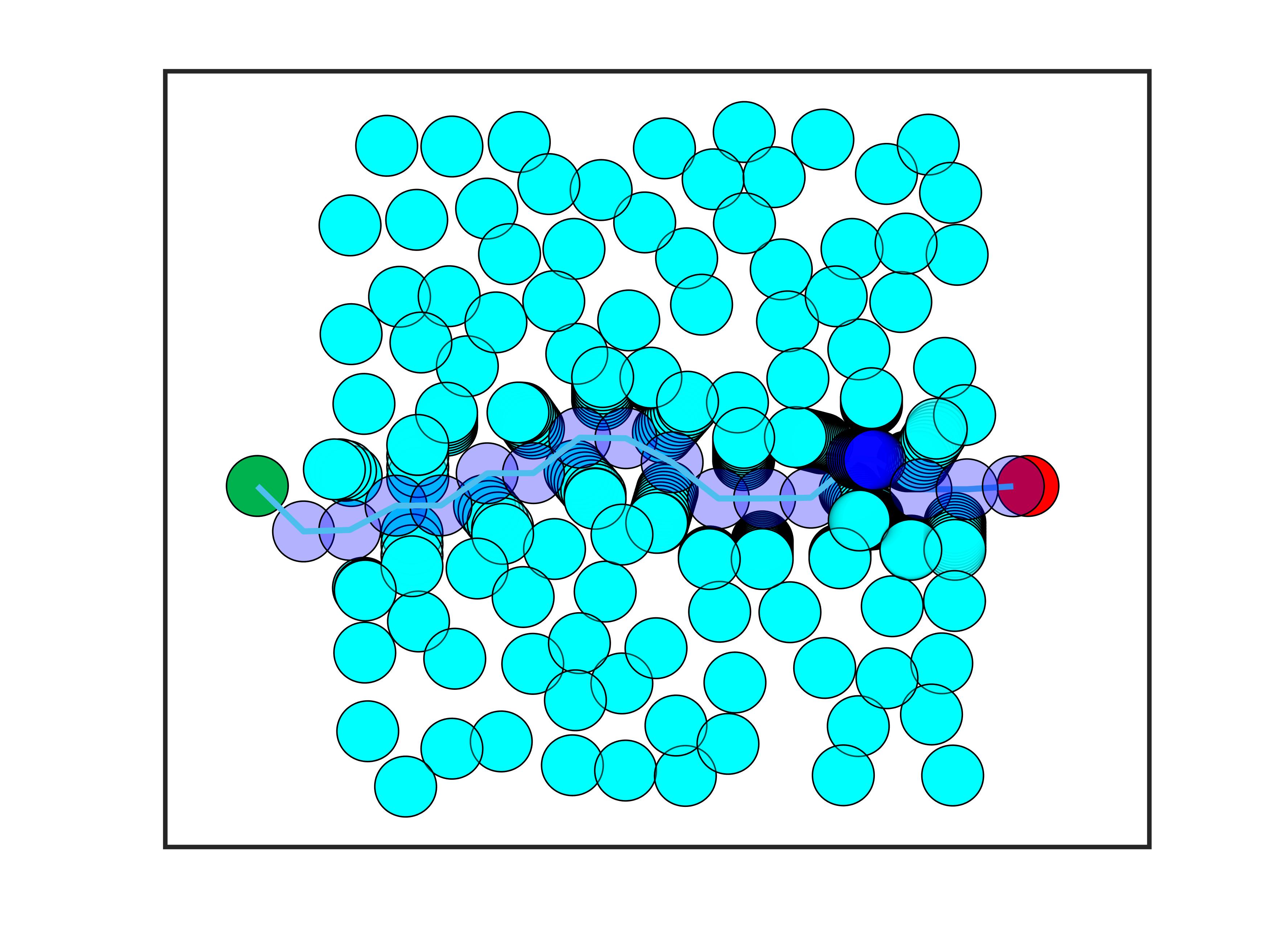}\label{fig:C42}}\\
     
\subfloat{\includegraphics[trim=61 125 50 115,clip,width=6.8cm,height=1.8cm]{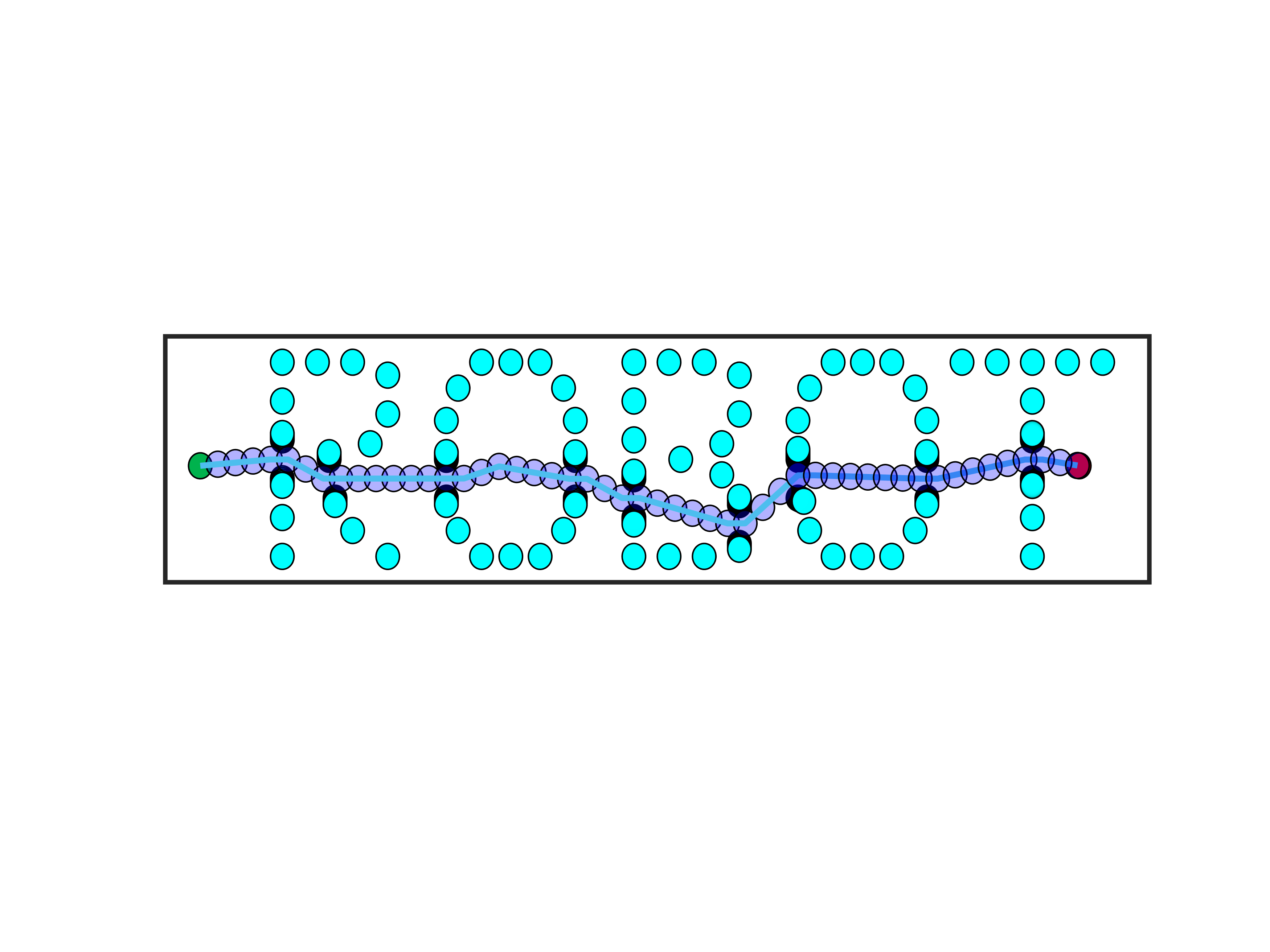}\label{fig:C33}}
\subfloat{\includegraphics[trim=475 330 360 220,clip,width=2.4cm,height=1.85cm]{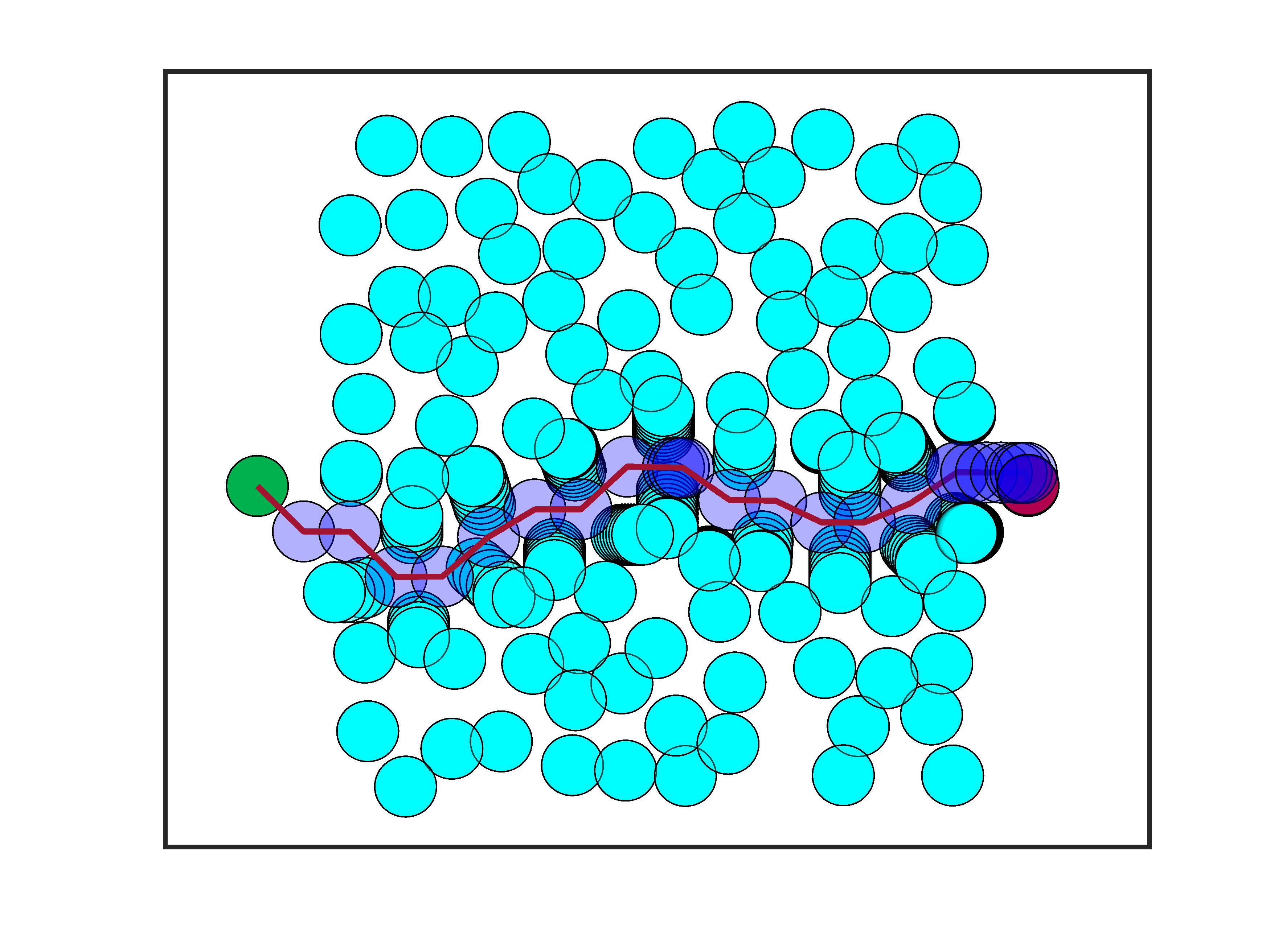}\label{fig:C43}}\\
     
\subfloat{\includegraphics[trim=61 125 50 115,clip,width=6.8cm,height=1.8cm]{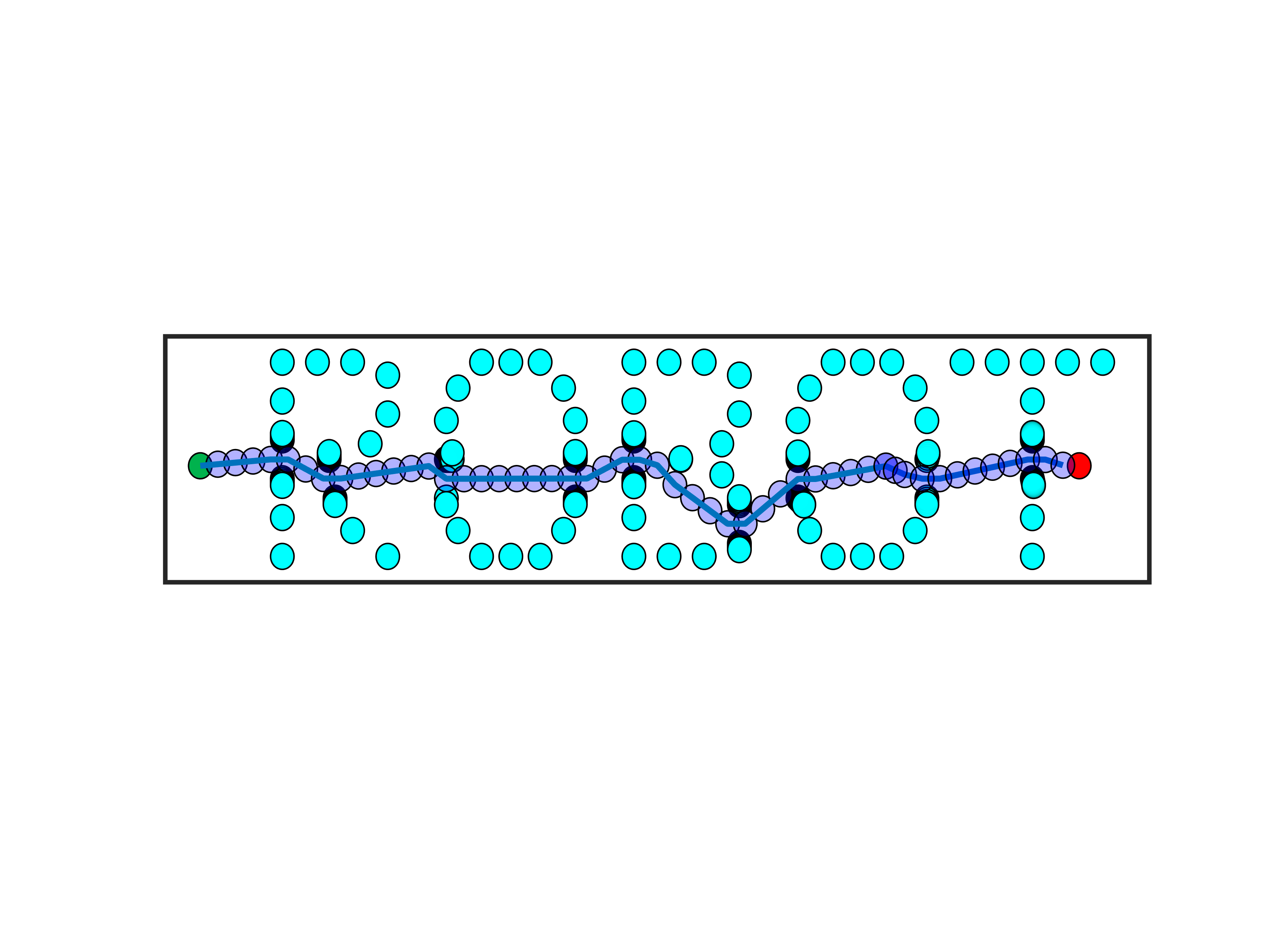}\label{fig:C34}}
\subfloat{\includegraphics[trim=475 330 360 220,clip,width=2.4cm,height=1.85cm]{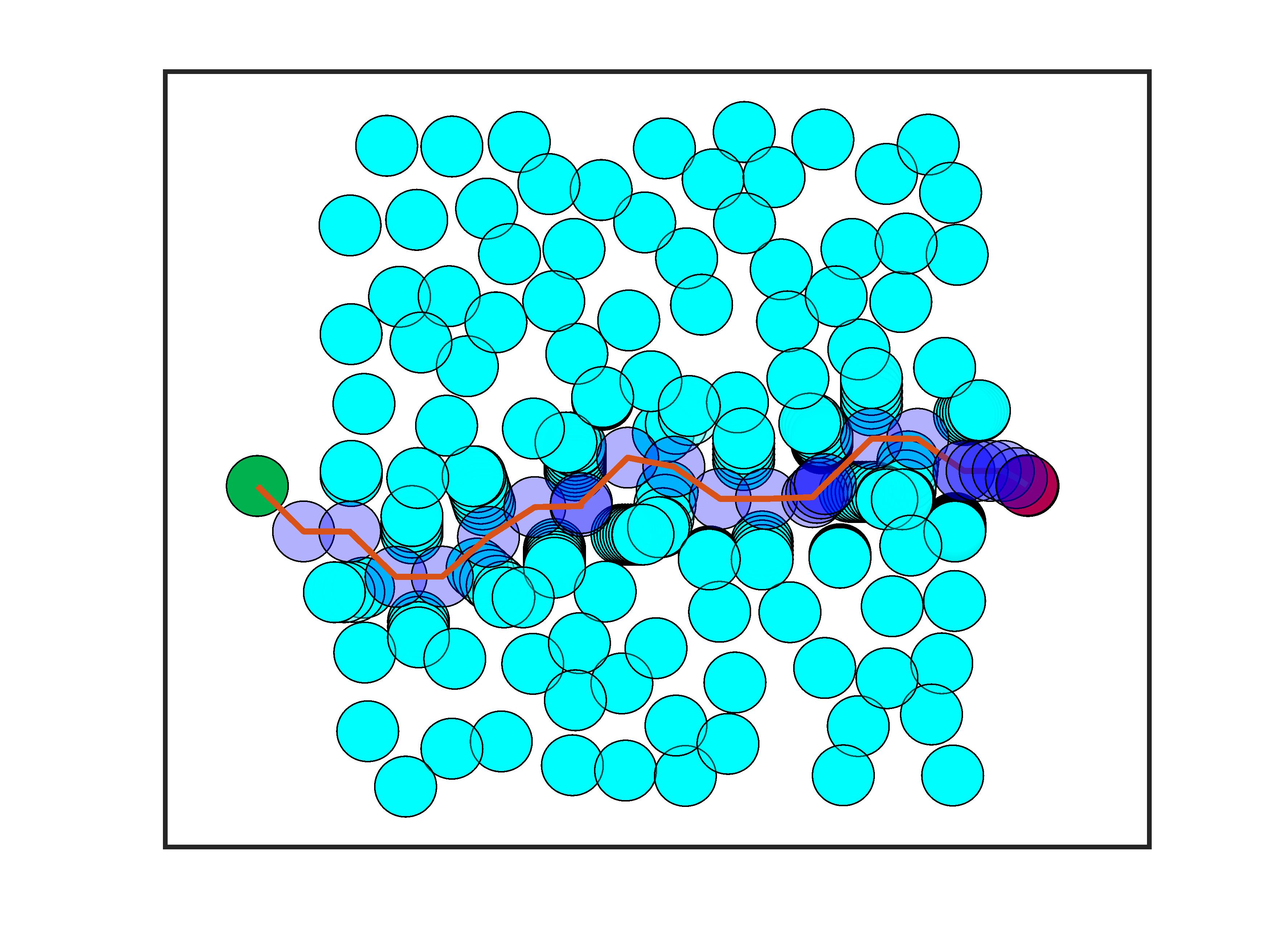}\label{fig:C44}}\\
    
\clearsubcaptcounter
\subfloat[\textsf{ROBOT}]{\includegraphics[trim=61 125 50 115,clip,width=6.8cm,height=1.8cm]{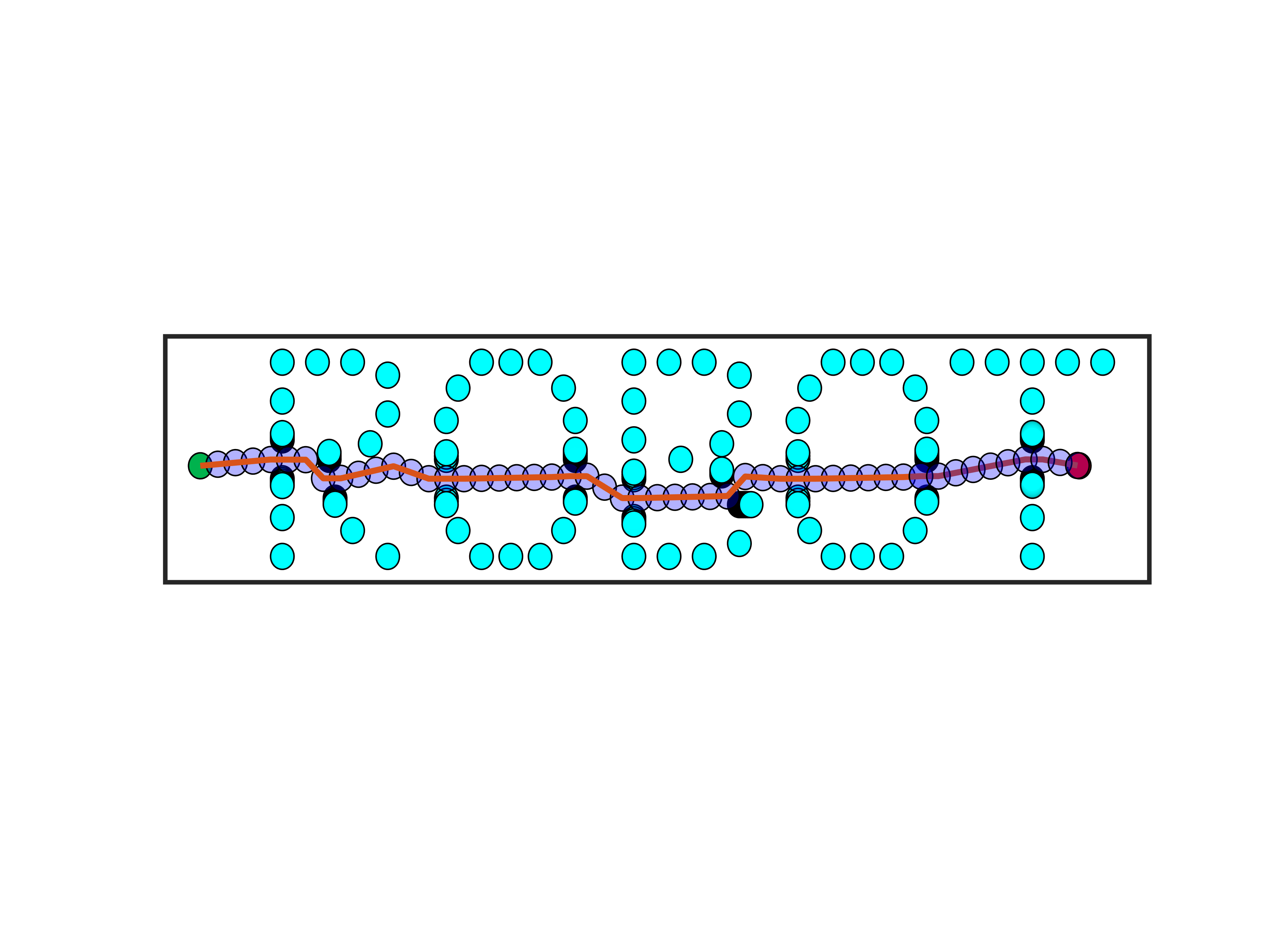}\label{fig:C35}}
\subfloat[\textsf{RANDOM}]{\includegraphics[trim=475 330 360 220,clip,width=2.4cm,height=1.85cm]{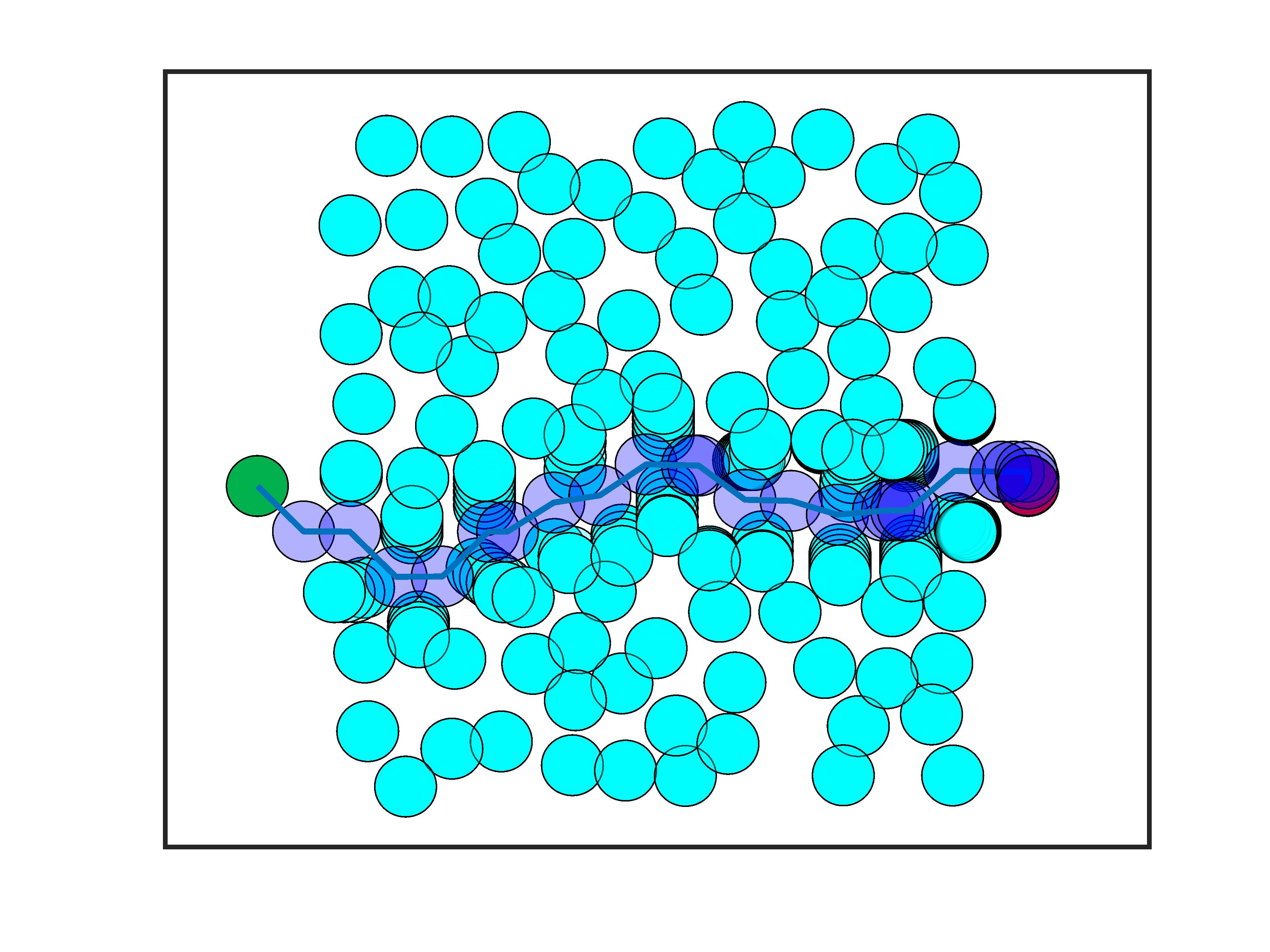}\label{fig:C45}}
\caption{
Robot trajectories and obstacle displacements for \textsf{ROBOT} and \textsf{RANDOM} while the robot moves according to the model in~\eqref{eq:robot_model1}. 
The rows are related to different values for $m$, the top row being related to $m=1$ (exact solution), the second row to $m=2$, and so on.}
\label{fig:linearmodel1b}
\end{figure*}

\begin{figure*}[]
\subfloat{\vspace{-7pt}\includegraphics[trim=500 315 380 310,clip,width=3.5cm,height=2.2cm]{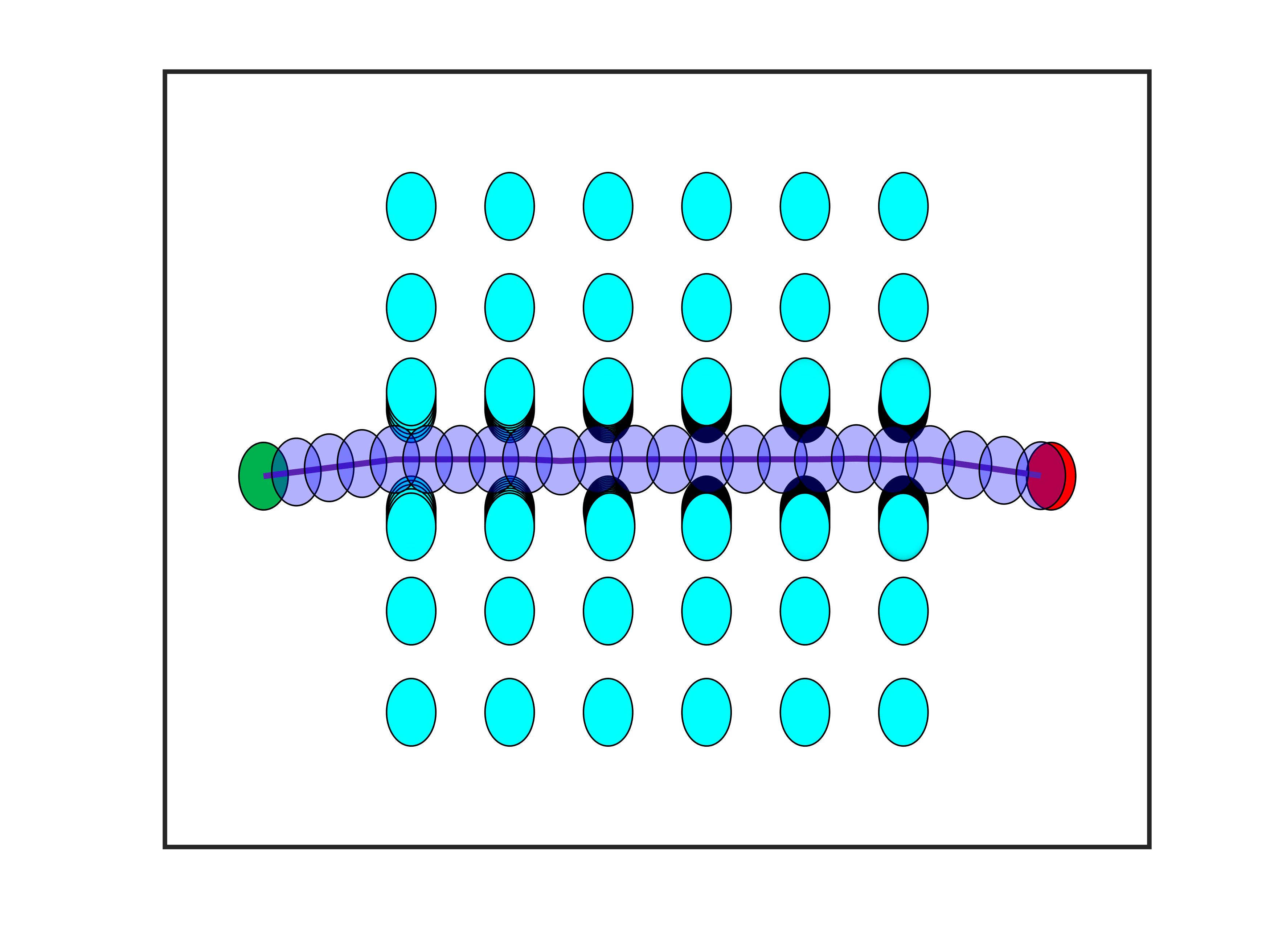}\label{fig:C51}}
\subfloat{\vspace{-10pt}\includegraphics[trim=500 315 380 310,clip,width=3.5cm,height=2.4cm]{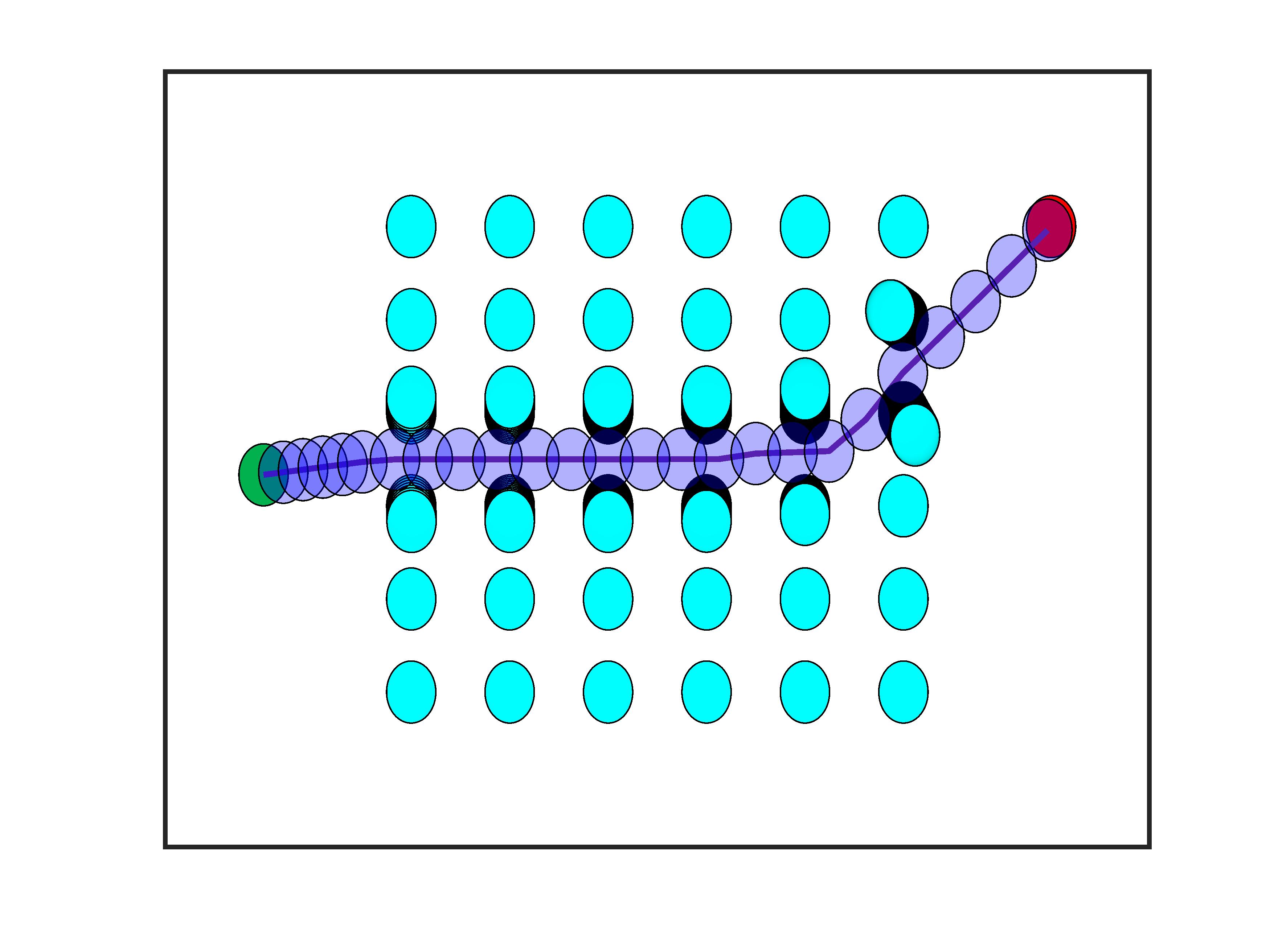}\label{fig:C61}}
    
\subfloat{\vspace{-7pt}\includegraphics[trim=500 315 380 310,clip,width=3.5cm,height=2.2cm]{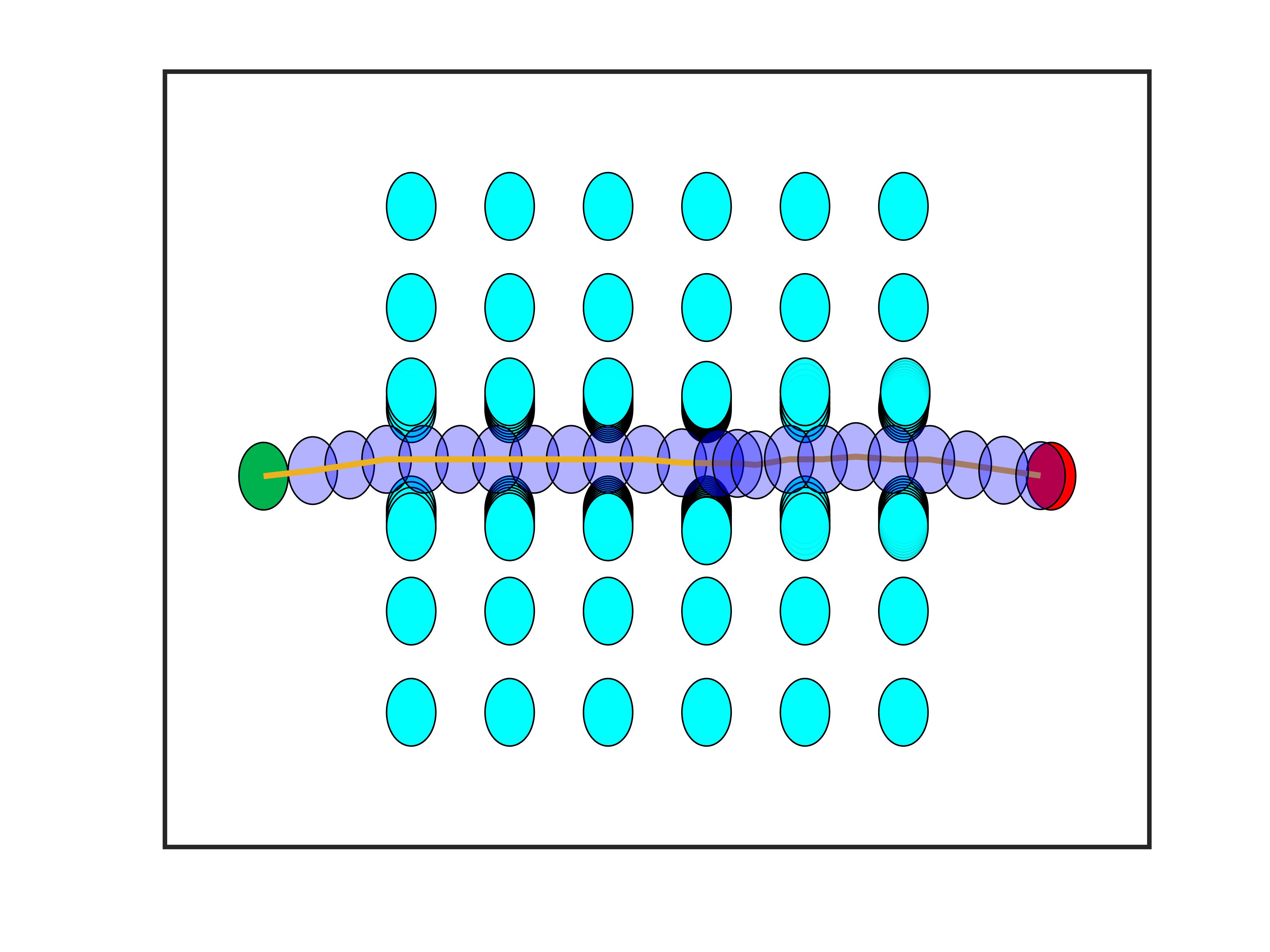}\label{fig:C52}}
\subfloat{\vspace{-10pt}\includegraphics[trim=480 330 360 270,clip,width=3.5cm,height=2.4cm]{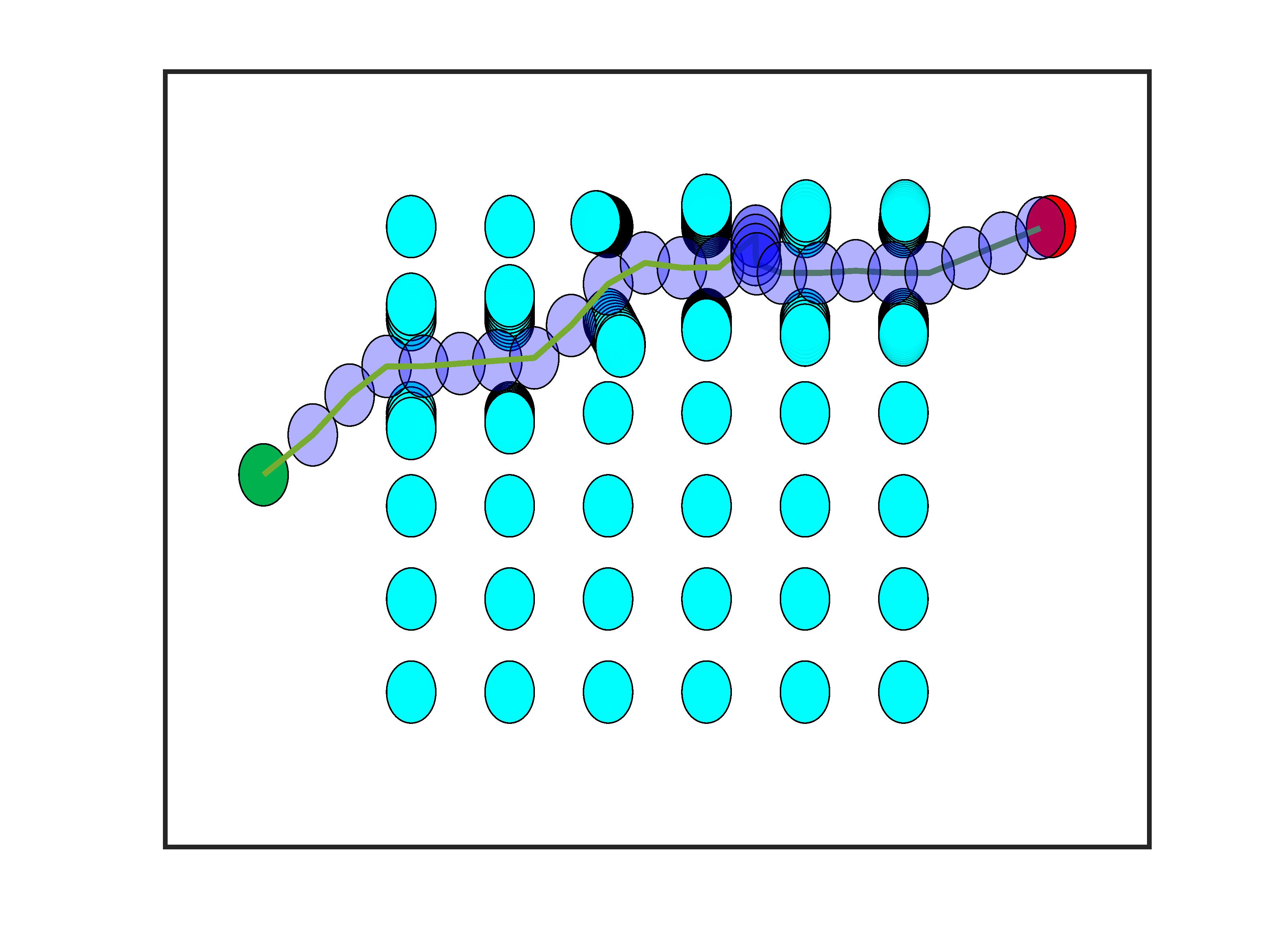}\label{fig:C62}}
     
\subfloat{\vspace{-8pt}\includegraphics[trim=500 315 380 310,clip,width=3.5cm,height=2.2cm]{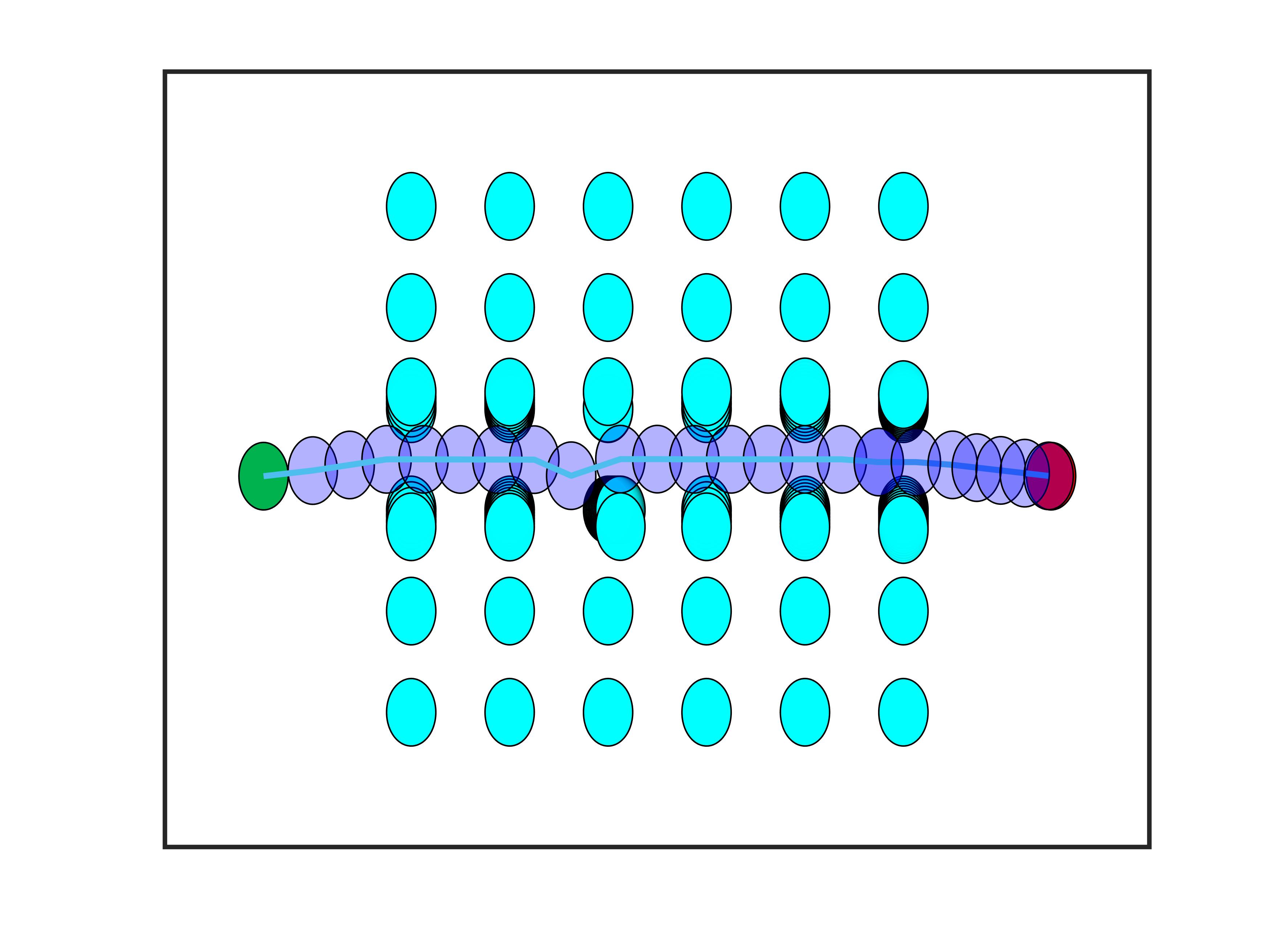}\label{fig:C53}}
\subfloat{\vspace{-10pt}\includegraphics[trim=480 330 360 270,clip,width=3.5cm,height=2.4cm]{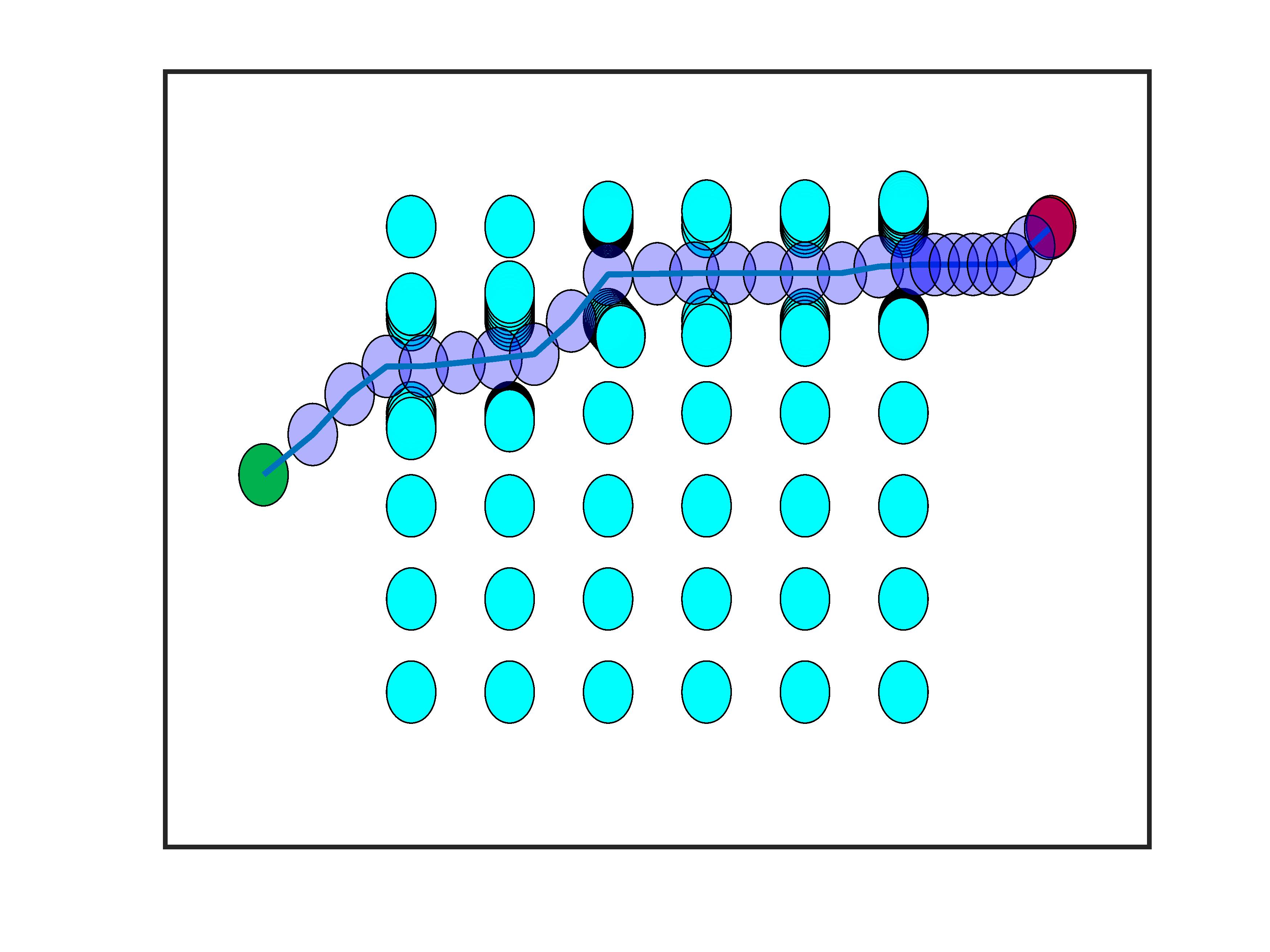}\label{fig:C63}}
     
\subfloat{\vspace{-8pt}\includegraphics[trim=500 315 380 310,clip,width=3.5cm,height=2.2cm]{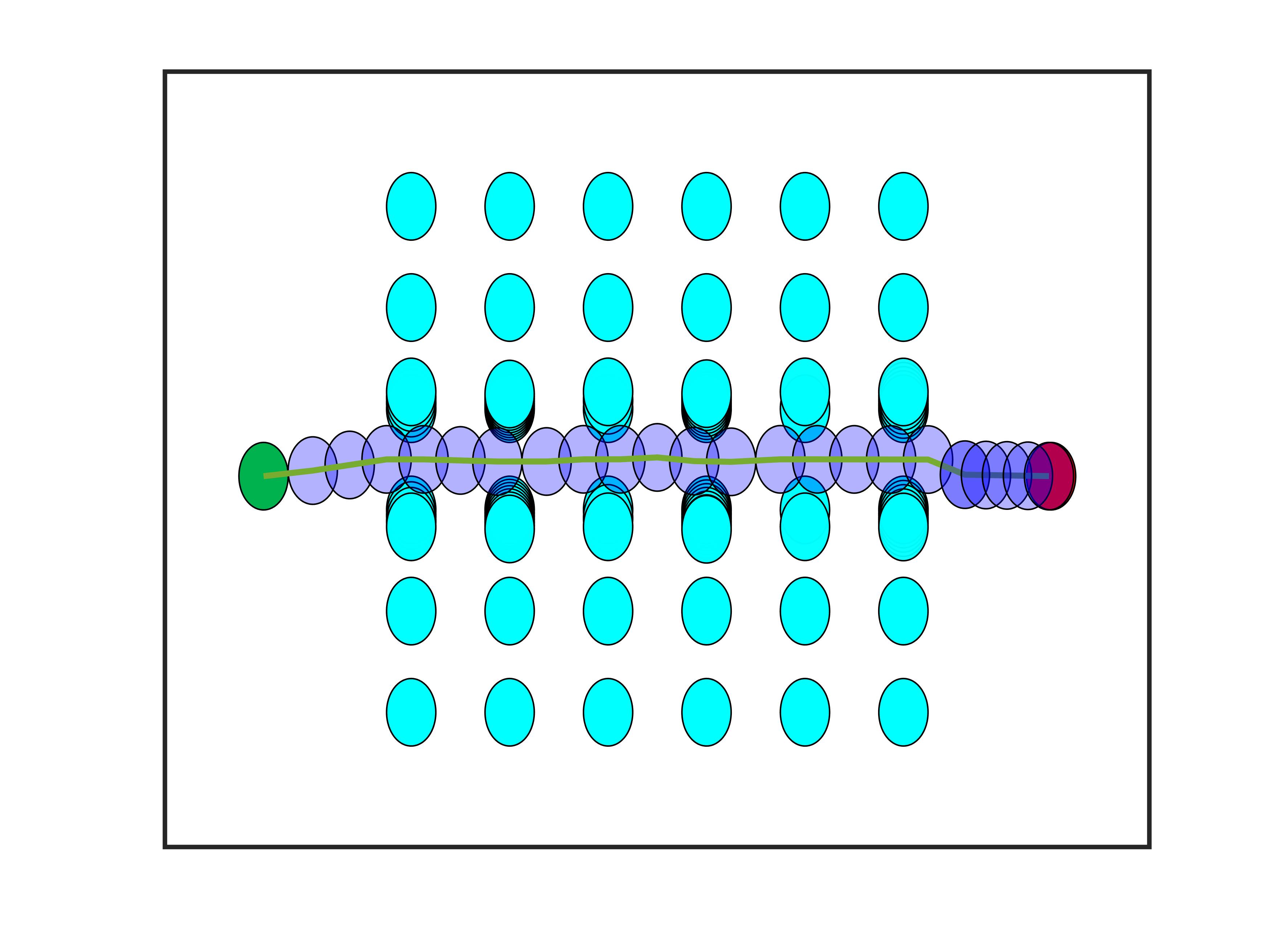}\label{fig:C54}}
\subfloat{\vspace{-10pt}\includegraphics[trim=480 330 360 270,clip,width=3.5cm,height=2.4cm]{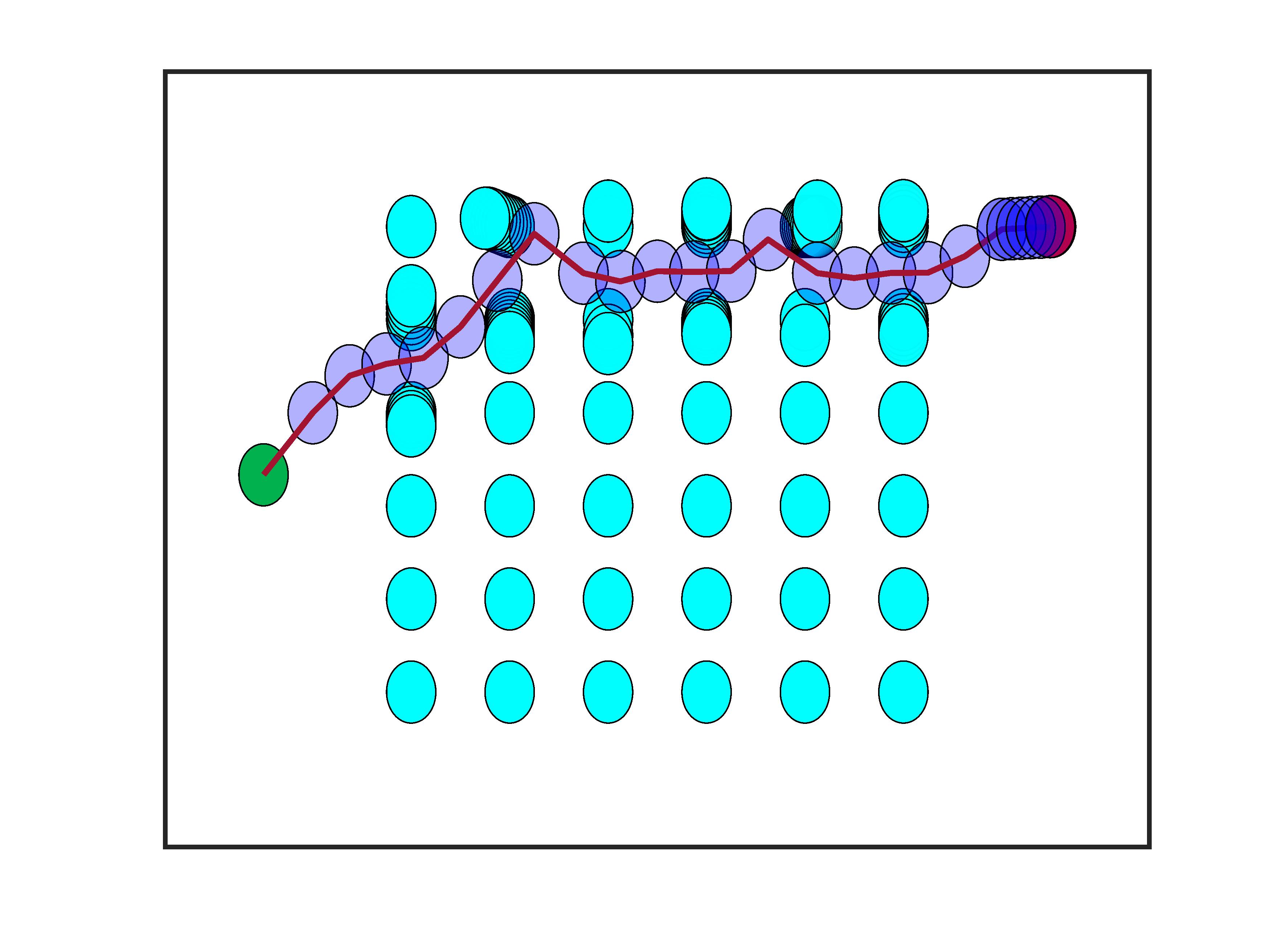}\label{fig:C64}}
    
\clearsubcaptcounter
\subfloat[\textsf{SQUARE\_1}]{\vspace{-8pt}\includegraphics[trim=500 315 380 310,clip,width=3.5cm,height=2.2cm]{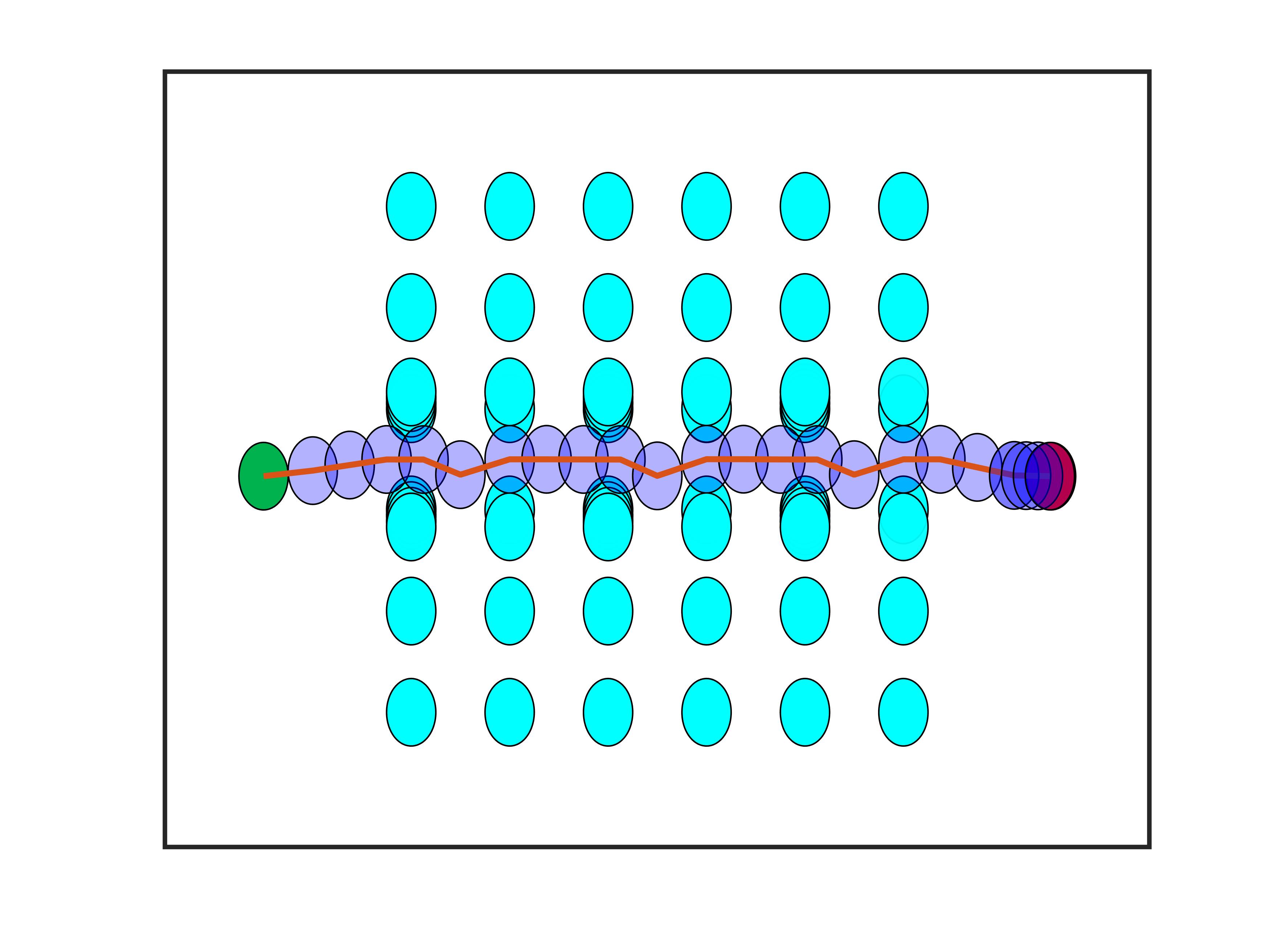}\label{fig:C55}}
\subfloat[\textsf{SQUARE\_2}]{\vspace{-10pt}\includegraphics[trim=480 330 360 270,clip,width=3.5cm,height=2.4cm]{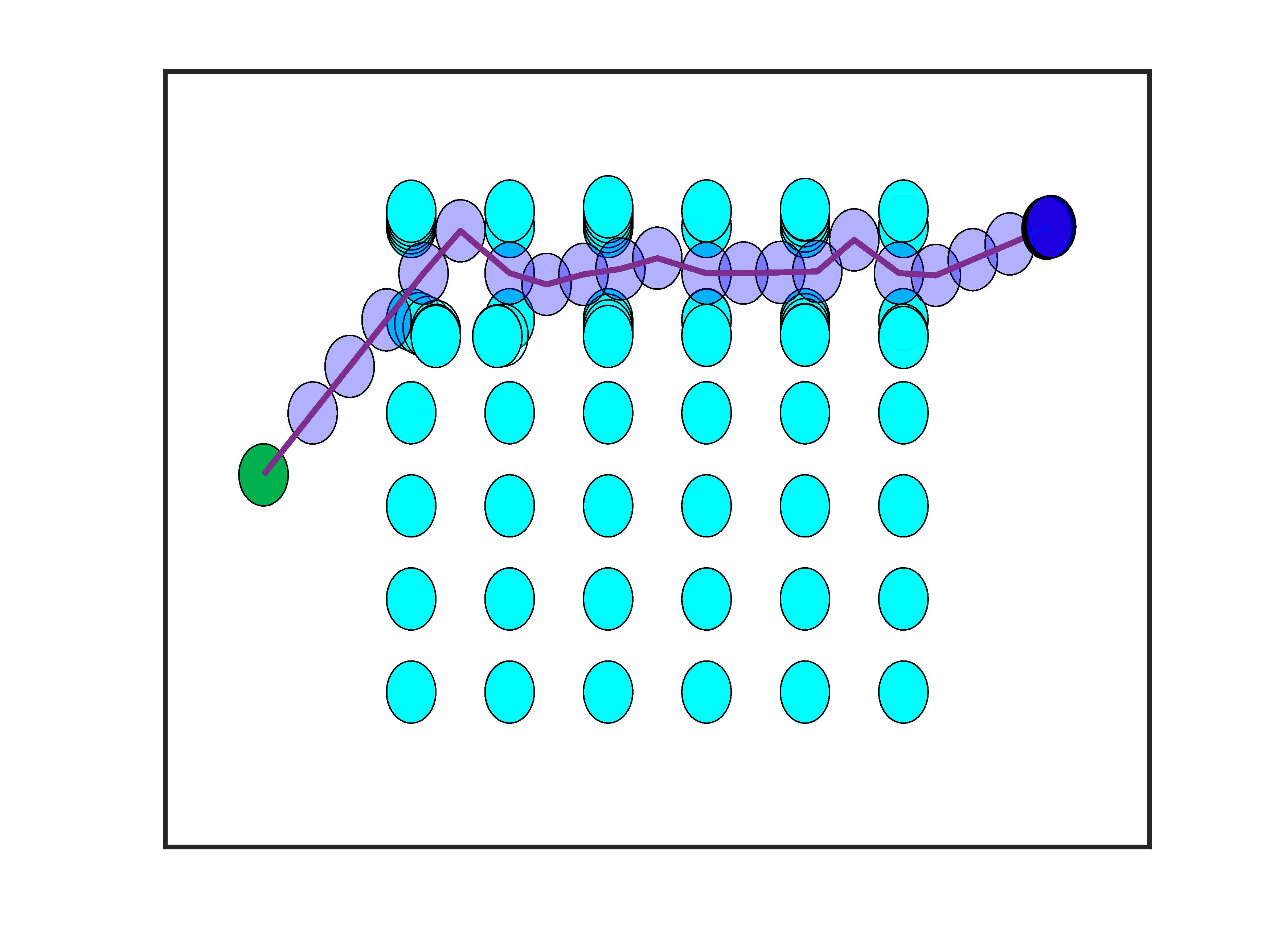}\label{fig:C65}}
\caption{
Robot trajectories and obstacle displacements for \textsf{SQUARE\_1} and \textsf{SQUARE\_2} while the robot moves according to the model in~\eqref{eq:robot_model2}. 
The first row corresponds to $m=1$ (no horizon slicing) for all four domains, the second row corresponds to $m=2$, and so on.
}
\label{fig:linearmodel2a}
\end{figure*}

\begin{figure*}[]
\subfloat{\includegraphics[trim=61 125 50 115,clip,width=6.8cm,height=1.8cm]{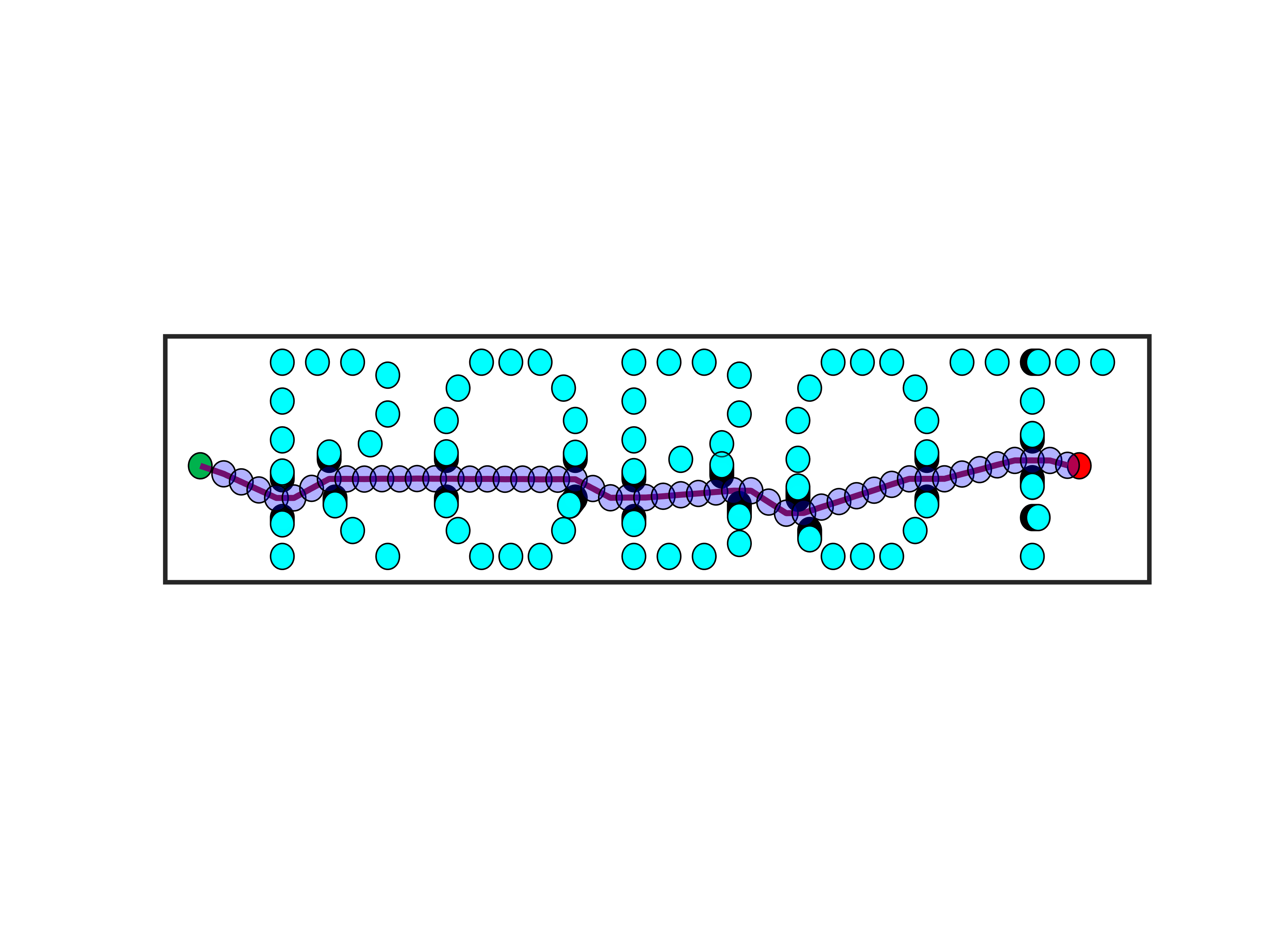}\label{fig:C71}}
\subfloat{\includegraphics[trim=475 330 360 220,clip,width=2.4cm,height=1.85cm]{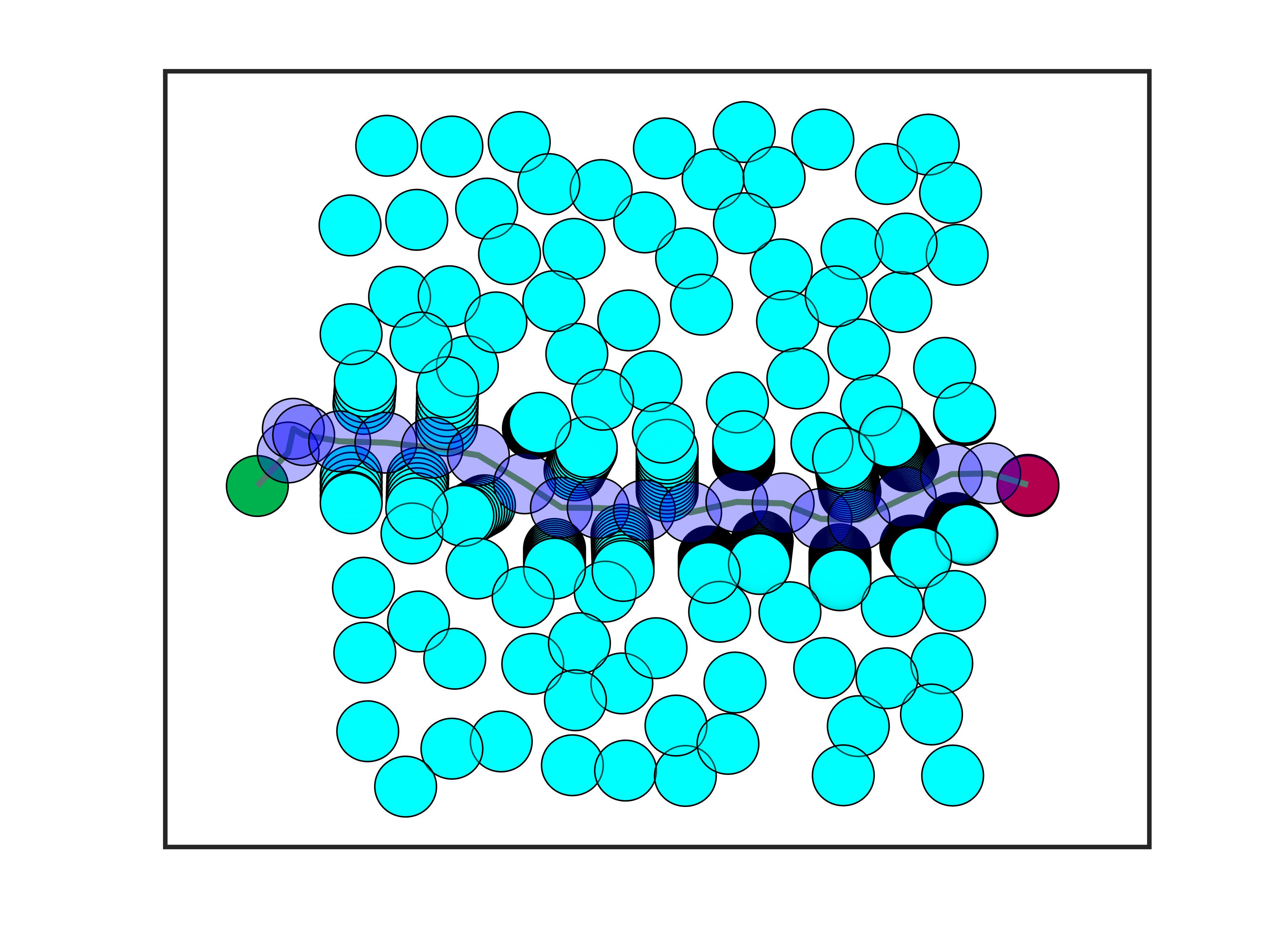}\label{fig:C81}}\\
    
\subfloat{\vspace{-2pt}\includegraphics[trim=501 325 410 315,clip,width=6.8cm,height=1.8cm]{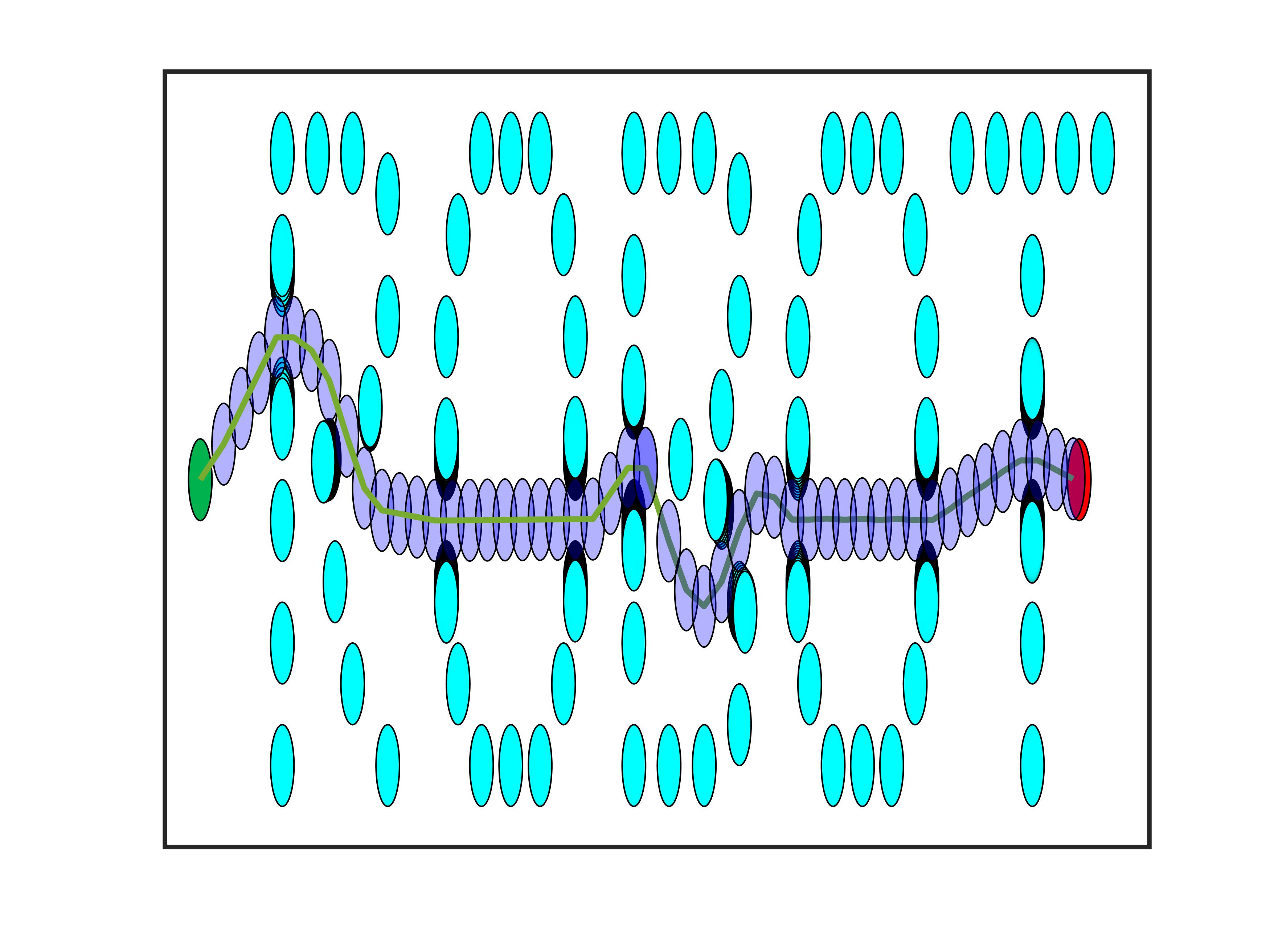}\label{fig:C72}}
\subfloat{\vspace{0pt}\includegraphics[trim=475 330 360 220,clip,width=2.4cm,height=1.85cm]{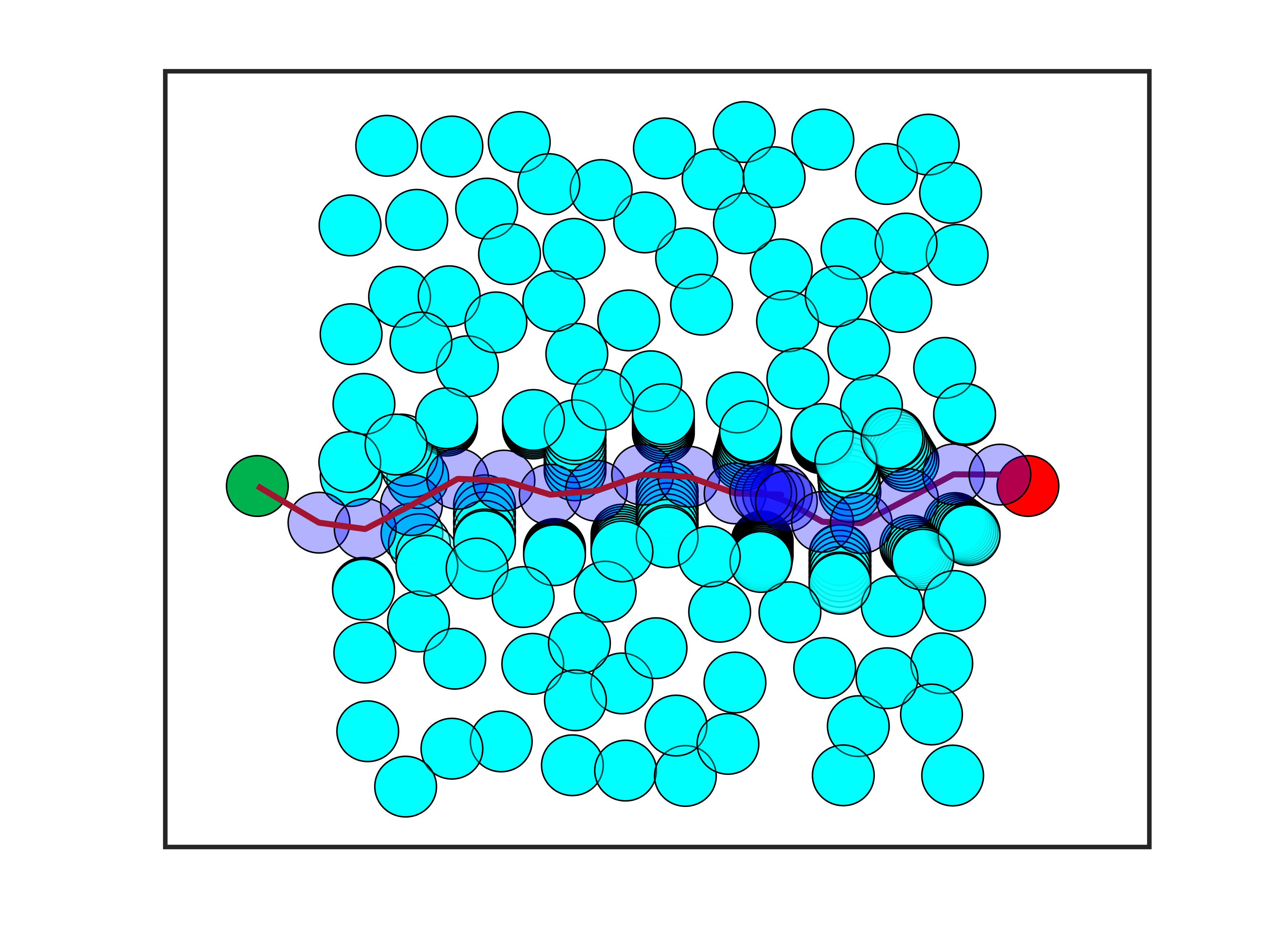}\label{fig:C82}}\\
     
\subfloat{\includegraphics[trim=61 125 50 115,clip,width=6.8cm,height=1.8cm]{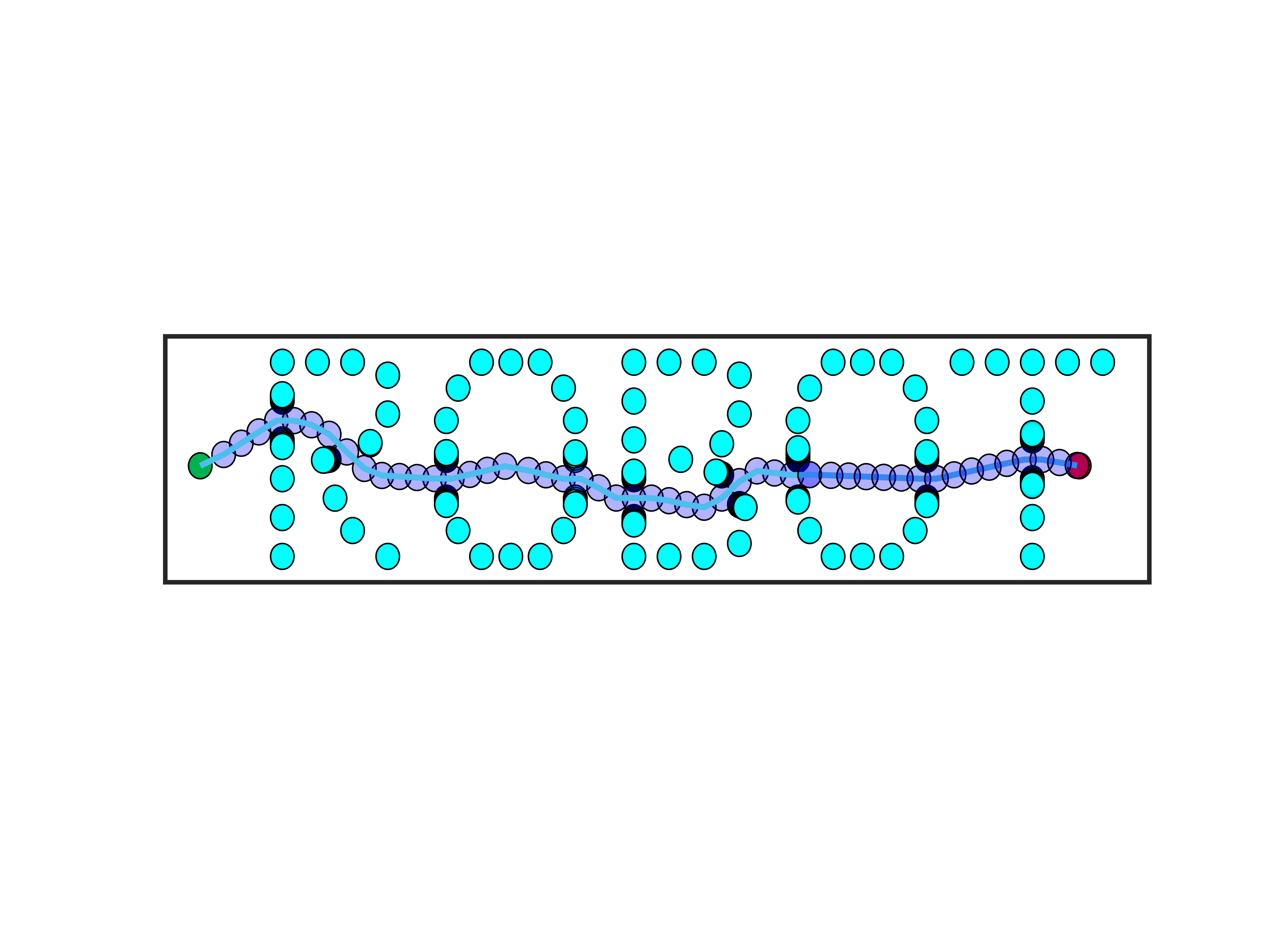}\label{fig:C73}}
\subfloat{\includegraphics[trim=475 330 360 220,clip,width=2.4cm,height=1.85cm]{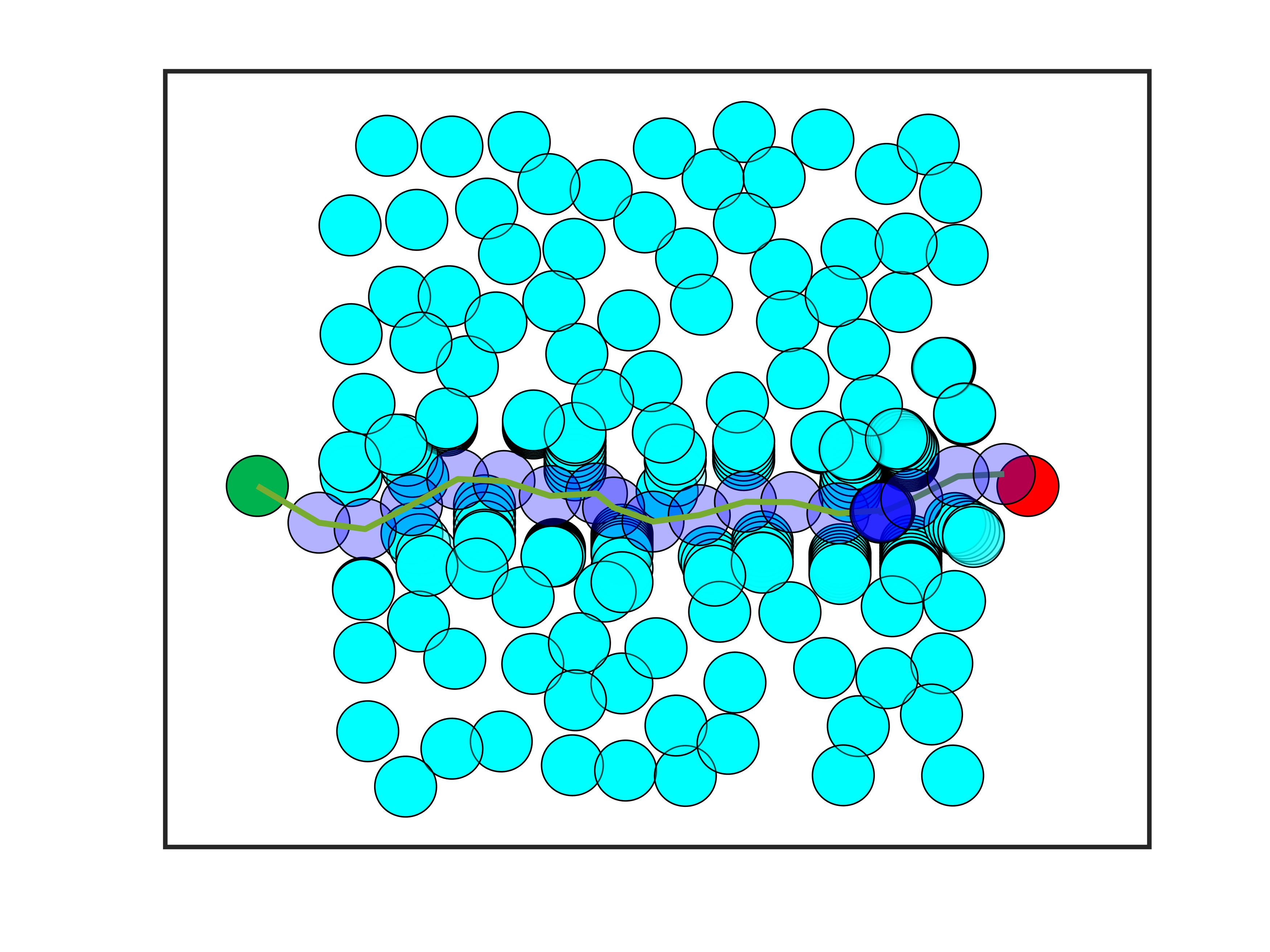}\label{fig:C83}}\\
     
\subfloat{\includegraphics[trim=61 125 50 115,clip,width=6.8cm,height=1.8cm]{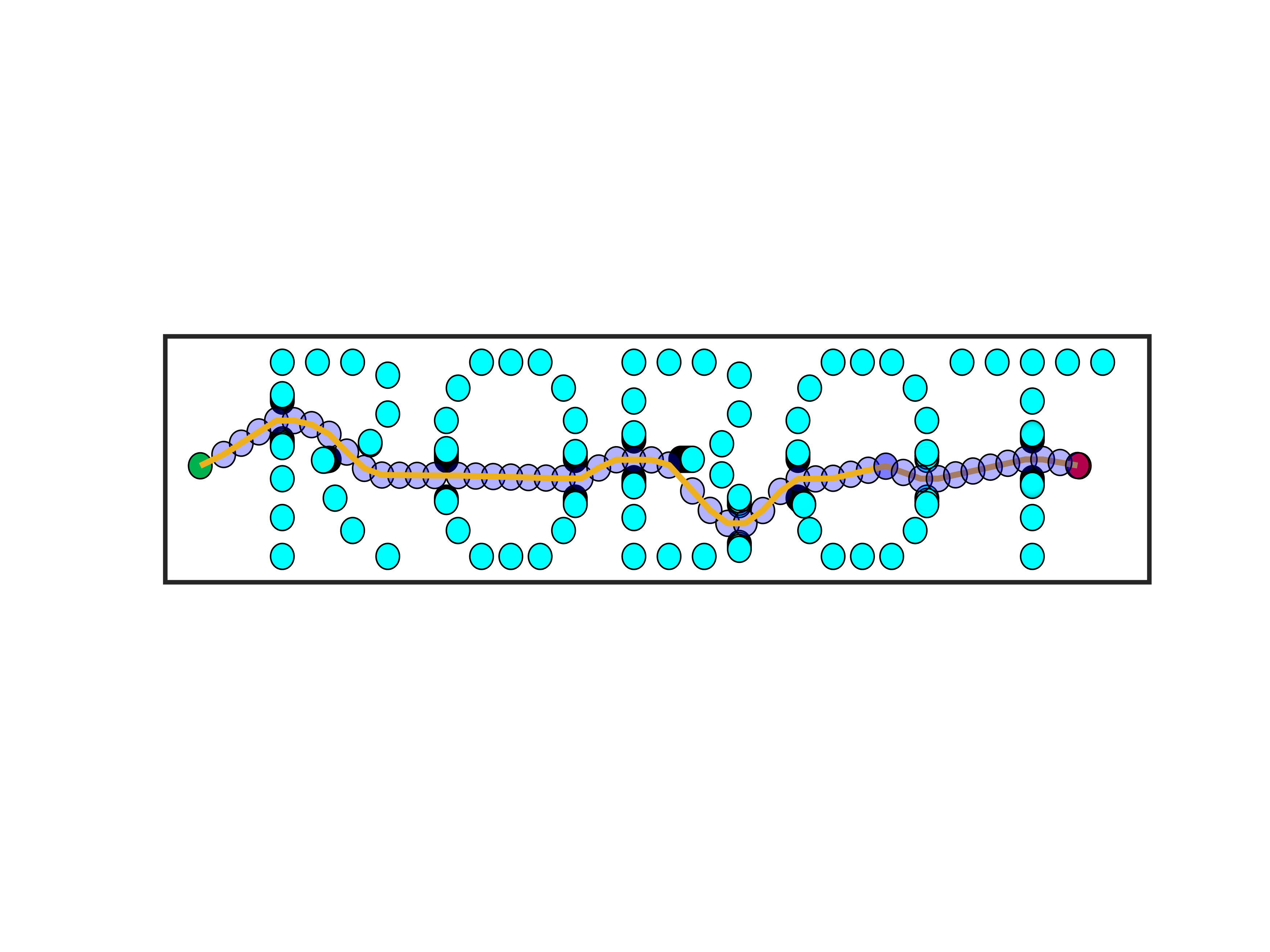}\label{fig:C74}}
\subfloat{\includegraphics[trim=475 330 360 220,clip,width=2.4cm,height=1.85cm]{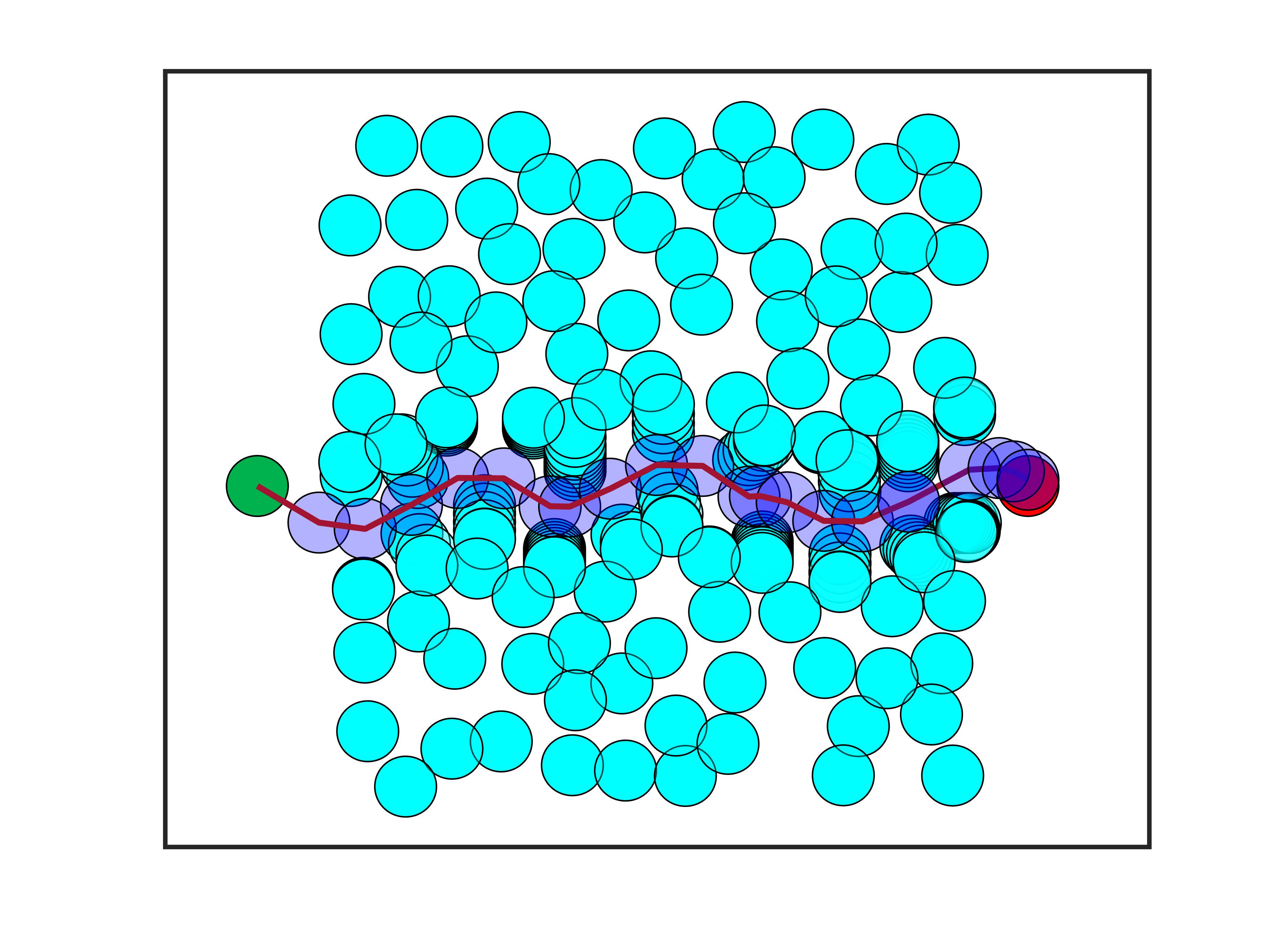}\label{fig:C84}}\\
    
\clearsubcaptcounter
\subfloat[\textsf{ROBOT}]{\includegraphics[trim=61 125 50 115,clip,width=6.8cm,height=1.8cm]{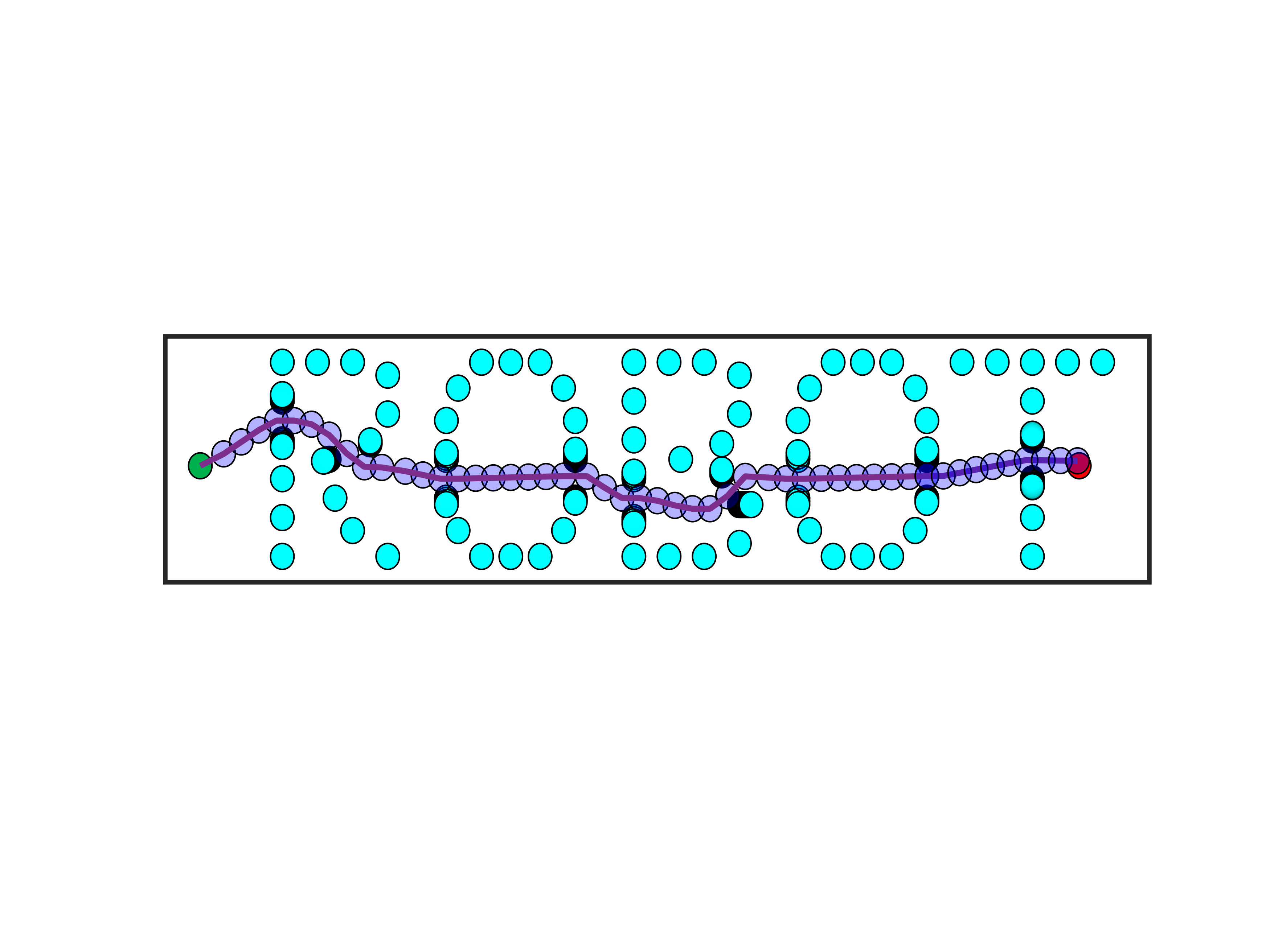}\label{fig:C75}}
\subfloat[\textsf{RANDOM}]{\includegraphics[trim=475 330 360 220,clip,width=2.4cm,height=1.85cm]{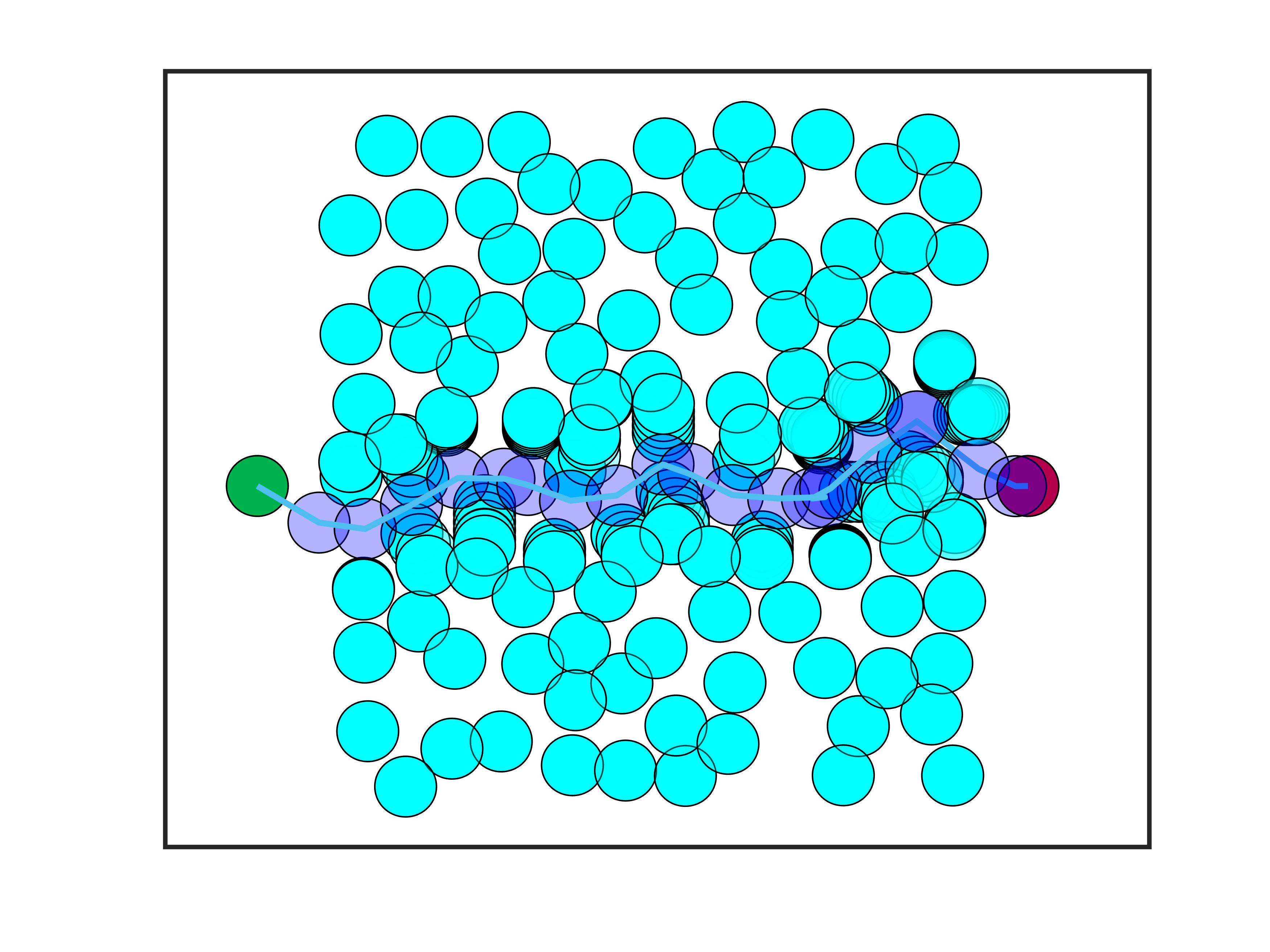}\label{fig:C85}}
\caption{
Robot trajectories and obstacle displacements for \textsf{ROBOT} and \textsf{RANDOM} while the robot moves according to the model in~\eqref{eq:robot_model2}. 
The first row corresponds to $m=1$ (no horizon slicing) for all four domains, the second row corresponds to $m=2$, and so on.
}
\label{fig:linearmodel2b}
\end{figure*}

\subsection{Linear Robot Model}

We first test \textsf{HS4MOD} adopting the robot dynamics in~\eqref{eq:robot_model1}. 
The robot trajectory and the computed obstacle displacements are visualized in Fig.~\ref{fig:linearmodel1a} and in Fig.~\ref{fig:linearmodel1b}.
Obstacles that have been displaced by the robot passage are characterized by a shaded trail of previous locations, whereas the cyan color represents their final location after the displacement.  
Various statistics for the same values are presented in Table~\ref{table:result1} under the robot model~\eqref{eq:robot_model1} column. 
The exact optimization case is denoted by $m=1$, and shows substantial computational load, as seen from Table~\ref{table:result1} --- the first line of each domain. 
As it has been argued before, the computational complexity is exponential in the number of binary variables, which is proportional to the number of movable obstacles, and it is particularly evident from the computational time in the \textsf{ROBOT} and \textsf{RANDOM} domains, with $74$ and $100$ obstacles, respectively.
The complexity depends also on 
(1) the spread of the obstacles with respect to the start and goal locations, and 
(2) the size of the environment. 
The first factor accounts for the difference in computation time for the cases $m=1$ and $m=2$ in the \textit{SQUARE\_1} and \textit{SQUARE\_2} domains, despite having the same number of binary variables. 
The two factors yield a higher computational time for the \textsf{ROBOT} domain than the \textsf{RANDOM} domain. 
Analogous results can be seen when employing the linear robot model in~\eqref{eq:robot_model2} --- the robot trajectory and obstacle movements visualized in Fig.~\ref{fig:linearmodel2a} and Fig.~\ref{fig:linearmodel2b}. 
Clearly, in all domains, for both robot models, as $m$ increases, the deviation from the optimal solution increases as well thereby resulting in sub-optimal solutions as corroborated by the rise in $\epsilon$ values. 
At the same time, higher values of $m$ result in great computational efficiency. 
Keeping in mind a pragmatic trade-off between optimality and computational efficiency, the statistics in Table~\ref{table:result1} suggest that for a small number of movable obstacles, for example in the domains \textsf{SQUARE\_1} and $\textsf{SQUARE\_2}$, $m=2$ may be used. For domains with a larger number of obstacles (\textsf{ROBOT} and \textsf{RANDOM}), $m=3$ may be employed.

\begin{figure}[]
\subfloat[]{\includegraphics[trim=70 125 50 110,clip,scale=0.7]{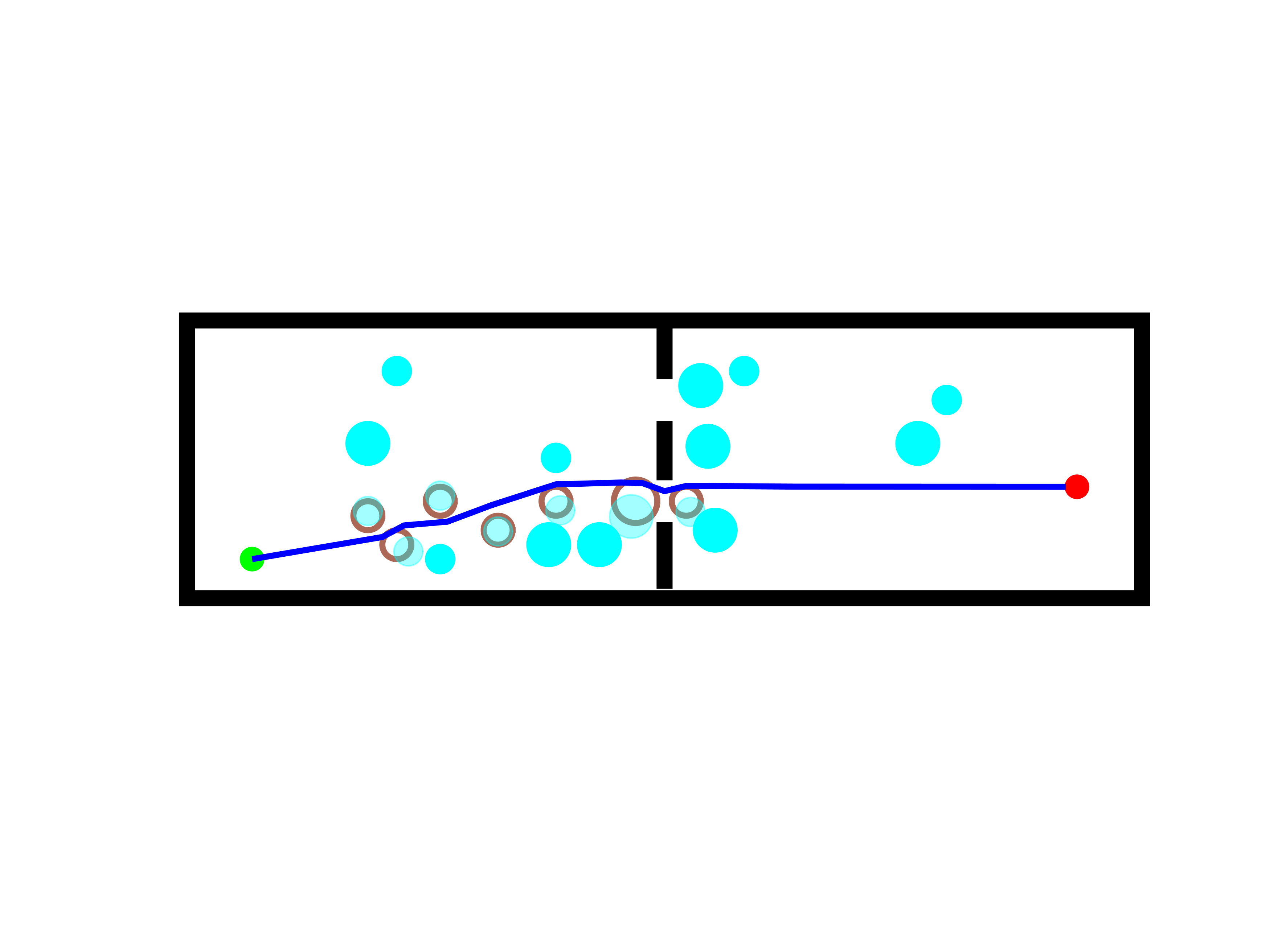}\label{fig:A2}}\vspace{10pt}\\
\subfloat[]{\includegraphics[trim=70 125 50 110,clip,scale=0.7]{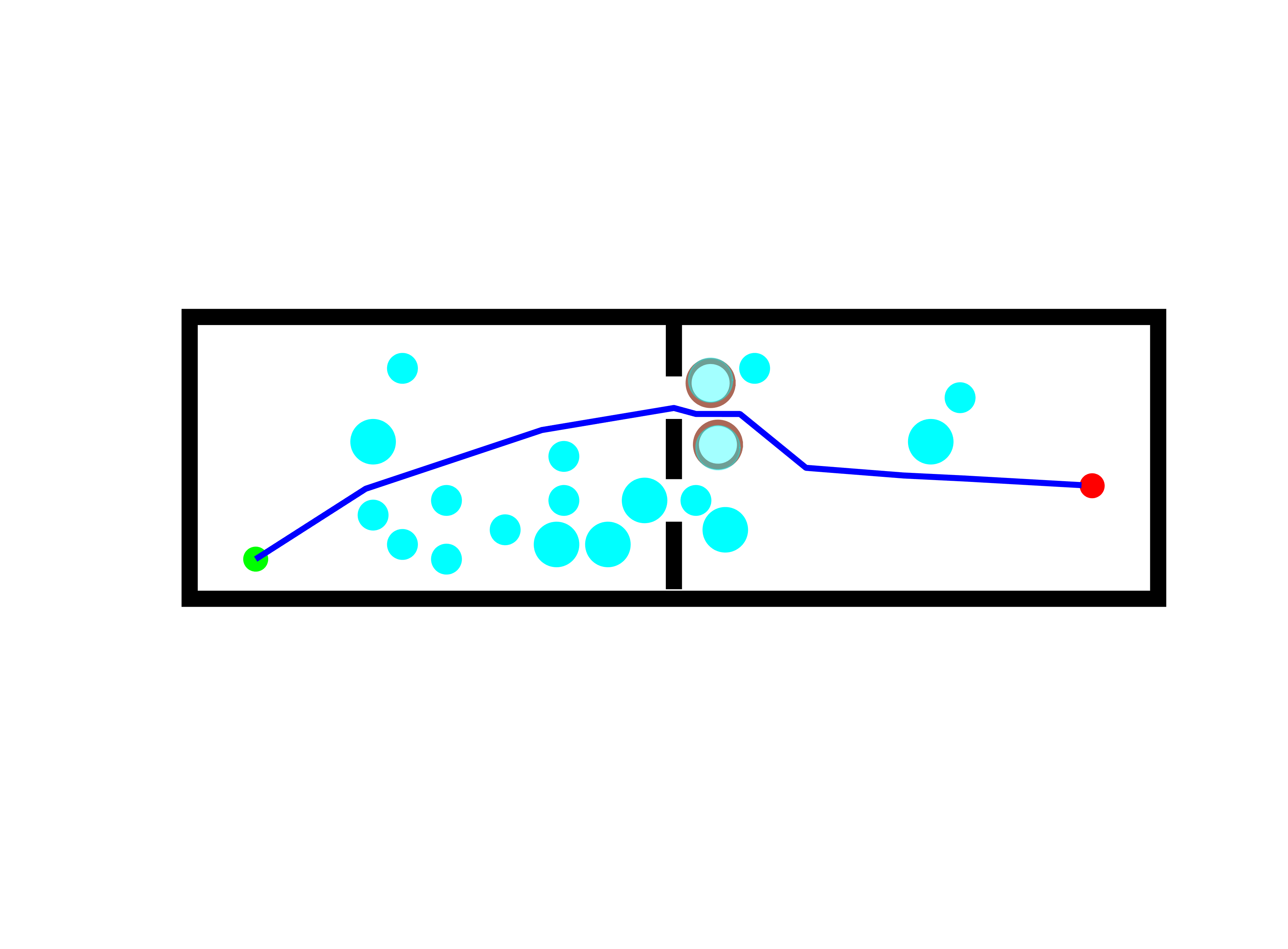}\label{fig:A3}}
\caption{
Two trajectories related to the example scenario that was introduced at the beginning of the paper.
A humanoid Pepper robot must navigate to the right-hand side of the map, and to that aim it must displace movable obstacles to pass through one of the two available passages.
(a) Pepper finds a short path that implies displacing $7$ obstacles.
(b) Increasing the obstacle displacement factor ($J^o$) weight from 10 to 1000, the robot finds a path with lower obstacle displacement but with a higher path length.}
\label{fig:application}
\end{figure}

Let us now consider again a simulated Pepper humanoid robot having to move through a cluttered environment as seen in Fig.~\ref{fig:example_scenario}. 
Pepper is asked to move to the other room, but its path is blocked by the specific configuration of the tables. 
Employing the robot model~\eqref{eq:robot_model1} with $m=2$ and $w^g = 1$, $w^x = 1$, $w^d = 10$, a 7-obstacle solution with a displacement magnitude of $4.4$ meters is obtained (see Fig.~\ref{fig:A2}). 
Penalizing large displacements by increasing $\alpha^o$ from 10 to 1000, a longer path with a 2-obstacle solution (see Fig.~\ref{fig:A3}) can be obtained with a lower displacement magnitude of $0.4$ meters. 
We recall here that the cost for the last time step is the deviation of the robot location from its goal.

\noindent \textbf{Baseline comparison}. We additionally compare \textsf{HS4MOD} against a NAMO baseline on the \textsf{SQUARE\_2} and \textsf{ROBOT} environments. Similar to~\cite{stilman2005IJHR, thomas2022IAS}, the baseline first constructs the robot's configuration space and computes a path to the goal using an $A^\star$ search that minimizes the number of obstacles intersecting the path. The obstacles intersecting the path are then displaced to obtain a collision-free trajectory, following the approach proposed in~\cite{thomas2026RAS}.

Table~\ref{tab:namo_compare} compares the robot path length, total obstacle displacement, and planning time \textsf{HS4MOD} and the NAMO baseline. In the \textsf{SQUARE\_2} environment, \textsf{HS4MOD} produces a shorter robot path with nearly identical obstacle displacement, while reducing planning time by more than a factor of three. In the \textsf{ROBOT} environment, both methods generate paths of comparable length, but \textsf{HS4MOD} requires less obstacle displacement and achieves a lower planning time. These results demonstrate that \textsf{HS4MOD}, even when $m=5$ offers a favorable trade-off between solution quality and runtime.
\begin{table}[t]
\centering
\caption{Comparison of robot path length, obstacle displacement and planning time for \textsf{HS4MOD} ($m=5$) and NAMO baseline.}
\label{tab:namo_compare}
\scalebox{0.7}{
\begin{tabular}{lccc|ccc}
\hline
& \multicolumn{3}{c|}{\textit{SQUARE\_2}} & \multicolumn{3}{c}{\textit{ROBOT}} \\
\cline{2-7}
Method & Length (m) & Displacement (m) & Time (s) &Length (m) & Displacement (m) & Time (s)\\
\hline
\textsf{HS4MOD} ($m=5$)  & 36.92  & 7.25 & 1.36 & 78.55 & 9.17 & 12.23  \\
NAMO Baseline &  41.00  &  7.26 & 4.57 &  78.20 & 10.80 & 17.82 \\
\hline
\end{tabular}}
\end{table}

\subsection{Nonlinear Robot Model}

\begin{figure}[]
\subfloat{\vspace{-7pt}\includegraphics[trim=500 315 380 310,clip,width=3.6cm,height=2.4cm]{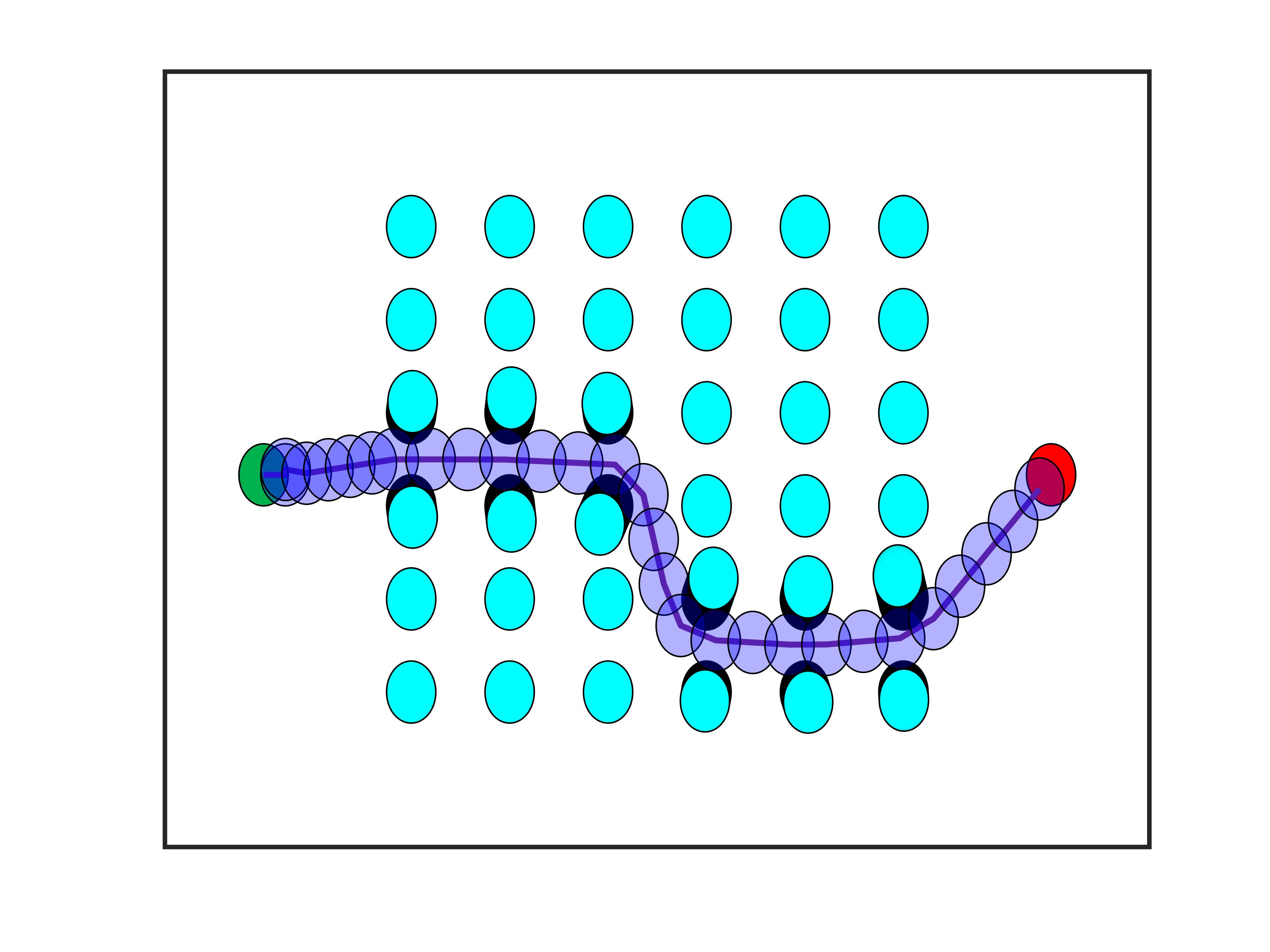}\label{fig:C91}}
\subfloat{\vspace{-7pt}\includegraphics[trim=100 51 58 51,clip,width=3.6cm,height=2.4cm]{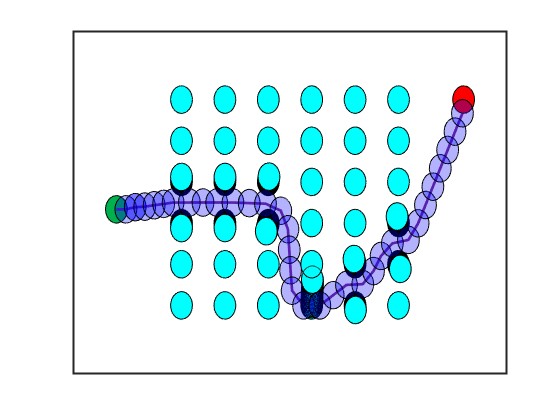}\label{fig:C101}}\\
    
\subfloat{\vspace{-7pt}\includegraphics[trim=500 315 380 310,clip,width=3.6cm,height=2.4cm]{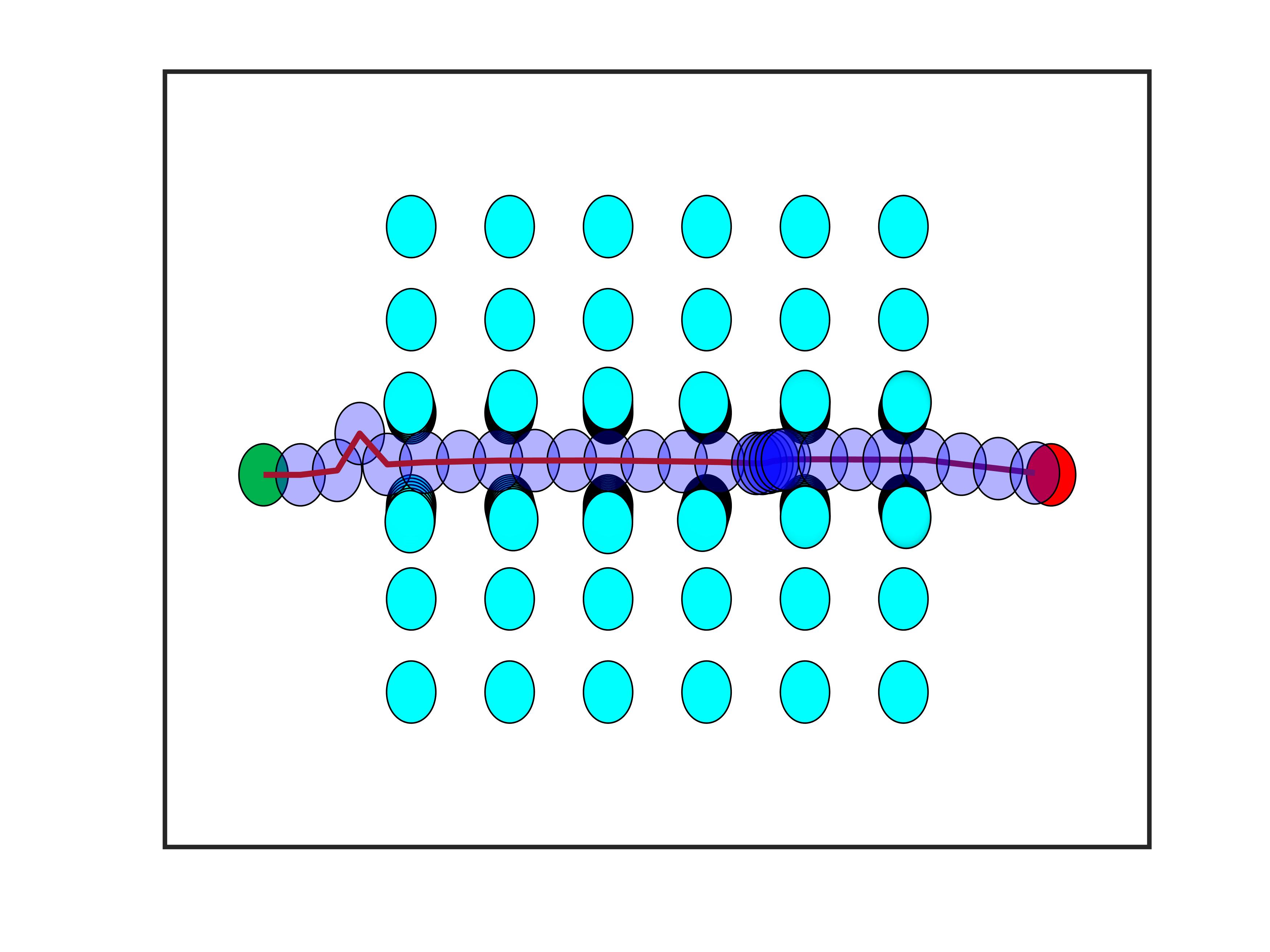}\label{fig:C92}}
\subfloat{\vspace{-7pt}\includegraphics[trim=100 51 58 51,clip,width=3.6cm,height=2.4cm]{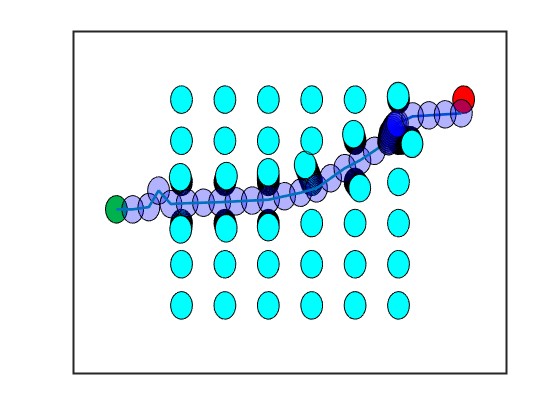}\label{fig:C102}}\\
     
\subfloat{\vspace{-7pt}\includegraphics[trim=500 315 380 310,clip,width=3.6cm,height=2.4cm]{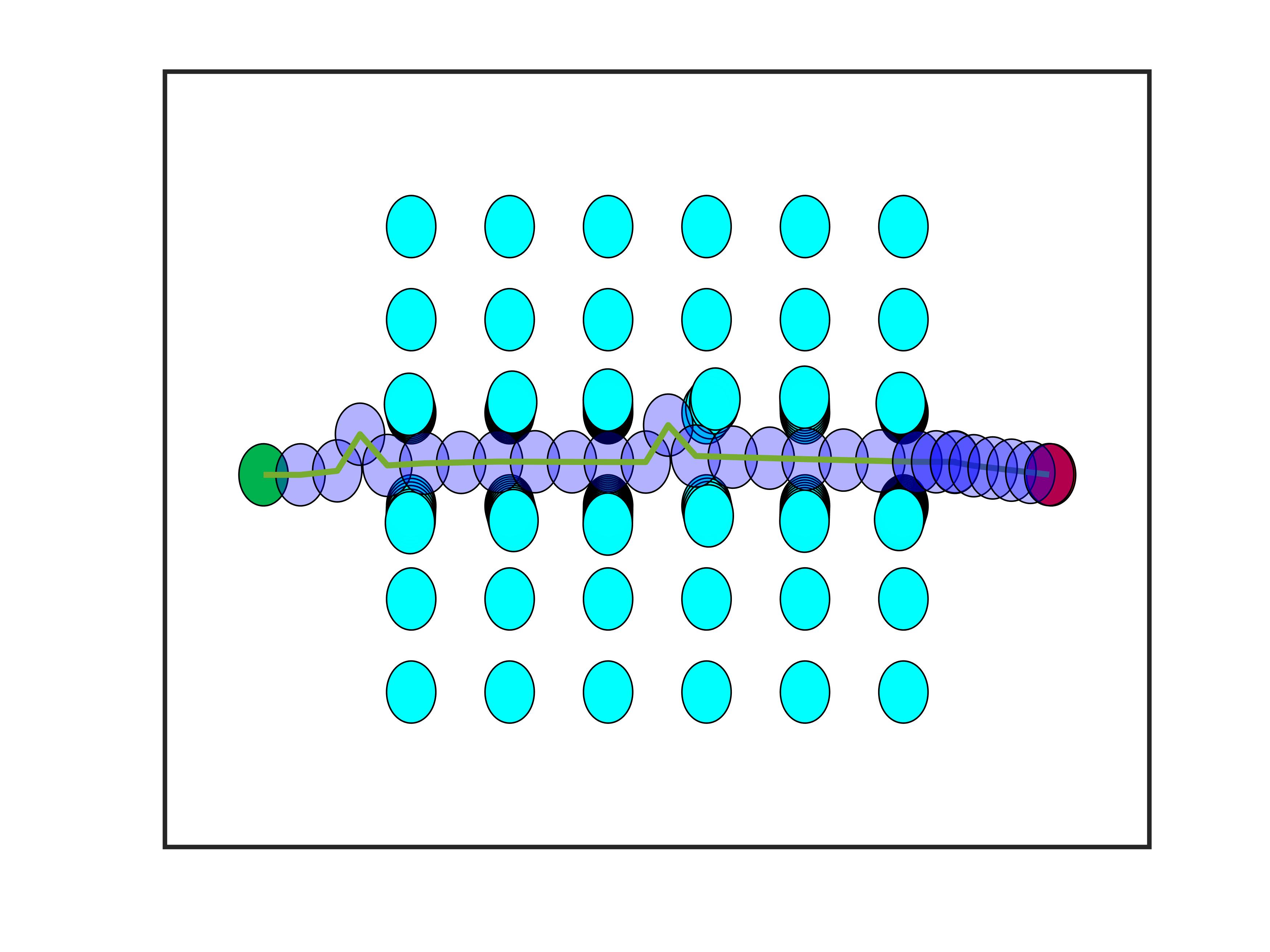}\label{fig:C93}}
\subfloat{\vspace{-7pt}\includegraphics[trim=100 51 58 51,clip,width=3.6cm,height=2.4cm]{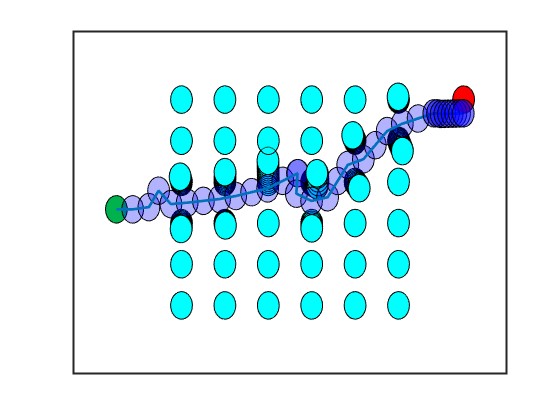}\label{fig:C103}}\\
     
\subfloat{\vspace{-7pt}\includegraphics[trim=500 315 380 310,clip,width=3.6cm,height=2.4cm]{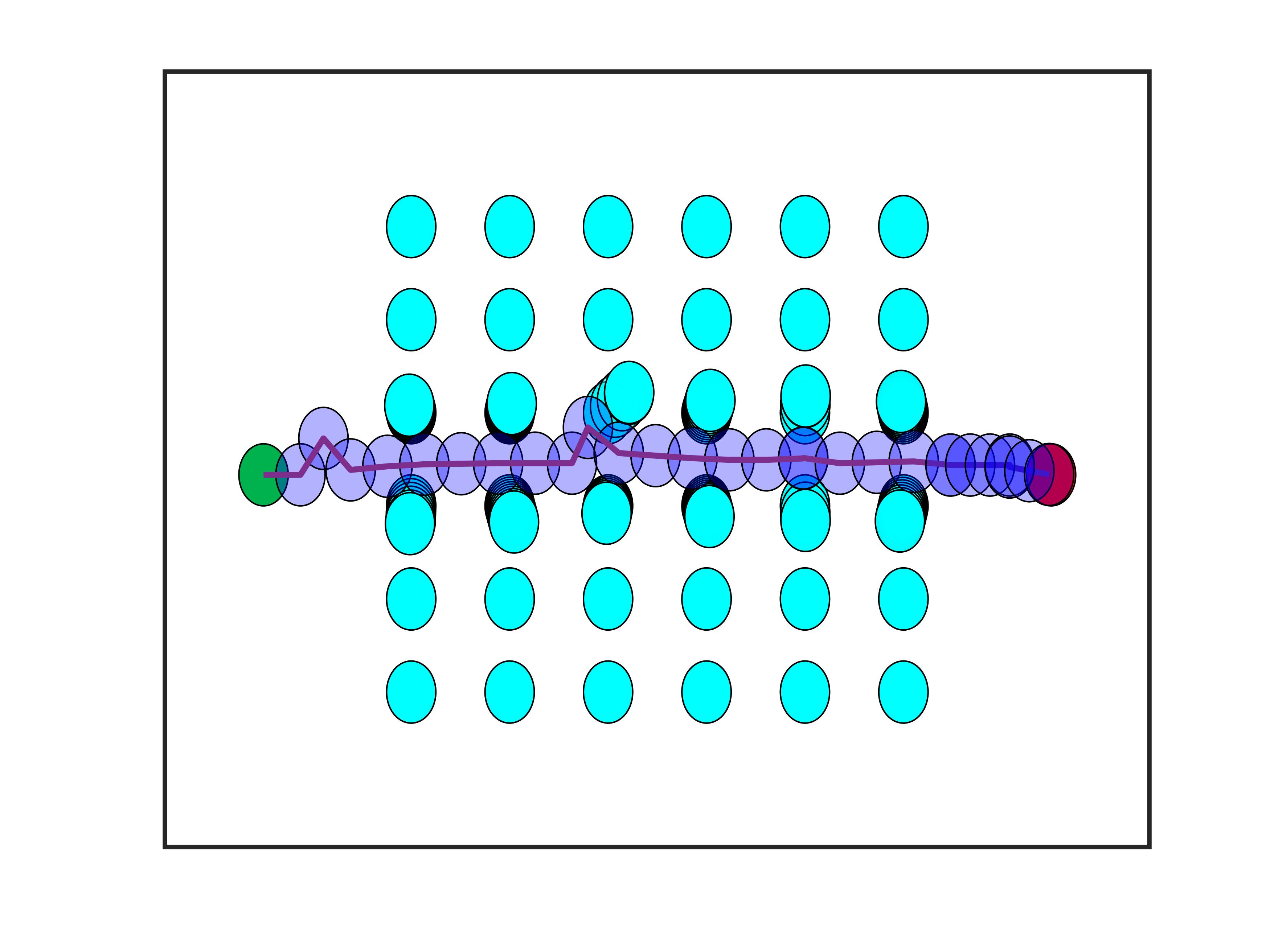}\label{fig:C94}}
\subfloat{\vspace{-7pt}\includegraphics[trim=100 51 58 51,clip,width=3.6cm,height=2.4cm]{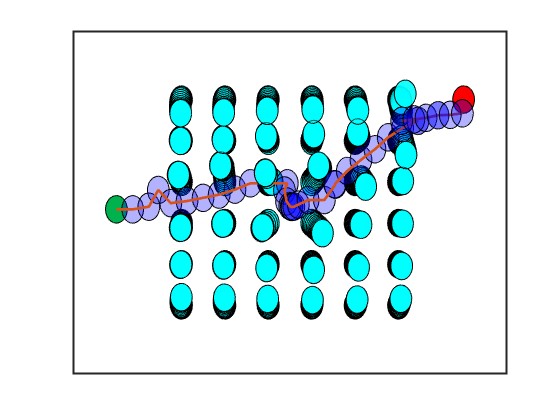}\label{fig:C104}}\\
    
\clearsubcaptcounter
\subfloat[\textsf{SQUARE\_1}]{\vspace{-7pt}\includegraphics[trim=500 315 380 310,clip,width=3.6cm,height=2.4cm]{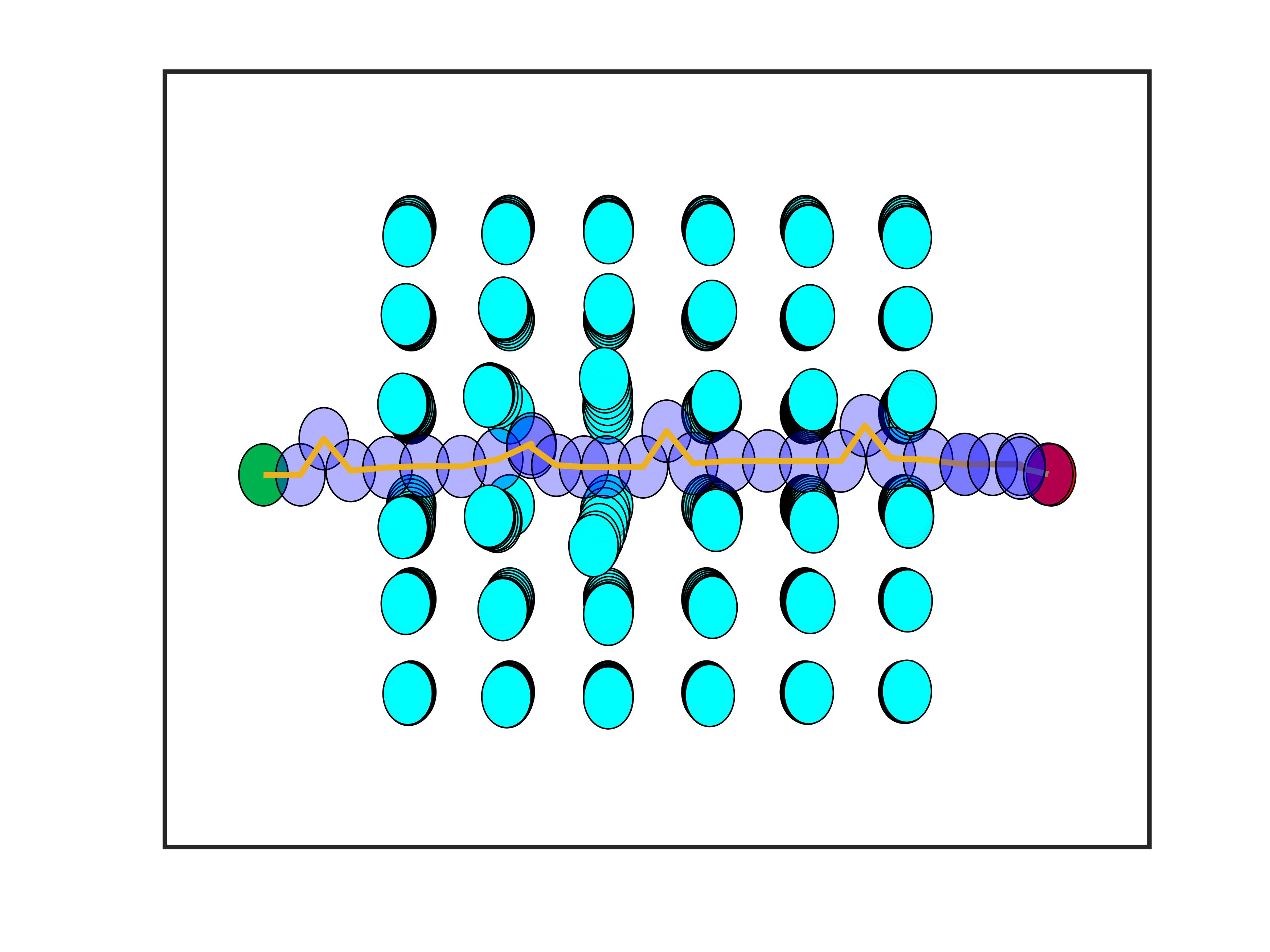}\label{fig:C95}}
\subfloat[\textsf{SQUARE\_2}]{\vspace{-7pt}\includegraphics[trim=100 51 58 51,clip,width=3.6cm,height=2.4cm]{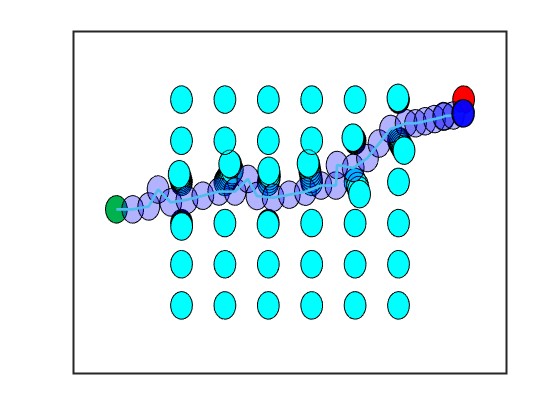}\label{fig:C105}}
\caption{
Robot trajectories and obstacle displacements for \textsf{SQUARE\_1} and \textsf{SQUARE\_2} domains while the robot moves according to the model in~\eqref{eq:robot_model3}. 
The first row corresponds to $m=1$ (no horizon slicing), the second row corresponds to $m=2$, and so on.}
\label{fig:nonlinearmodel1}
\end{figure}

\begin{figure}[]
\subfloat{\vspace{-10pt}\includegraphics[trim=100 51 74 51,clip,width=3.6cm,height=2.4cm]{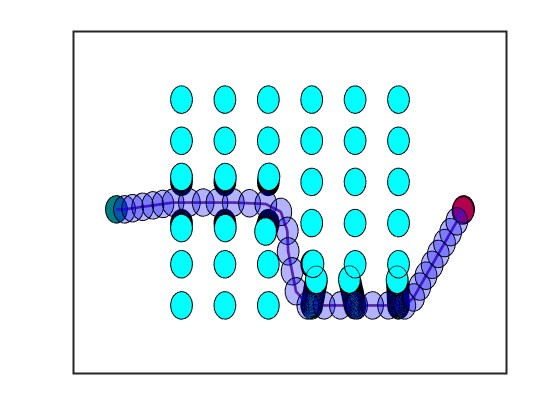}\label{fig:C111}}
\subfloat{\vspace{-10pt}\includegraphics[trim=100 51 74 51,clip,width=3.6cm,height=2.5cm]{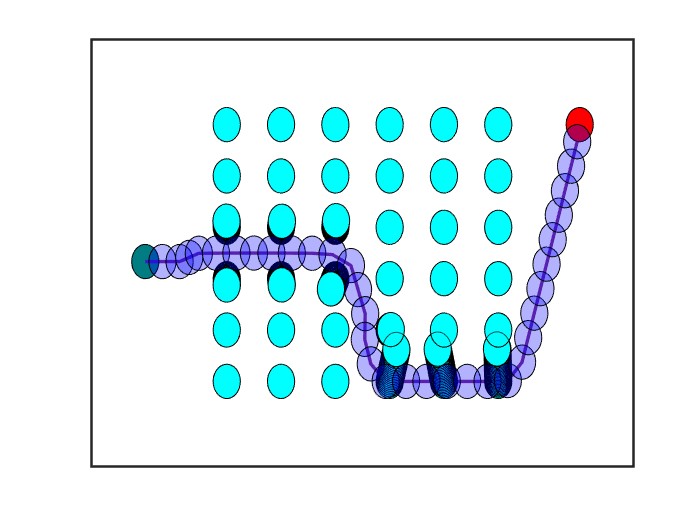}\label{fig:C121}}\\
    
\subfloat{\vspace{-10pt}\includegraphics[trim=100 51 74 51,clip,width=3.6cm,height=2.4cm]{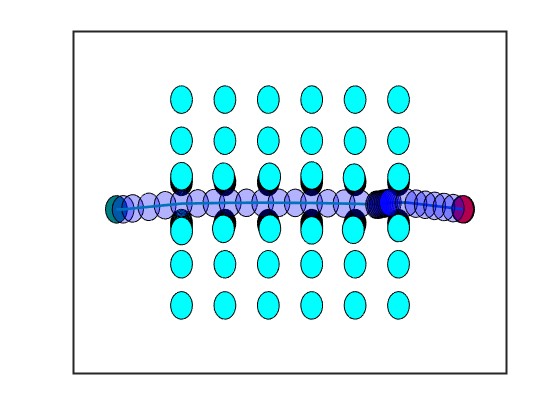}\label{fig:C112}}
\subfloat{\vspace{-10pt}\includegraphics[trim=100 51 74 51,clip,width=3.6cm,height=2.5cm]{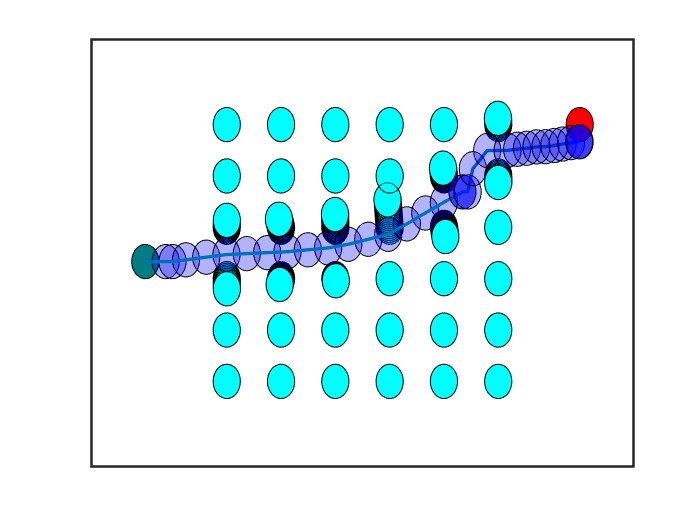}\label{fig:C122}}\\
     
\subfloat{\vspace{-10pt}\includegraphics[trim=100 51 74 51,clip,width=3.6cm,height=2.4cm]{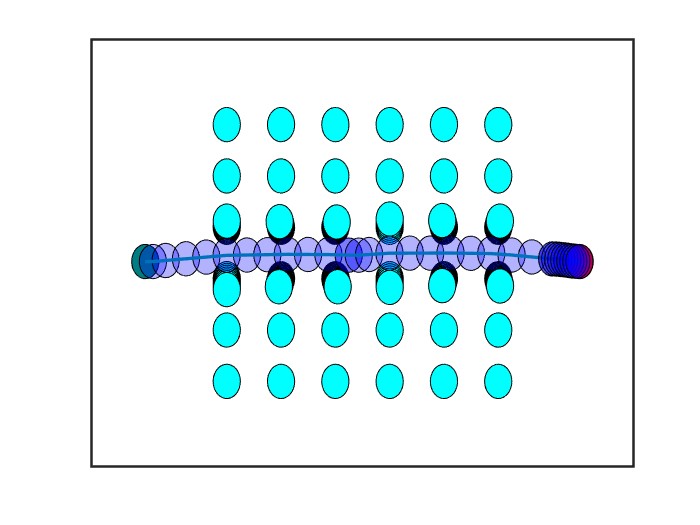}\label{fig:C113}}
\subfloat{\vspace{-10pt}\includegraphics[trim=100 51 74 51,clip,width=3.6cm,height=2.4cm]{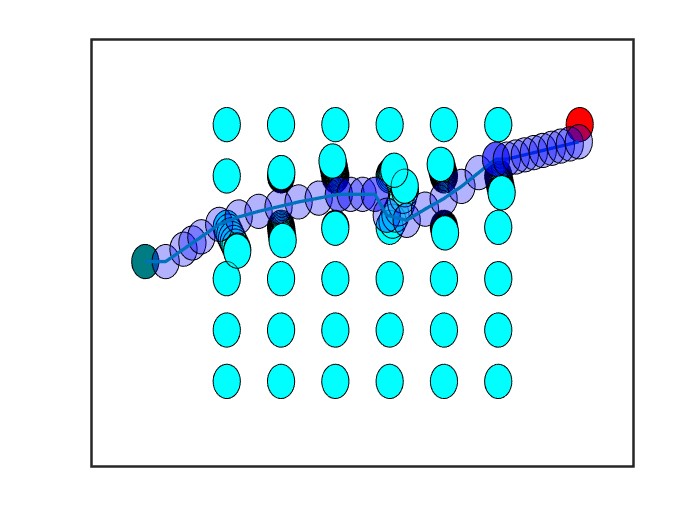}\label{fig:C123}}\\
     
\subfloat{\vspace{-10pt}\includegraphics[trim=100 51 74 51,clip,width=3.6cm,height=2.4cm]{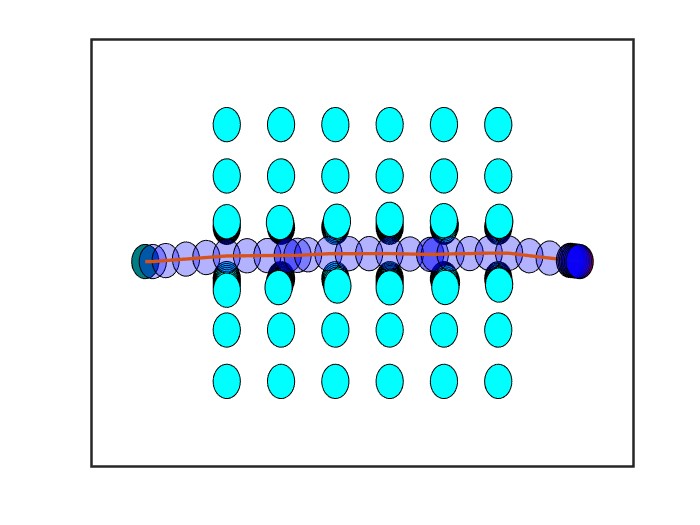}\label{fig:C114}}
\subfloat{\vspace{-10pt}\includegraphics[trim=100 51 74 51,clip,width=3.6cm,height=2.4cm]{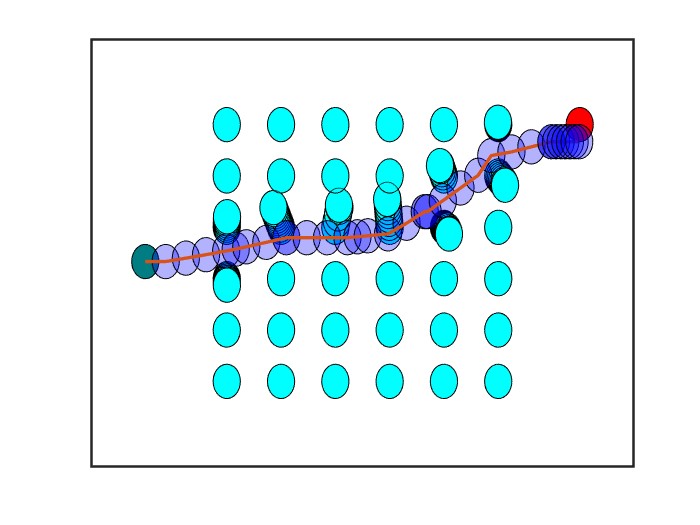}\label{fig:C124}}\\
    
\clearsubcaptcounter
\subfloat[\textsf{SQUARE\_1}]{\vspace{-10pt}\includegraphics[trim=100 71 74 51,clip,width=3.6cm,height=2.4cm]{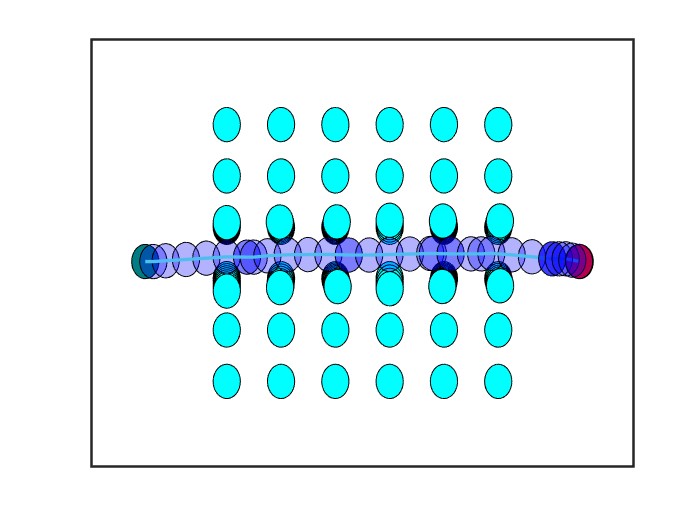}\label{fig:C115}}
\subfloat[\textsf{SQUARE\_2}]{\vspace{-10pt}\includegraphics[trim=100 71 74 51,clip,width=3.6cm,height=2.4cm]{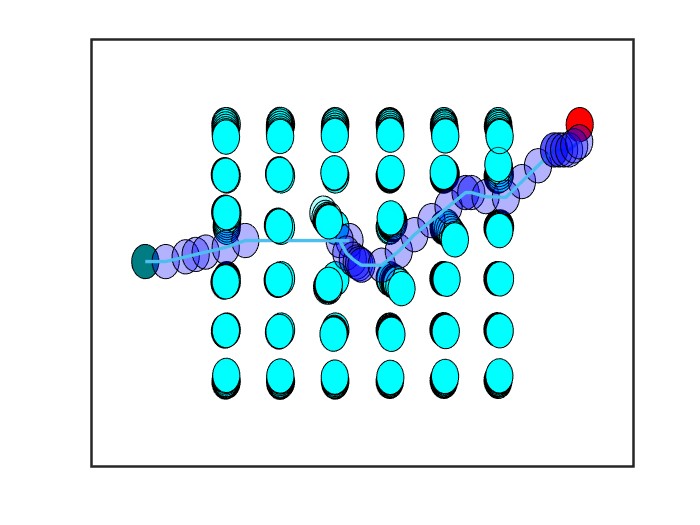}\label{fig:C125}}
\caption{
Robot trajectories and obstacle displacements are depicted for the \textsf{SQUARE\_1} and \textsf{SQUARE\_2} domains, where the robot moves according to the model in~\eqref{eq:robot_model4}. 
Each row corresponds to a different value of $m$: the first row corresponds to $m=1$ (no horizon slicing), the second row corresponds to $m=2$, and so on.}
\label{fig:nonlinearmodel2}
\end{figure}

As argued in Section~\ref{sec:approach}, when a nonlinear robot model is adopted for the optimization problem~\eqref{eq:optimization_problem}, optimality cannot be guaranteed. 
We test \textsf{HS4MOD} with the nonlinear models~\eqref{eq:robot_model3} and~\eqref{eq:robot_model4} on \textsf{SQUARE\_1} and \textsf{SQUARE\_2}.
The corresponding robot trajectories and the obstacle displacements with varying $m$ for both the robot models are visualized in Fig.~\ref{fig:nonlinearmodel1} and Fig.~\ref{fig:nonlinearmodel2}, respectively. 
The results indicate that regardless of the robot model, the trajectories exhibit a similar pattern. 
However, it is observed that certain obstacles are displaced unnecessarily, such as, when $m=5$ (last row) in Fig.~\ref{fig:nonlinearmodel1}(a), $m=4$ in Fig.~\ref{fig:nonlinearmodel1}(b), and $m=5$ in Fig.~\ref{fig:nonlinearmodel2}(b). 
This phenomenon can be attributed to the non-convexity of the optimization problem, resulting in solutions that are local optima.
Numerical results are presented in Table~\ref{table:result2}. 
As argued in Section~\ref{sec:approach}, due to the nonlinearity of the models, $m=1$ does not guarantee an optimal solution. 
In contrast to the findings in Table~\ref{table:result1}, in Table~\ref{table:result2}, the change in $\epsilon$ is not monotonic with an increase in $m$ (note that $\epsilon$ is computed with respect to $m=1$). This behavior results from the interaction between horizon discretization and nonlinear dynamics, whereby increasing the number of slices can alter the feasible trajectory and the associated obstacle displacements without necessarily yielding a progressively improved solution. Therefore, no particular value of $m$ can be identified as providing a consistent trade-off between solution quality and computational efficiency for the nonlinear models. Moreover, $\epsilon$ should be interpreted as a relative measure with respect to the $m=1$ solution, rather than as a direct measure of the true optimality gap.

Therefore, using a nonlinear model in~\eqref{eq:cost_fn1} precludes any validation on  $\epsilon$. 
Nonetheless, in such scenarios, a linear model can initially be employed to determine the necessary path and displacements using \textsf{HS4MOD}.
As a consequence, a trajectory-tracking algorithm can be employed to calculate the requisite controls for the nonlinear model.

\setlength{\tabcolsep}{4pt}
\begin{table*}[]
\resizebox{\columnwidth}{!}{%
\begin{tabular}{l c rccc | rccc }        
\multicolumn{1}{c}{\multirow{ 1}{*}{}} & \multicolumn{1}{c}{} & \multicolumn{4}{c}{Robot model~\eqref{eq:robot_model1}} & \multicolumn{4}{c}{Robot model~\eqref{eq:robot_model2}}\\ [0.5ex]
\cline{3-6}\cline{7-10} \\
{} & {m} & \multicolumn{1}{c}{CPU [$s$]} & \multicolumn{1}{c}{Gap/$\epsilon$} & \multicolumn{1}{c}{$J^r [m^2]$} & \multicolumn{1}{c}{$J^o [m^2]$} & \multicolumn{1}{c}{CPU [$s$]} & \multicolumn{1}{c}{Gap/$\epsilon$} & \multicolumn{1}{c}{$J^r [m^2]$} & \multicolumn{1}{c}{$J^o [m^2]$} \\ [0.5ex]
\hline
\hline
                            & 1 & 1011.41   & 0.00  & 45.12     & 0.27  & 1001.34   & 0.00  & 47.35     & 0.35 \\ 
                            & 2 & 516.44    & 0.59  & 48.02     & 0.55  & 11.85     & 0.45  & 48.67     & 0.61 \\ 
\textsf{SQUARE\_1} & 3 & 3.58      & 0.89  & 44.64     & 0.72  & 3.06      & 1.28  & 48.49     & 1.09 \\
                            & 4 & 2.99      & 1.97  & 45.45     & 1.25  & 2.58      & 1.50  & 49.09     & 1.22 \\ 
                            & 5 & 1.50      & 2.76  & 46.21     & 1.64  & 1.37      & 2.83  & 51.97     & 1.99 \\
\hline 	
\hline    
                            & 1 & 1601.58   & 0.00  & 51.20     & 0.28  & 1000.98   & 0.00  & 51.22     & 0.28 \\
                            & 2 & 27.92     & 0.62  & 59.17     & 0.56  & 116.37    & 0.76  & 59.33     & 0.65 \\
\textsf{SQUARE\_2} & 3 & 4.79      & 0.90  & 53.87     & 0.74  & 7.19      & 1.06  & 54.38     & 0.84 \\ 
                            & 4 & 1.32      & 2.10  & 59.44     & 1.35  & 1.95      & 3.03  & 64.91     & 1.85 \\
                            & 5 & 1.02      & 3.08  & 59.29     & 1.88  & 1.36      & 3.72  & 65.90     & 2.22 \\
\hline 
\hline    
                            & 1 & 28814.74  & 0.00  & 122.99    & 0.28  & 28808.67  & 0.00  & 132.70    & 0.32 \\ 
                            & 2 & 17367.71  & 0.25  & 121.64    & 0.52  & 19564.10  & 0.18  & 133.00    & 0.49 \\ 
\textsf{ROBOT}	    & 3 & 128.68    & 0.37  & 121.08    & 0.63  & 147.85    & 0.31  & 127.93    & 0.66 \\
                            & 4 & 24.23     & 0.98  & 119.32    & 1.19  & 28.83     & 0.71  & 136.78    & 0.99 \\ 
                            & 5 & 11.62     & 1.21  & 142.74    & 1.28  & 12.24     & 0.75  & 128.16    & 1.08 \\              
\hline 
\hline    
                            & 1 & 16009.18  & 0.00  & 51.86     & 1.13  & 18010.40  & 0.00  & 39.15     & 1.60 \\
                            & 2 & 12813.12  & 0.78  & 49.56     & 2.24  & 14413.58  & 0.77  & 52.28     & 2.93 \\ 
\textsf{RANDOM}    & 3 & 6403.36   & 0.99  & 46.68     & 2.55  & 771.53    & 1.35  & 48.39     & 3.99 \\
                            & 4 & 897.14    & 1.83  & 48.85     & 3.71  & 44.76     & 1.43  & 43.72     & 4.16 \\ 
                            & 5 & 14.43     & 2.05  & 46.70     & 4.03  & 20.47     & 2.16  & 47.55     & 5.44 \\  
\hline
\end{tabular}   }                               
\caption{
Different statistics for the four domains considered in this paper, while employing the linear robot models in~\eqref{eq:robot_model1} and~\eqref{eq:robot_model2}. 
$m=1$ corresponds to the scenario with no horizon-slicing.
CPU denotes the computation time in seconds for the optimization problem in~\eqref{eq:optimization_problem}, $J^r = c(\B{p})$, and $J^o = \sum_{i=1}^{\mathcal{M}} \sum_{k=0}^T d^i_k$.} 
\label{table:result1}
\end{table*}

\setlength{\tabcolsep}{4pt}
\begin{table*}[]
\resizebox{\columnwidth}{!}{%
 \begin{tabular}{l c rccc | rccc }        
\multicolumn{1}{c}{\multirow{ 1}{*}{}} & \multicolumn{1}{c}{} & \multicolumn{4}{c}{Robot model~\eqref{eq:robot_model3}} & \multicolumn{4}{c}{Robot model~\eqref{eq:robot_model4}}\\ [0.5ex]
\cline{3-6}\cline{7-10} \\ 
{} & {m} & \multicolumn{1}{c}{CPU [$s$]} & \multicolumn{1}{c}{Gap/$\epsilon$} & \multicolumn{1}{c}{$J^r [m^2]$} & \multicolumn{1}{c}{$J^o [m^2]$} & \multicolumn{1}{c}{CPU [$s$]} & \multicolumn{1}{c}{Gap/$\epsilon$} & \multicolumn{1}{c}{$J^r [m^2]$} & \multicolumn{1}{c}{$J^o [m^2]$} \\ [0.5ex]
\hline
\hline

                            & 1 & 88.59   & 0.00  & 52.19     & 0.20 & 307.17   & 0.00  & 52.31    & 0.61 \\ 
                            & 2 & 96.99    & 0.09  & 46.73     & 0.27  & 150.38     & 0.92  & 41.86     & 0.24 \\ 
\textsf{SQUARE\_1} & 3 & 61.81   & 0.43  & 46.89     & 0.46  & 128.93      & 0.46  & 43.27     & 0.38 \\
                            & 4 & 26.53    & 1.35  & 47.75     & 0.89  & 97.34     & 0.37  & 43.83     & 0.41 \\ 
                            & 5 & 12.22    & 4.69  & 50.00     & 2.49  & 35.24     & 0.12  & 42.28     & 0.57 \\
\hline 	
\hline    
                            & 1 & 141.78   & 0.00  & 62.43     & 0.30 & 71.53   & 0.00  & 67.75    & 0.65 \\
                            & 2 & 47.33     & 0.89  & 49.75     & 0.45  & 29.18    & 0.44  & 45.15     & 0.45 \\
\textsf{SQUARE\_2} & 3 &  125.09      & 0.22  & 52.67     & 0.85  & 146.78      & 0.49  & 43.76     & 1.81 \\ 
                            & 4 & 86.29      & 0.61 & 60.75     & 3.28 & 68.66     & 0.37  & 45.16     & 1.42 \\
                            & 5 & 60.09      & 0.18  & 51.55    & 1.40  & 74.10      & 0.72  & 95.50     & 3.15 \\

\hline
\end{tabular}  }                                
\caption{
Average computation time, Gap- wrt to $m=1$, path $ \left( J^r = c(\B{p}) \right)$ and displacement function $\left(J^o = \sum_{i=1}^{\mathcal{M}} \sum_{k=0}^T d^i_k\right)$ values while employing the nonlinear robot models in~\eqref{eq:robot_model3} and~\eqref{eq:robot_model4}.} 
\label{table:result2}
\end{table*}

\begin{figure*}[]
\subfloat[]{\includegraphics[trim=70 125 50 110,clip,scale=0.7]{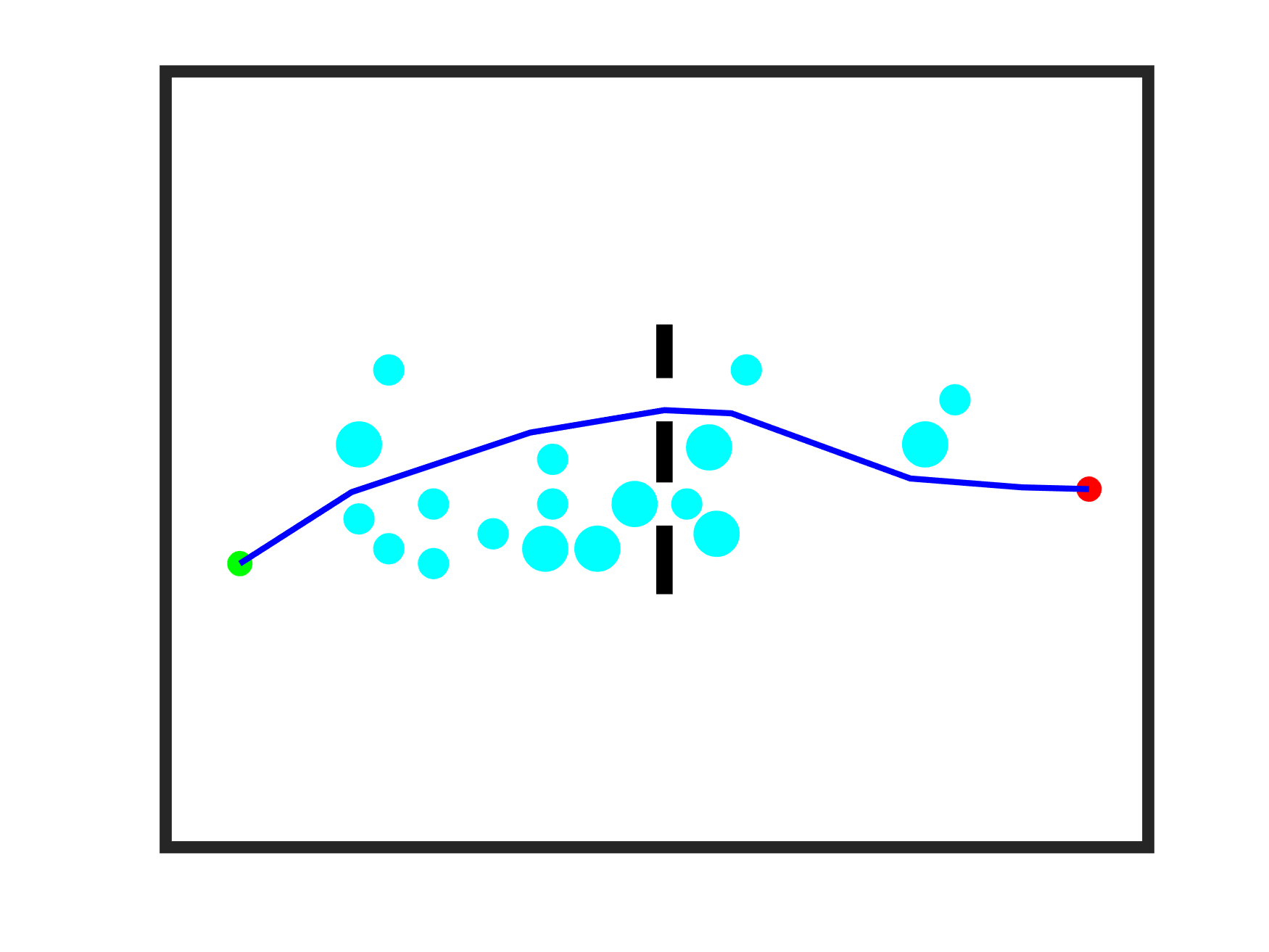}\label{fig:diff1}} \\ \vspace{10pt}
\subfloat[]{\includegraphics[trim=70 125 50 110,clip,scale=0.7]{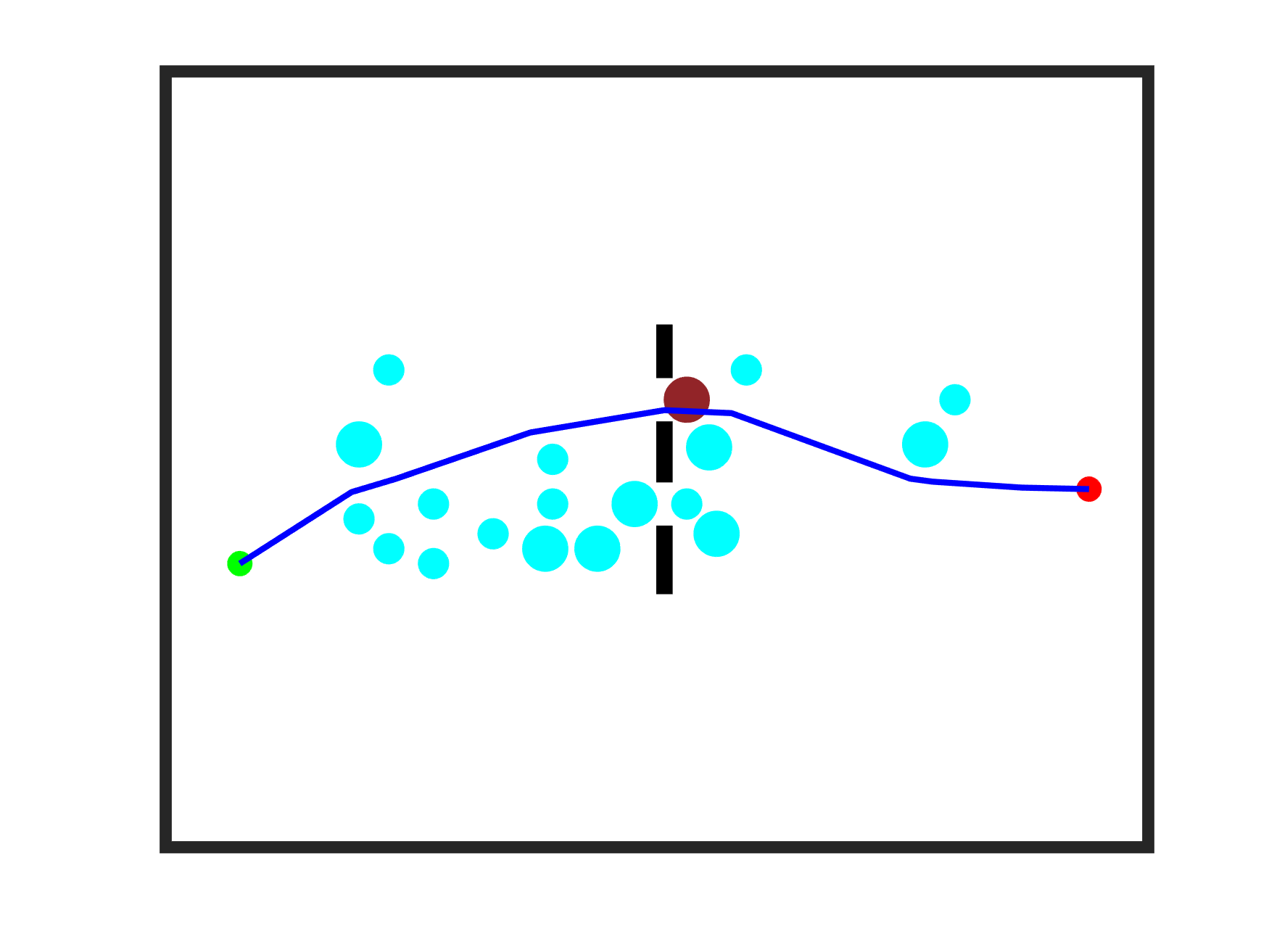}\label{fig:diff2}} \\ \vspace{10pt}
\subfloat[]{\includegraphics[trim=70 125 50 110,clip,scale=0.7]{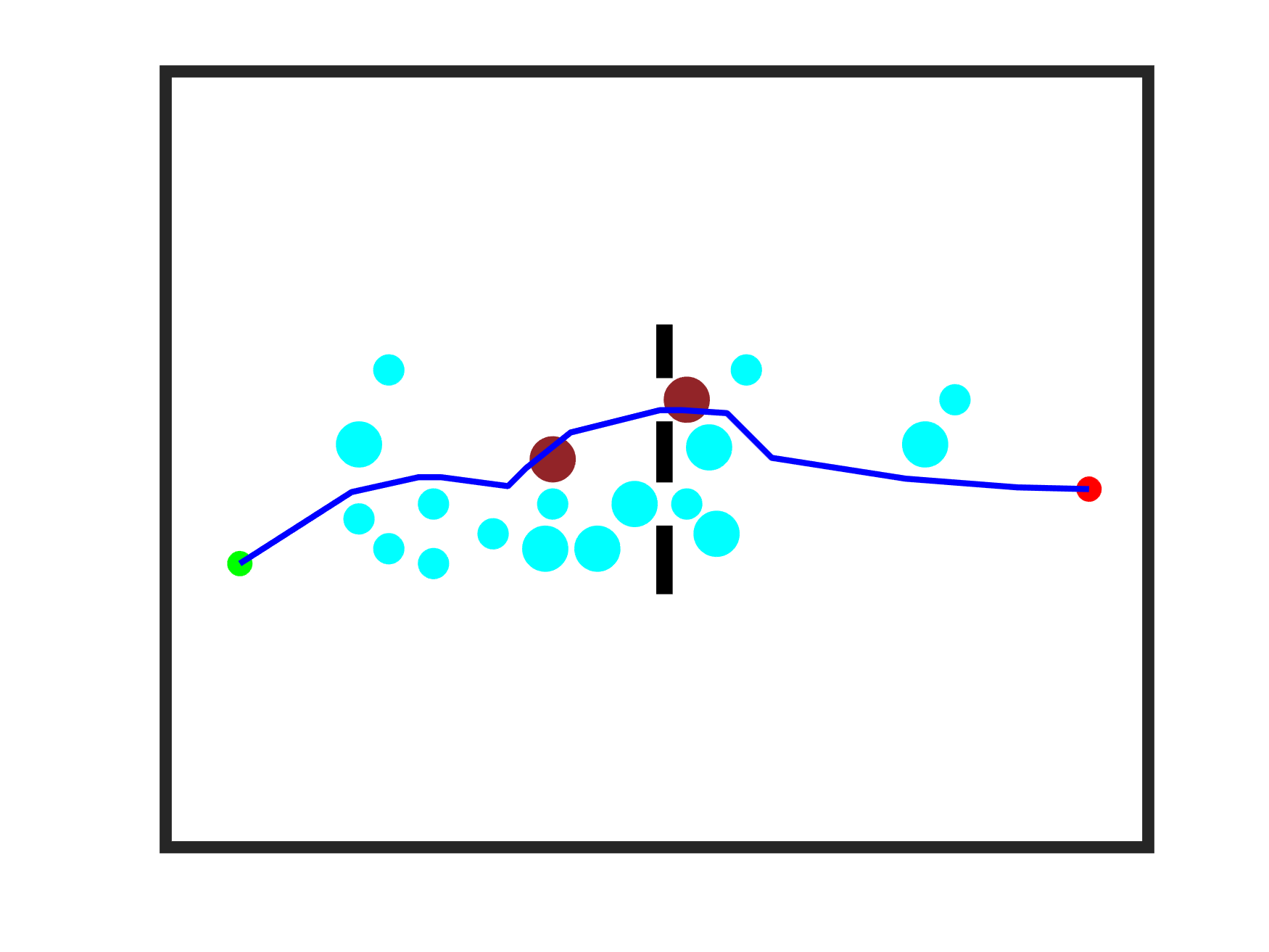}\label{fig:diff3}}
\caption{
(a) Example of \textsf{MOD} solving the classical motion planning problem when a collision-free path exists. 
(b) The \textsf{MOD} cost is modified to obtain the \textsf{MCR} path using the exact approach.
(c) \textsf{HS4MOD} with $m=3$ gives an approximate 2-obstacle \textsf{MCR} path.}
\label{fig:different}
\end{figure*}

\subsection{Two Different Problems}

We have previously noted how \textsf{MOD} encompasses various classes of motion planning problems. 
In this Section, we explore two such problems using a linear robot model. 
Let us consider again the simulated Pepper humanoid robot scenario depicted in Fig.\ref{fig:example_scenario}, where we remove a table blocking the top doorway. 
By employing the exact approach, a solution to the classical motion planning problem is obtained in approximately $8.31$ seconds (visualized in Fig.\ref{fig:diff1}). 
This demonstrates that when a collision-free path exists, \textsf{MOD} resolves the classical motion planning problem with all displacements set to zero.

The \textsf{MCR} problem identifies the minimum number of obstacles that must be removed to yield a feasible path. 
To transform \textsf{MOD} into \textsf{MCR}, we allow collisions with obstacles and modify the cost function~\eqref{eq:disp_def} such that it assigns a cost of $1$ for collision with an obstacle, and $0$ otherwise. 
The exact approach produces a solution in approximately $6.37$ seconds (see Fig.~\ref{fig:diff2}), revealing that at least one obstacle must be removed to achieve a feasible path. \textsf{MCR}, likewise \textsf{MOD}, has been proven to be NP-hard, and \textsf{HS4MOD} may be equally applied to \textsf{MCR}. 
To evaluate this possibility, we solve the problem using $m = 3$, and obtain a 2-obstacle approximation in $0.67$ seconds, as shown in Fig.~\ref{fig:diff3}. 
This result is interesting in so far as \textsf{MCR} is known to have an \( O(\sqrt{n}) \) approximation for rectilinear polygons and disks~\cite{bandyapadhyay2020CG}. 
While the analysis of \textsf{HS4MOD} in the context of \textsf{MCR} is an intriguing avenue for exploration, it falls beyond the scope of this work and is left for future research.
\section{Conclusions and Future Work}
\label{sec:discussion}
This paper formalizes and discusses a \textit{minimum obstacle displacement} (\textsf{MOD}) problem for robot navigation tasks requiring the manipulation of obstacles in the way. 
This work proves that the problem is NP-hard for polygonal obstacles in the plane, even when such obstacles are modeled as simple line segments. 
The paper presents an exact approach, which generalizes a number of existing problems in the literature, which is computationally intractable in the worst case.
Furthermore, the paper discusses a trade-off between path length and the amount of object displacements, and it introduces an approximate albeit sub-optimal horizon-slicing approach characterized by a high computational gain when compared to the exact approach. 
Both approaches are validated using different theoretical examples, scenarios with up to $100$ obstacles, and different robot motion models. 
The solution quality of the approximate method is measured in terms of its optimality gap, and the validations suggest that the approximate method can provide a desirable trade-off between the optimality gap and computation time. 
  
The models presented in the paper are characterized by a few limitations. 
First, we do not deal with obstacles \textit{interacting} with each other, that is, obstacles themselves are not collision-free. Interaction between obstacles can be incorporated by adding collision avoidance constraints among obstacles; similar to~\eqref{eq:cost_fn7} an additional constraint of the form $O(\B{o}_k^j) \cap \bigcup\limits_{i=1, i \neq j}^{|\mathcal{O}|} O(\B{o}_k^i) = \emptyset $ should be added to~\eqref{eq:optimization_problem}.
However, for a large number of obstacles, taking such collisions into account increases the computational burden due to the addition of the corresponding binary variables. 
Second, the approaches discussed in the paper are restricted to robot and obstacle classes allowing for linearizing the collision constraint since the optimization problem for \textsf{MOD} requires the linearisation of the collision constraint in~\eqref{eq:cost_fn7}. 
For the interaction between circles or axis-aligned rectangles, this can be achieved by introducing four binary constraints (for an example, see~\eqref{eq:linear_constraints1}-\eqref{eq:linear_constraints5}). 
Other polygonal projections of the robot and the obstacles require a careful formulation of the linear constraints~\eqref{eq: milp_ploy_1}-\eqref{eq: milp_ploy_2}. 
However, with an increasing number of obstacles, there is a higher computational burden due to the increased number of auxiliary variables.

As argued throughout the paper, we are not concerned about the weight of the obstacles and assume that the computed displacements can be achieved. 
Yet, an obstacle might be too heavy for being moved even though it accounts for a minimum displacement solution. 
This aspect may be incorporated with heavy obstacles being penalized more by assuming a higher weight than other obstacles. 
Future work includes incorporating additional aspects such as the work done by the robot which addresses similar concerns through the force required to move the obstacles. Furthermore, the proposed method is intended as an offline planning approach, where the required displacement for each obstacle is computed prior to execution under the assumption that the environment map is known. In a practical deployment, however, the robot may have access only to a locally observed portion of the environment through its onboard sensors. In such a setting, HS4MOD could be extended to an online incremental planning framework by repeatedly constructing and solving the horizon-sliced optimization problem using the currently observed environment, and updating the plan as the robot moves and new obstacles or obstacle configurations become available. While the development and evaluation of such an online replanning strategy are beyond the scope of the present work, investigating this extension represents an interesting direction for future research.

We consider environments with movable obstacles, where the robot itself is responsible for displacing the obstacles and therefore the focus is on scenarios where no feasible path initially exists for the robot. Thus, the change in the environment arises from deliberate robot–object interactions, rather than from independently evolving or time-varying dynamics.

One could argue that partitioning the planning horizon based on obstacle encounters would be more effective than using equal-length horizon slices. However, such an approach requires either prior knowledge of the solution trajectory or an additional preprocessing stage to identify the obstacle encounter events, thereby increasing the computational complexity. In contrast, equal horizon slicing provides a simple, general, and computationally efficient decomposition strategy that can be applied without any prior knowledge of the trajectory. Nevertheless, the investigation of adaptive horizon partitioning strategies represents an interesting direction for future research.

Finally, it is worth noting that the evaluation domains used in this paper have been rather abstract in nature. 
Yet, this is by no means a limitation, and many practical applications exist to the minimum obstacle displacement problem, such as for example those related to automated logistics \cite{Mohamedetal2019}.
\bibliographystyle{elsarticle-num}
\bibliography{references}
\end{document}